%% file: main.tex
\makeatletter
\def\input@path{{styles/nvidia/}}
\makeatother
\documentclass[10pt,twocolumn,logo,copyright]{nvidiatechreport}

\ifPDFTeX\else
  \microtypesetup{tracking=false}
\fi

\graphicspath{{styles/nvidia/}{figures/}}
\input{preamble_common}

\renewcommand{\tablefont}{\scriptsize}
\renewcommand{\earlyfloat}[1]{\input{#1}}
\renewcommand{\latefloat}[1]{}
\renewcommand{\reportonly}[1]{#1}
\renewenvironment{colgather}{\begin{gathered}}{\end{gathered}}
\renewcommand{\colpair}{\\}
\renewcommand{\colbreak}{\\}
\makeatletter
\def\@secpenalty{0}
\g@addto@macro\appendix{}
\makeatother
\AtEndDocument{}

\setlist[itemize]{noitemsep,labelsep=0.3em}
\usepackage[nameinlink,capitalise]{cleveref}
\crefname{section}{Sec.}{Secs.}
\Crefname{section}{Section}{Sections}
\crefname{table}{Tab.}{Tabs.}
\Crefname{table}{Table}{Tables}
\input{metadata}

\fancypagestyle{plain}{%
  \fancyhead[L,R]{}
  \fancyhead[C]{\footerfont \papertitle}
  \fancyfoot[C]{}
  \fancyfoot[L]{}
  \fancyfoot[R]{\footerfont \thepage}
}

\hypersetup{
  colorlinks=true,
  linkcolor=fpblue,
  citecolor=fpblue,
  urlcolor=fpblue,
  pdftitle={\papertitle},
  pdfauthor={Yuhang He, Harrison Zhang, Fanrong Li, Chang Liu, RuQing G. Xu, Perkz Zheng, Zhenhua Wang, Rajeev Rao, Mengdi Wang, Sharan Chetlur, Jun Yang},
  pdfsubject={NVIDIA Technical Report},
  pdfkeywords={attention, quantization, FP4, NVFP4, video generation, diffusion transformer}
}

\renewcommand{\today}{November 2026}
\reportnumber={}

\title{\papertitle}
\author[1,*]{Yuhang He}
\author[1,*]{Harrison Zhang}
\author[1]{Fanrong Li}
\author[1]{Chang Liu}
\author[1]{RuQing G.~Xu}
\author[1]{Perkz Zheng}
\author[1]{Zhenhua Wang}
\author[1]{Rajeev Rao}
\author[1]{Mengdi Wang}
\author[1]{Sharan Chetlur}
\author[1]{Jun Yang}
\affil[1]{NVIDIA}
\affil[*]{Equal contribution}
\date{}

\begin{document}
\begin{abstract} {\noindent\Large\bfseries Abstract\par}\medskip \input{docs/abstract}
\end{abstract}

\maketitle
\suppressfloats  

\input{paper}
\end{document}

%% file: preamble_common.tex
\usepackage{mathtools}
\usepackage{bm}
\usepackage{natbib}
\usepackage{xspace}
\usepackage{float}
\usepackage{needspace}
\usepackage{multirow}
\usepackage{makecell}
\usepackage{colortbl}
\usepackage{tikz}
\usetikzlibrary{arrows.meta,positioning,calc,fit,decorations.pathreplacing}

\definecolor{fpblue}{HTML}{1F77B4}
\definecolor{bodyfill}{HTML}{DCE9F5}
\definecolor{compfill}{HTML}{F8DDC4}
\definecolor{resfill}{HTML}{EDA46B}
\definecolor{vfill}{HTML}{DCEFD8}
\definecolor{keyink}{HTML}{2A78D6}
\definecolor{compink}{HTML}{D9581F}
\definecolor{vcompfill}{HTML}{9ED39A}
\definecolor{vcompink}{HTML}{3E9A3A}
\definecolor{basisfill}{HTML}{D6CBEE}
\definecolor{basisink}{HTML}{6A4FB3}
\definecolor{wfill}{HTML}{BFE5DC}
\definecolor{wink}{HTML}{2A8C7A}
\definecolor{bestfill}{HTML}{DCE9F5}

\newcommand{\earlyfloat}[1]{}
\newcommand{\latefloat}[1]{\input{#1}}
\newcommand{\llmstatement}{}
\newcommand{\reportonly}[1]{}

\newcommand{\tablefont}{\footnotesize}
\newcommand{\widetablefont}{\scriptsize}
\newenvironment{spreadtabular}[1]{\begin{tabular*}{\linewidth}{@{\extracolsep{\fill}}#1}}{\end{tabular*}}
\usepackage{tabularx}
\newcolumntype{Y}{>{\centering\arraybackslash}X}
\newenvironment{joinedtabular}[1]{\tabularx{\linewidth}{#1}}{\endtabularx}
\usepackage{algorithm}
\usepackage{algpseudocode}

\newcommand{\tablehead}[1]{\textbf{\boldmath #1}}
\newcommand{\colplot}[1]{\includegraphics{#1}}
\newcommand{\colplotwide}[1]{\includegraphics[width=\linewidth]{#1}}
\newcommand{\colfigwidth}{\linewidth}
\newcommand{\algfont}{\footnotesize}
\newcommand{\best}[1]{\cellcolor{bestfill}#1}
\newcommand{\tablecite}[1]{}
\newcommand{\spancell}[3]{\multirow{#1}{*}[-\dimexpr(\aboverulesep+\belowrulesep+\cmidrulewidth)*#2/2\relax]{#3}}

\newenvironment{colgather}{}{}
\newcommand{\colpair}{\qquad}
\newcommand{\colbreak}{}

\newcommand{\deco}{DeCo-Attention\xspace}
\newcommand{\qkpv}{\mbox{QK16PV8}\xspace}

\newcommand{\Qq}{\mathcal{Q}}

\newcommand{\cfgmain}{\mbox{DeCo-Attention, FP8 PV}\xspace}
\newcommand{\cfgfour}{\mbox{DeCo-Attention, NVFP4 PV}\xspace}

%% file: metadata.tex
\newcommand{\papertitle}{DeCo-Attention: Decompose to Compensate for Low-Precision Attention}

%% file: docs/abstract.tex
Self-attention over tens of thousands of tokens dominates the generation time of video diffusion transformers,
because its cost grows quadratically with sequence length. The FP4 matrix products of modern GPUs can speed up
attention, but the 4-bit rounding of the queries $Q$ and the keys $K$ introduces large errors. We show that the
exponential in softmax amplifies the error of their product $QK$, which therefore dominates the attention error. We
propose \deco, a co-design of algorithm and system for low-precision attention. \deco is training-free and
compensates the rounding error of $Q$ and $K$ along a few principal directions found by a per-head
eigendecomposition. The lightweight compensation terms are appended to $Q$ and $K$ as extra channels,
so the FP4 matrix product corrects its own rounding error while it computes the attention scores. Once the products
run in FP4, the softmax bounds the kernel, and we optimize the softmax so that the gain of FP4 computation
translates into realized speedup. With
$QK$ in FP4, we implement \deco with the rest of attention in FP8 on GB300 and VR200 and in FP4 on RTX PRO
6000. On Wan2.2,
these kernels run 2.13$\times$, 2.49$\times$, and 2.82$\times$ faster than BF16 attention, and the whole generation
runs 1.55$\times$, 1.68$\times$, and 1.80$\times$ faster. On four video generation models, the FP8 configuration
achieves LPIPS comparable to prior 8-bit $QK$ methods, and the fully FP4 configuration lower LPIPS than prior fully
4-bit kernels.

%% file: paper.tex
\section{Introduction}
\label{sec:intro}

\begin{figure}[tbp]
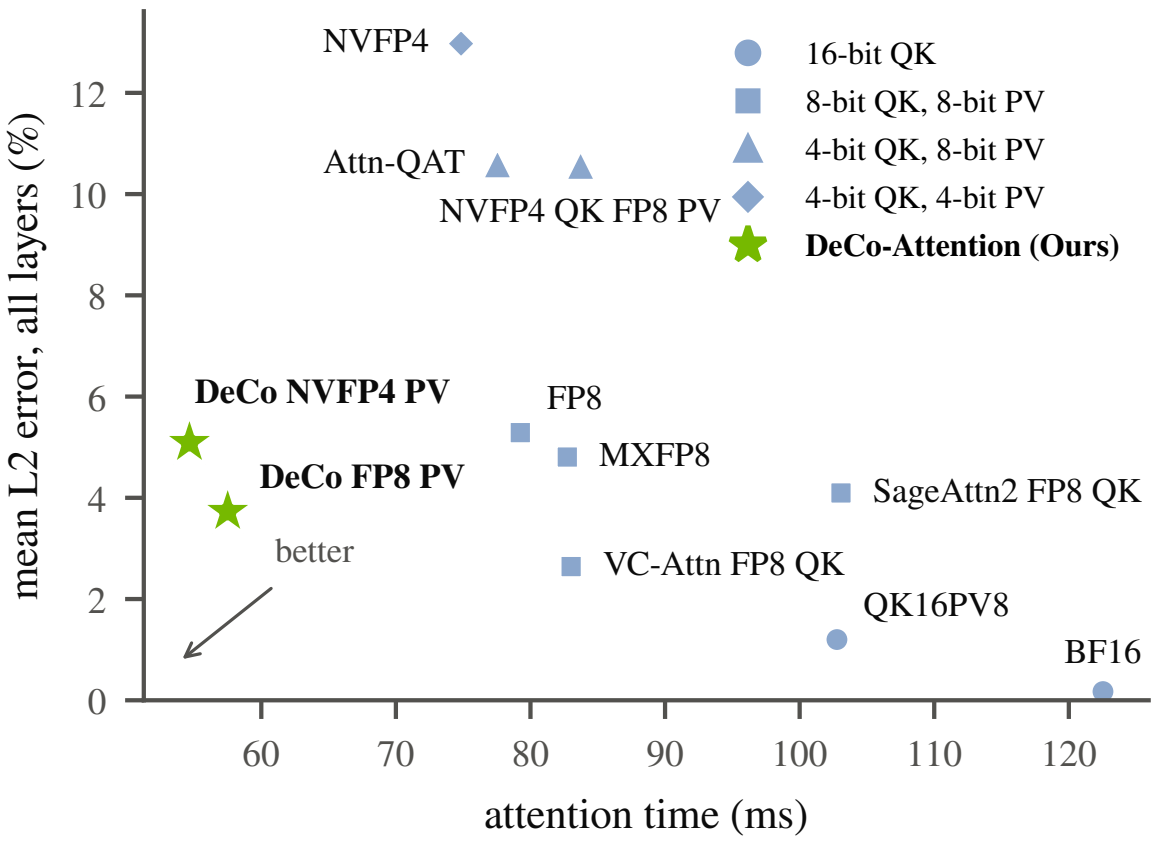

\centering
\colplot{plots/speed_accuracy_wan22.pdf}
\caption{\textbf{Accuracy and speed.} Mean relative L2 error over all layers of Wan2.2 against the attention time of a
Wan2.2 call on GB300. With a 4-bit $QK$, \deco advances the Pareto frontier of accuracy and speed.}
\label{fig:speed-accuracy}
\end{figure}

Video diffusion transformers (DiTs)~\citep{dit} have become the mainstream approach to video generation, as in
Wan~\citep{wan}, HunyuanVideo~\citep{hunyuanvideo,hunyuanvideo15}, CogVideoX~\citep{cogvideox},
LongCat-Video~\citep{longcatvideo}, and Movie Gen~\citep{moviegen}. These models generate a video by denoising the
latent of the whole clip over tens of steps, and in every step each self-attention layer attends over all spatiotemporal
tokens of the clip. Because of its quadratic cost and the large number of tokens, attention dominates the generation
time. For example, an 81-frame 720p clip of Wan2.2 has 75,600 tokens. Generating it with
TensorRT-LLM~\citep{tensorrtllm} takes 302.7~s on GB300, and attention takes 67.6\% of each step.

Modern GPUs support increasingly low-precision data formats on their tensor cores. Hopper introduced FP8 matrix
multiply-accumulates (MMAs)~\citep{fp8formats}, and Blackwell and Rubin added MMAs in block-scaled FP4 formats such as
NVFP4~\citep{microscaling,nvfp4pretrain}. On GB300 and VR200, the FP4 MMAs run at two to three times the FP8 rate and
six to nearly nine times the BF16 rate~\citep{nvfp4pretrain,verarubinnvl72}. Quantizing attention to these formats is
thus a direct way to accelerate it, but the lower precision can compromise the accuracy.

Many recent methods study low-precision attention and achieve promising
results~\citep{fa3,sage1,sage2,vcattention,attnqat,sage3}. However, the existing kernels with a 4-bit $QK$ have large errors in the attention output
(\cref{fig:speed-accuracy}), which cause a non-negligible quality loss in the generated videos. This accuracy loss is the main
challenge of accelerating attention with the FP4 tensor cores. Our analysis shows that the loss comes
from the exponential in softmax. The exponential turns the error that the rounding of $Q$ and $K$ causes in each score
into a relative error of the attention weight, so this rounding error dominates the error of attention and of the
generated video.

To address this challenge, we propose \deco (Decompose to Compensate), a co-design of algorithm and system that
decomposes the quantization error of $QK$ and compensates its main part inside the FP4 $QK$ product. Specifically,
it decomposes this error with a per-head eigendecomposition of $Q$ and $K$ and keeps its components along the top
$r$ eigenvectors.\reportonly{ We show that most of this error comes from these few top directions.} These compensation
terms are appended to $Q$ and $K$ as extra channels, so the same FP4 MMA computes
the scores together with their correction. To keep this lightweight, \deco computes the
eigenvectors from a random sample of 1/256 of the tokens and shares them between $Q$ and $K$, so
one eigendecomposition per head suffices and adds only a few percent to the attention time. The 4-bit $QK$ is now
accurate, but the softmax between the two MMAs becomes the kernel bottleneck. To turn the speed of FP4 into a
kernel speedup, we propose I-ExpCast and adopt the row sum in the $PV$ MMA~\citep{hifa4} and the lazy
rescale~\citep{fa4} to shorten the softmax.

We implement \deco for GB300, VR200, and RTX PRO 6000, with $PV$ in FP8 on the first two GPUs and in FP4 on the
third, and evaluate it on four video generation models. With FP8 $PV$, \deco is on par with FP8-$QK$ methods in
generation quality and narrows the gap to INT8-$QK$ attention. With FP4 $PV$ on RTX PRO 6000, it is more accurate
than prior fully 4-bit kernels. On Wan2.2, the kernels of \deco
run 2.13$\times$, 2.49$\times$, and 2.82$\times$ faster than the fastest BF16 kernel, and the end-to-end generation
runs 1.55$\times$, 1.68$\times$, and 1.80$\times$ faster.\reportonly{ \deco is the fastest of the compared attention
kernels and gives the largest end-to-end speedup on every GPU.}

In summary, this paper makes three contributions.
\begin{itemize}[leftmargin=*,itemsep=1pt,topsep=2pt]
\item \textbf{Accuracy.} We analyze the quantization error of a \mbox{4-bit} $QK$ and propose \deco, a training-free and
calibration-free method that compensates this error directly inside attention. \deco makes a 4-bit $QK$ competitive
with 8-bit $QK$ methods.
\item \textbf{Performance.} We implement high-performance kernels for \deco in CuTe DSL. The FP4 $QK$ product exposes
the softmax as the bottleneck of low-precision attention. Our kernels relieve this bottleneck and turn the speed of the FP4 $QK$
into realized speedup.
\item \textbf{Evaluation.} A systematic evaluation on three GPU platforms, at the kernel level and end to end, validates the
accuracy and the speedups of \deco. We show that \deco advances the Pareto frontier of accuracy and speed
(\cref{fig:speed-accuracy}).
\end{itemize}

\input{docs/background}

\input{docs/method}
\input{docs/performance}

\section{Evaluation}
\label{sec:evaluation}

\input{docs/setup}
\input{docs/results}

\section{Conclusion}
\label{sec:conclusions}

We presented \deco, a co-design of algorithm and system for low-precision attention. We showed that the rounding error
of a 4-bit $QK$ lies mostly along a few principal directions, which \deco compensates inside the matrix product that
computes the scores. Softmax optimizations then turn the speed of the FP4 products into kernel speedup. On four video
models, \deco with FP8 $PV$ competes with the generation quality of FP8-$QK$ methods, its fully 4-bit kernel is more
accurate than prior 4-bit kernels, and it is the fastest of the compared kernels on GB300, VR200, and RTX PRO 6000.
\deco thereby advances the Pareto frontier of accuracy and speed.

\deco has limitations that we plan to address. Our kernels target video diffusion models, and LLM inference
with KV cache cannot yet use them directly. The runtime
estimation of the principal directions adds a small overhead to each call, which matters in low-latency scenarios.
Over all layers, the error of \deco is not yet below that of the most accurate FP8-$QK$ kernel. The kernel code will be released publicly.

\llmstatement
\bibliography{references}

\ifdefined\mainonly\else  
\appendix
\input{docs/appendix}

\fi

%% file: docs/background.tex
\section{Related Work}
\label{sec:background}
\label{sec:related}

\input{figures/overview}

\input{docs/related}

%% file: figures/overview.tex
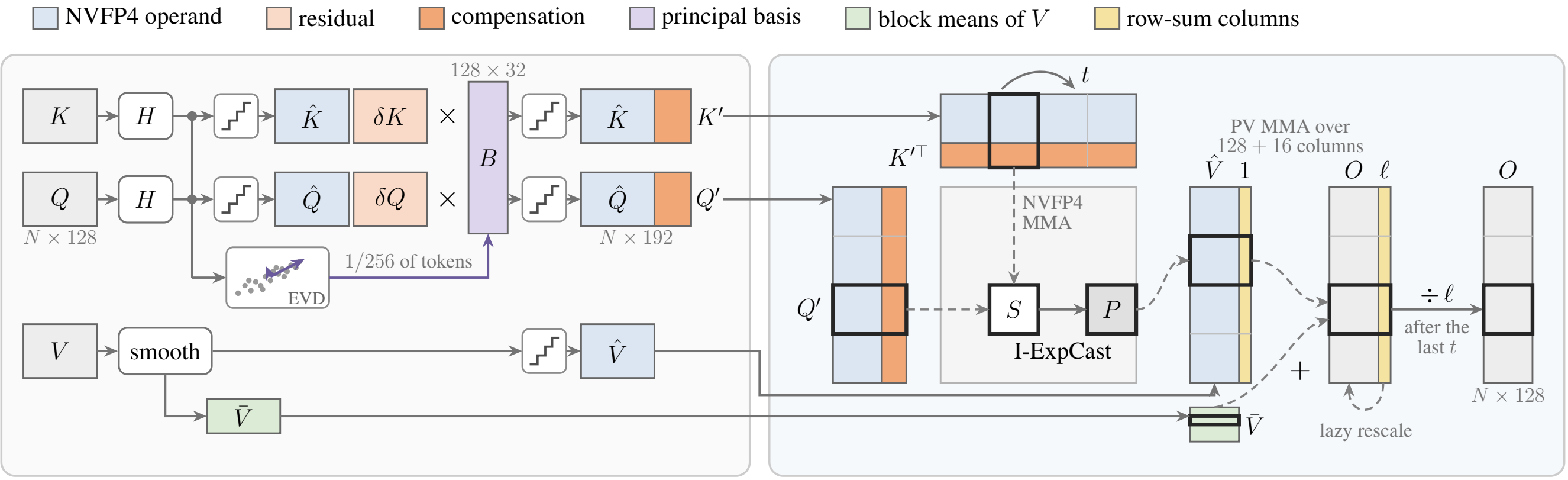
\begin{figure*}[t]
\centering
\definecolor{opfill}{HTML}{DCE6F2}
\definecolor{lossfill}{HTML}{F9D9C7}
\definecolor{fixfill}{HTML}{F2A777}
\definecolor{basefill}{HTML}{DDD5EE}
\definecolor{baseink}{HTML}{6A5A9C}
\definecolor{meanfill}{HTML}{DCEBD5}
\definecolor{sumfill}{HTML}{F5E3A1}
\providecommand{\fs}[1]{\fontsize{#1}{#1}\selectfont}
\resizebox{\textwidth}{!}{%
\begin{tikzpicture}[x=1cm, y=1cm, font=\fs{14}, >={Stealth[length=8pt, width=6.5pt]},
  mat/.style={draw=black!60, line width=1pt},
  dims/.style={font=\fs{11.5}, inner sep=2pt, text=black!55},
  op/.style={font=\fs{16.5}},
  stage/.style={draw=black!20, rounded corners=8pt, line width=1.2pt},
  icon/.style={draw=black!45, rounded corners=4pt, fill=white, line width=1pt},
  module/.style={draw=black!55, rounded corners=5pt, fill=white, line width=1.1pt, minimum width=1.1cm,
                 minimum height=0.95cm, inner sep=3pt},
  flow/.style={->, black!55, line width=1.3pt},
  link/.style={->, black!50, line width=1.2pt, dash pattern=on 5pt off 3pt},
  tile/.style={draw=black!85, line width=2.2pt},
  junction/.style={circle, fill=black!55, inner sep=2pt},
  panel/.style={font=\fs{15}, anchor=north},
  quant/.pic={
    \draw[icon] (-0.45, -0.45) rectangle (0.45, 0.45);
    \draw[black!70, line width=1.3pt] (-0.3, -0.27) -- (-0.14, -0.27) -- (-0.14, -0.09) -- (0.02, -0.09) -- (0.02, 0.09)
      -- (0.16, 0.09) -- (0.16, 0.27) -- (0.3, 0.27);
  },
  evd/.pic={
    \draw[icon] (-1.05, -0.62) rectangle (1.05, 0.62);
    \foreach \x/\y in {-0.33/-0.13, -0.26/-0.07, -0.2/-0.11, -0.15/-0.04, -0.1/-0.07, -0.05/0.0, 0.0/-0.03, 0.03/0.04,
                       0.08/0.02, 0.12/0.06, 0.17/0.04, 0.21/0.09, 0.26/0.07, 0.3/0.12, -0.22/-0.03, 0.06/0.07,
                       -0.08/-0.1, 0.14/0.0}
      \fill[black!40] (2*\x - 0.15, 2.6*\y + 0.02) circle (1.6pt);
    \draw[->, baseink, line width=1.6pt] (-0.15, 0.02) -- (0.55, 0.38);
    \draw[->, baseink, line width=1.2pt] (-0.15, 0.02) -- (-0.25, 0.24);
    \node[font=\fs{10.5}, text=black!60, anchor=south east, inner sep=2pt] at (1.05, -0.62) {EVD};
  }]
\begin{scope}[shift={(0.6, 8.75)}]
  \node[anchor=west, inner sep=0] (l0) at (0, 0) {};
  \foreach \c/\t [count=\i, remember=\i as \j (initially 0)] in {opfill/NVFP4 operand, lossfill/residual,
      fixfill/compensation, basefill/principal basis, meanfill/block means of $V$, sumfill/{row-sum columns}} {
    \draw[mat, fill=\c] ($(l\j.east) + (\ifnum\i=1 0\else 0.9\fi, -0.22)$) rectangle ++(0.5, 0.44) coordinate (s\i);
    \node[anchor=west, inner sep=0] (l\i) at ($(s\i) + (0.15, -0.22)$) {\t};
  }
\end{scope}
\draw[stage, fill=black!2] (0, -0.6) rectangle (15.3, 8.0);
\node[panel] at (7.65, -0.75) {(a) Preparation, per head};
\foreach \y/\x in {6.75/K, 5.05/Q} {
  \draw[mat, fill=black!7] (0.45, \y - 0.55) rectangle (1.95, \y + 0.55);
  \node at (1.2, \y) {$\x$};
  \draw[flow] (1.95, \y) -- (2.4, \y);
  \node[module] at (2.95, \y) {$H$};
  \draw[flow] (3.5, \y) -- (4.35, \y);
  \pic at (4.8, \y) {quant};
  \draw[flow] (5.25, \y) -- (5.6, \y);
  \draw[mat, fill=opfill] (5.6, \y - 0.55) rectangle (7.1, \y + 0.55);
  \node at (6.35, \y) {$\hat \x$};
  \draw[mat, fill=lossfill] (7.2, \y - 0.55) rectangle (8.7, \y + 0.55);
  \node at (7.95, \y) {$\delta \x$};
  \node[op] at (9.12, \y) {$\times$};
  \draw[flow] (10.35, \y) -- (10.65, \y);
  \pic at (11.1, \y) {quant};
  \draw[flow] (11.55, \y) -- (11.85, \y);
  \fill[opfill] (11.85, \y - 0.55) rectangle (13.35, \y + 0.55);
  \fill[fixfill] (13.35, \y - 0.55) rectangle (14.1, \y + 0.55);
  \draw[mat] (11.85, \y - 0.55) rectangle (14.1, \y + 0.55) (13.35, \y - 0.55) -- (13.35, \y + 0.55);
  \node at (12.6, \y) {$\hat \x$};
  \node[anchor=west, inner sep=2pt] at (14.12, \y) {$\x'$};
}
\node[dims, anchor=north] at (1.2, 4.5) {$N \times 128$};
\node[dims, anchor=north] at (12.975, 4.5) {$N \times 192$};
\draw[mat, fill=basefill] (9.55, 4.35) rectangle (10.35, 7.45);
\node at (9.95, 5.9) {$B$};
\node[dims, anchor=south] at (9.95, 7.45) {$128 \times 32$};
\draw[black!55, line width=1.3pt] (3.9, 6.75) -- (3.9, 3.45);
\draw[flow] (3.9, 3.45) -- (4.6, 3.45);
\node[junction] at (3.9, 6.75) {};
\node[junction] at (3.9, 5.05) {};
\pic at (5.65, 3.45) {evd};
\draw[flow, baseink] (6.7, 3.45) -- (9.95, 3.45) -- (9.95, 4.35);
\node[dims, anchor=south] at (8.325, 3.47) {$1/256$ of tokens};
\draw[mat, fill=black!7] (0.45, 1.4) rectangle (1.95, 2.5);
\node at (1.2, 1.95) {$V$};
\node[module, minimum width=1.9cm] (smooth) at (3.35, 1.95) {smooth};
\draw[flow] (1.95, 1.95) -- (smooth.west);
\draw[flow] (smooth.east) -- (10.65, 1.95);
\pic at (11.1, 1.95) {quant};
\draw[flow] (11.55, 1.95) -- (11.85, 1.95);
\draw[mat, fill=opfill] (11.85, 1.4) rectangle (13.35, 2.5);
\node at (12.6, 1.95) {$\hat V$};
\draw[flow] (smooth.south) -- (3.35, 0.62) -- (4.2, 0.62);
\draw[mat, fill=meanfill] (4.2, 0.27) rectangle (5.7, 0.97);
\node at (4.95, 0.62) {$\bar V$};

\draw[stage, fill=opfill!25] (15.7, -0.6) rectangle (31.95, 8.0);
\node[panel] at (23.825, -0.75) {(b) Attention kernel, per tile};
\fill[opfill] (19.2, 6.2) rectangle (23.2, 7.2);
\fill[fixfill] (19.2, 5.7) rectangle (23.2, 6.2);
\draw[mat] (19.2, 5.7) rectangle (23.2, 7.2) (19.2, 6.2) -- (23.2, 6.2);
\foreach \x in {20.2, 21.2, 22.2} \draw[black!25, line width=0.8pt] (\x, 5.7) -- (\x, 7.2);
\draw[tile] (20.2, 5.7) rectangle (21.2, 7.2);
\node[anchor=east, inner sep=5pt] at (19.15, 5.95) {$K'^{\top}$};
\draw[->, black!55, line width=1.3pt] (20.45, 7.3) .. controls (20.9, 7.75) and (21.5, 7.75) .. (21.95, 7.3);
\node[anchor=west, inner sep=2pt] at (22.0, 7.6) {$t$};
\fill[opfill] (17.0, 1.3) rectangle (18.0, 5.3);
\fill[fixfill] (18.0, 1.3) rectangle (18.5, 5.3);
\draw[mat] (17.0, 1.3) rectangle (18.5, 5.3) (18.0, 1.3) -- (18.0, 5.3);
\foreach \y in {2.3, 3.3, 4.3} \draw[black!25, line width=0.8pt] (17.0, \y) -- (18.5, \y);
\draw[tile] (17.0, 2.3) rectangle (18.5, 3.3);
\node[anchor=east, inner sep=5pt] at (16.95, 2.8) {$Q'$};
\fill[black!4] (19.2, 1.3) rectangle (23.2, 5.3);
\draw[black!30, line width=1pt] (19.2, 1.3) rectangle (23.2, 5.3);
\draw[tile, fill=white] (20.2, 2.3) rectangle (21.2, 3.3);
\node at (20.7, 2.8) {$S$};
\draw[link] (18.5, 2.8) -- (20.2, 2.8);
\draw[link] (20.7, 5.7) -- (20.7, 3.3);
\node[dims, anchor=north west, align=left] at (20.8, 5.2) {NVFP4\\MMA};
\draw[tile, fill=black!12] (22.2, 2.3) rectangle (23.2, 3.3);
\node at (22.7, 2.8) {$P$};
\draw[flow] (21.2, 2.8) -- (22.2, 2.8);
\node[anchor=north, inner sep=2pt] at (21.7, 2.2) {I-ExpCast};
\fill[opfill] (24.3, 1.3) rectangle (25.3, 5.3);
\fill[sumfill] (25.3, 1.3) rectangle (25.55, 5.3);
\draw[mat] (24.3, 1.3) rectangle (25.55, 5.3) (25.3, 1.3) -- (25.3, 5.3);
\foreach \y in {2.3, 3.3, 4.3} \draw[black!25, line width=0.8pt] (24.3, \y) -- (25.55, \y);
\draw[tile] (24.3, 3.3) rectangle (25.55, 4.3);
\node[anchor=south, inner sep=5pt] at (24.8, 5.3) {$\hat V$};
\node[anchor=south, inner sep=5pt] at (25.425, 5.3) {$1$};
\draw[mat, fill=meanfill] (24.3, 0.1) rectangle (25.3, 0.8);
\draw[tile] (24.3, 0.45) rectangle (25.3, 0.625);
\node[anchor=west, inner sep=2pt] at (25.35, 0.45) {$\bar V$};
\draw[link] (23.2, 2.8) .. controls (23.75, 2.8) and (23.75, 3.8) .. (24.3, 3.8);
\node[dims, anchor=south, align=center] at (26.35, 5.9) {PV MMA over\\$128 + 16$ columns};
\fill[black!7] (27.15, 1.3) rectangle (28.15, 5.3);
\fill[sumfill] (28.15, 1.3) rectangle (28.4, 5.3);
\draw[mat] (27.15, 1.3) rectangle (28.4, 5.3) (28.15, 1.3) -- (28.15, 5.3);
\foreach \y in {2.3, 3.3, 4.3} \draw[black!25, line width=0.8pt] (27.15, \y) -- (28.4, \y);
\draw[tile] (27.15, 2.3) rectangle (28.4, 3.3);
\node[anchor=south, inner sep=5pt] at (27.65, 5.3) {$O$};
\node[anchor=south, inner sep=5pt] at (28.275, 5.3) {$\ell$};
\draw[link] (25.55, 3.8) .. controls (26.35, 3.8) and (26.35, 2.95) .. (27.15, 2.95);
\draw[link] (24.8, 0.8) .. controls (25.9, 1.0) and (26.4, 2.4) .. (27.15, 2.6);
\node[op] at (26.55, 1.55) {$+$};
\draw[link] (28.25, 1.3) .. controls (28.25, 0.6) and (27.55, 0.6) .. (27.55, 1.3);
\node[dims, anchor=north] at (27.9, 0.55) {lazy rescale};
\draw[flow] (28.4, 2.8) -- (30.3, 2.8);
\node[anchor=south, inner sep=3pt] at (29.35, 2.8) {$\div\,\ell$};
\node[dims, anchor=north, align=center, inner sep=3pt] at (29.35, 2.7) {after the\\last $t$};
\fill[black!7] (30.3, 1.3) rectangle (31.3, 5.3);
\draw[mat] (30.3, 1.3) rectangle (31.3, 5.3);
\foreach \y in {2.3, 3.3, 4.3} \draw[black!25, line width=0.8pt] (30.3, \y) -- (31.3, \y);
\draw[tile] (30.3, 2.3) rectangle (31.3, 3.3);
\node[anchor=south, inner sep=5pt] at (30.8, 5.3) {$O$};
\node[dims, anchor=north] at (30.8, 1.28) {$N \times 128$};
\draw[flow] (14.75, 6.75) -- (19.2, 6.75);
\draw[flow] (14.75, 5.05) -- (17.0, 5.05);
\draw[flow] (13.35, 1.95) -- (15.5, 1.95) -- (15.5, 1.05) -- (24.8, 1.05) -- (24.8, 1.3);
\draw[flow] (5.7, 0.62) -- (24.3, 0.62);
\end{tikzpicture}}
\caption{\textbf{Overview of \deco.} (a) For every head, $Q$ and $K$ are rotated by a Hadamard matrix
$H$ and quantized to NVFP4, leaving the residuals $\delta Q$ and $\delta K$. An eigendecomposition (EVD) of the
Gram matrices of 1/256 of the tokens gives the 32 principal directions $B$. The projections of the residuals
and of the quantized operands onto $B$ are appended as NVFP4 compensation channels, so $Q'$ and $K'$ have 192
channels. $V$ is smoothed by subtracting its block means $\bar V$ and quantized to $\hat V$. (b) Per tile, one NVFP4
MMA over $Q'$ and $K'$ computes the compensated scores $S$, I-ExpCast forms $P$, and the $PV$ MMA
returns the output with the row sums $\ell$ of $P$ through columns of ones appended to $\hat V$. The block means
are added back, the softmax rescales lazily, and $O$ is finally divided by $\ell$.}
\label{fig:overview}
\end{figure*}

%% file: docs/related.tex
\subsection{Low-precision attention}
\label{sec:related-lowp}

Low-precision attention speeds up attention by computing its two matrix products with the 8-bit and 4-bit MMAs of
the tensor cores. At 8 bits, FlashAttention-3 runs attention in FP8 with block scales and a Hadamard rotation~\citep{fa3}, and
SageAttention, SageAttention2, and VC-Attention smooth or center $Q$ and $K$ to keep the accuracy of an 8-bit or 4-bit
$QK$~\citep{sage1,sage2,vcattention}. With a 4-bit $QK$ and an 8-bit $PV$, the NVFP4 kernels of TensorRT-LLM,
cuDNN, and FlashInfer quantize $Q$ and $K$ with one scale per 16
elements~\citep{tensorrtllm,cudnn,flashinfer,nvfp4pretrain}, and Attn-QAT retrains the model for such a
kernel~\citep{attnqat}. Fully 4-bit kernels such as SageAttention3, PAROAttention, and HiFA4 also quantize $P$ and
$V$~\citep{sage3,paroattention,hifa4}, and ThriftAttention, DraftAttention2, and DMA mix two precisions across the
score matrix~\citep{thriftattention,draftattention2,dma}. These methods reduce the quantization error by smoothing, by Hadamard
rotations~\citep{fa3,mxattention}, or by finer scales. However, a floating-point element keeps its relative rounding error at any scale, so the rounding error
of $Q$ and $K$ stays in the scores and grows from FP8 to FP4. \deco compensates this error inside the $QK$ product.

\subsection{Residual and low-rank compensation}
\label{sec:related-compensation}

Compensation recovers the accuracy lost in rounding by adding the main part of the residual back. Residual products refine FP16 matrix products on tensor
cores~\citep{markidis,ootomo}. LoRC, LQER, QERA, and SVDQuant compensate the rounding error or the outliers of static
weights offline with low-rank factors~\citep{zeroquantv2,lqer,qera,svdquant}.
ResQ keeps the activations along principal directions of
calibration data in 8 bits and the others in 4 bits~\citep{resq}. ARCQuant appends the residuals of
outlier activation channels as extra channels of the same GEMM~\citep{arcquant}. MpFA restores the mean term of the centered $Q$ and $K$ of an NVFP4 $QK$ in an
additional BF16 MMA~\citep{mpfa}. These methods rely on calibration data, a static operand, or a second product, which normally costs more than an
FP8 $QK$. \deco compensates the rounding error of both $Q$ and $K$ along directions estimated at runtime for every
head, in the same FP4 MMA as the scores.

\subsection{Efficient attention}
\label{sec:related-efficient}

With low-precision matrix products, the softmax and the output rescale bound the attention kernel. FlashAttention-4 computes part of the exponentials on the fused multiply-add (FMA) units in parallel with the
special-function unit (SFU), which still occupies the SFU and takes many instructions~\citep{fa4}. Its lazy rescale moves the row base, which the online softmax keeps at the
running maximum of a row~\citep{flashattention,onlinesoftmax}, only when a tile exceeds it by more than a
threshold. ExpCast of VC-Attention forms the FP8 code of a probability from its score with one FMA,
because the bit pattern of a floating-point number read as an integer grows linearly with the logarithm of its
value~\citep{vcattention,schraudolph}. Direct-P forms the E2M1 codes of an MXFP4 $P$ from a
linear function of the score and a hardware conversion~\citep{fa4fp4}. HiFA4 appends a column of ones to $V$, so that the $PV$ product returns the row sum
of $P$~\citep{hifa4}. \deco improves ExpCast and adopts the lazy rescale and the row sum in the $PV$ MMA
to build high-performance kernels.\reportonly{\looseness=-1}  

%% file: docs/method.tex
\section{DeCo-Attention}
\label{sec:method}

This section presents \deco (\cref{fig:overview}). We first show that the exponential
in softmax amplifies the quantization error of $QK$. We then introduce \textit{principal residual compensation}, which
corrects the main part of this error in extra channels of $Q$ and $K$. Finally, we describe the kernel optimizations
that turn the speed of FP4 into realized speedup. The appendix gives the derivations and details.

\input{figures/overview}
\subsection{Error of QK quantization}
\label{sec:motivation}
\label{sec:numerics}
\label{sec:num-motivation}

\begin{figure}[tbp]
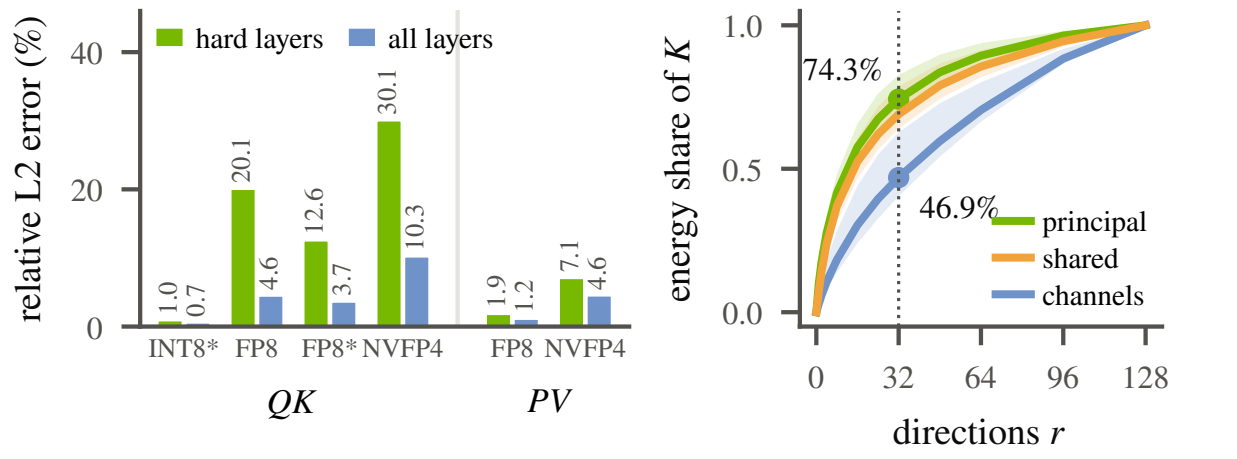

\centering
\colplot{plots/motivation_wan22-t2v-a14b.pdf}
\caption{\textbf{Quantization error and energy of K.} (a) Relative L2 error on Wan2.2 with only $Q$ and $K$ or only
$P$ and $V$ quantized (*, per-block scales of SageAttention). (b) Share of the energy of $K$ along $r$ directions over the
heads of Wan2.2.}
\label{fig:motivation}
\label{fig:qk-formats}
\label{fig:qk-bases}
\end{figure}

FP4 MMAs offer the highest throughput of the tensor cores on modern GPUs, but the existing kernels with a 4-bit $QK$
lose accuracy. To locate the source of this loss, we quantize one product of attention at a
time and report the relative L2 error~\citep{higham} of the attention output against FP32 attention. On the hard
layers\footnote{The three layers of a model with the largest error of per-tensor FP8 attention.} of Wan2.2,
quantizing only $Q$ and $K$ gives a larger mean error than quantizing only $P$ and $V$ in the same format, 20.14\%
against 1.90\% in FP8 and 30.12\% against 7.14\% in NVFP4, and the order is the same over all layers
(\cref{fig:motivation}a). We therefore focus on the error of~$QK$~\mbox{quantization}.

Low-precision attention methods usually quantize a high-precision tensor $x$ to a low-precision tensor
$\hat x = \Qq(x)$ with a quantization function $\Qq$, which leaves the residual $\delta x = x - \hat x$. A query $q$
and a key $k$, both row vectors, thus split into $q = \hat q + \delta q$ and $k = \hat k + \delta k$, and their exact
product decomposes as
\begin{equation}
q\cdot k = \hat q\cdot\hat k + \delta q\cdot\hat k + \hat q\cdot\delta k + \delta q\cdot\delta k .
\label{eq:qk-split}
\end{equation}
Attention computes the score of $q$ and the key $k_j$ of token $j$ as $s_j = q\cdot k_j/\sqrt d$, where $d$ is the head
dimension. A quantized kernel computes the score $\hat s_j$ from the first term of \cref{eq:qk-split} alone, so the
other three terms form the score error $e_j = \hat s_j - s_j$.

Softmax turns the scores into the attention weights $p = \operatorname{softmax}(s)$, and its exponential can amplify
the score errors. The quantized kernel computes the weights $\tilde p$ from the quantized scores
$\hat s_j = s_j + e_j$, so each weight is the exact weight $p_j$ multiplied by $\exp(e_j)$ and renormalized,
\begin{equation}
\tilde p_j = \frac{\exp(\hat s_j)}{\sum_l \exp(\hat s_l)} = \frac{p_j \exp(e_j)}{\sum_l p_l \exp(e_l)}
\approx p_j\,(1 + e_j - \bar e) ,
\label{eq:softmax-linear}
\end{equation}
where $\bar e = \sum_l p_l e_l$ is the $P$-weighted mean of the score errors. For small errors, the relative change
of each weight is close to its centered error $e_j - \bar e$, so the errors pass through softmax without amplification.
With an NVFP4 $QK$, however, the centered errors can exceed 1.0, and the linear form of \cref{eq:softmax-linear} no
longer holds. The exponential then amplifies the errors instead of passing them on linearly, and most of the attention
weight can move to a different key. This amplification explains why the attention output is more sensitive to the
quantization of $QK$ than to that of $PV$.

An NVFP4 $QK$ produces such large score errors mainly because of the floating-point format itself. A
floating-point format with $t$ mantissa bits rounds every element in its normal range with a relative error of up to
$2^{-(t+1)}$ regardless of its scale. This bound is 1/16 for FP8 and 1/4 for NVFP4. Common ways to reduce
quantization error, such as finer scales, cannot avoid this error. In addition, $Q$ and $K$ often contain outliers with
large values~\citep{massivevalues}, which, together with the exponential in softmax, further amplify these rounding
errors. \Cref{fig:motivation}a confirms the effect of the format. With the same per-block scales of SageAttention, the
floating-point FP8 has a much larger error than the integer INT8~\citep{intvsfp}, 12.63\% against about 1\%.
NVFP4 has the largest error although it has the finest scales. An accurate 4-bit $QK$ therefore has to mitigate this
error by other means.

\subsection{Principal residual compensation}
\label{sec:qk}
\label{sec:num-qk}
\label{sec:qk-why}

\deco addresses the $QK$ error numerically. Since the
residuals $\delta q$ and $\delta k$ are known, the first-order terms $\delta q\cdot\hat k$ and $\hat q\cdot\delta k$ of
\cref{eq:qk-split} can be added back to the scores. We neglect the second-order term. Adding them exactly takes
two more full-width products, more than an FP8 $QK$ costs. We therefore propose \textit{principal residual
compensation}, which approximates each term over a few channels along the principal directions that hold most of its
error, so that the $QK$ MMA computes them with the scores.

\paragraph{Error along the principal directions.} For one query, the $Q$-side term of key $j$ is $\delta q\cdot\hat k_j$. Its sum of
squares over all keys is a quadratic form with the Gram matrix $G_K$ of the quantized keys $\hat K$. For every head, we
decompose $G_K$ into eigenvalues $\lambda_1 \ge \dots \ge \lambda_d$ and orthonormal column eigenvectors $u_m$, which
splits the sum of squares along the eigenvectors,
\begin{equation}
\begin{colgather}
\sum_j (\delta q\cdot\hat k_j)^2 = \delta q\, G_K\, \delta q^\top = \sum_{m=1}^{d} \lambda_m\, (\delta q\, u_m)^2 ,
\colpair G_K = \hat K^\top \hat K = \sum_{m=1}^{d} \lambda_m u_m u_m^\top .
\end{colgather}
\label{eq:qk-energy}
\end{equation}
We call $\lambda_m = \sum_j (\hat k_j u_m)^2$ the \emph{energy} of the keys along $u_m$. \deco rotates $Q$ and $K$
by a Hadamard matrix before the quantization, a rotation widely used for low-precision
quantization~\citep{quarot,fa3,mxattention}. In most cases the rotation lowers the rounding error of $Q$ and $K$
and the second-order term. It also spreads the outliers and their residual over all channels, so that $\delta q$ has no
preferred direction and $(\delta q\, u_m)^2$ is about the same along every $u_m$. The rotation leaves the scores and the
eigenvalues unchanged, so each term $\lambda_m (\delta q\, u_m)^2$ of \cref{eq:qk-energy} is then proportional to its eigenvalue
$\lambda_m$. Without the rotation, the outliers hold most of the residual, and the eigenvalues no longer predict how much
error the compensation along their eigenvectors removes.

Compensating the error along an eigenvector $u_m$ cancels the term $\lambda_m (\delta q\, u_m)^2$ of
\cref{eq:qk-energy}. Since these terms are proportional to the eigenvalues, compensating along the eigenvectors with
the largest eigenvalues removes the most error. A few eigenvectors hold most of the energy of $K$. On Wan2.2, the
32 leading eigenvectors of $G_K$ hold a median of 74.3\% of this energy over the heads, against 46.9\% for the 32 channels of largest energy
(\cref{fig:motivation}b). Compensating along a small number $r$ of leading eigenvectors therefore recovers most of the
error.

\paragraph{Decomposing the first-order terms.} Let $B_K = [\,u_1, \dots, u_r\,]$ hold the $r$ leading eigenvectors of $G_K$. In the eigenbasis of $G_K$, the $Q$-side term of key $j$ is a sum over $d$ directions, of which \deco keeps the
first $r$,
\begin{equation}
\delta q\cdot\hat k_j = \sum_{m=1}^{d} (\delta q\, u_m)(\hat k_j\, u_m) \approx (\delta q\, B_K)\cdot(\hat k_j\, B_K) .
\label{eq:qk-truncation}
\end{equation}
The $K$-side term $\hat q\cdot\delta k_j$ splits in the same way along the $r$ leading eigenvectors $B_Q$ of $G_Q = \hat Q^\top \hat Q$.

\input{figures/compensation}

\paragraph{Extended operands.} Adding the kept parts of both first-order terms to $\hat q\cdot\hat k$ gives one dot
product $q\cdot k \approx q'\cdot k'$ of the extended operands
\begin{equation}
q' = [\hat q \,|\, \delta q B_K \,|\, \hat q B_Q], \ \
k' = [\hat k \,|\, \hat k B_K \,|\, \delta k B_Q] .
\label{eq:qk-channels}
\end{equation}
The compensation channels are quantized to NVFP4 as well. As \cref{fig:compensation} shows, one block-scaled NVFP4
MMA over the extended operands then computes the scores together with the first-order correction of both operands.
On the hard layers of Wan2.2, the compensation lowers the error of the kernel from 18.70\% to 2.80\%
(\cref{tab:abl-acc}). It brings the centered score errors from 1.0 or more down to the range where softmax is
approximately linear.

\paragraph{Efficient estimation.}
\label{sec:basis}
\deco estimates the principal directions at runtime for every head, in a preparation step that we keep lightweight in three ways. First, the directions are estimated from a random sample $\mathcal S$ of
$N/256$ of the $N$ tokens. Second, $Q$ and $K$ share one basis. Both are
linear projections of the same activations, and we find that their leading eigenvectors span largely the same
subspace. On Wan2.2, the two subspaces overlap by 0.75 (against 0.25 for two unrelated subspaces). We add the two Gram
matrices of the sample,
each divided by its trace so that $Q$ and $K$ weigh equally,
\begin{equation}
G = \frac{K_{\mathcal S}^\top K_{\mathcal S}}{\operatorname{tr}(K_{\mathcal S}^\top K_{\mathcal S})} + \frac{Q_{\mathcal S}^\top Q_{\mathcal S}}{\operatorname{tr}(Q_{\mathcal S}^\top Q_{\mathcal S})} , \qquad
B \in \mathbb{R}^{d \times r} ,
\label{eq:shared-basis}
\end{equation}
and take the $r$ leading eigenvectors of $G$ as the basis $B$ in place of both $B_K$ and $B_Q$ in
\cref{eq:qk-channels}. The shared basis holds 69.0\% of the energy of $K$, against 74.3\% for $B_K$ (\cref{fig:motivation}b), and takes
only one eigendecomposition per head. Third, the preparation runs as a fast chain of fused kernels. A GPU
eigensolver finds the bases of all heads, and one fused pass over $Q$ and $K$ rotates, quantizes, and projects every
token. These passes are linear in the sequence length, whereas attention is quadratic, and the whole preparation
adds only 2.7 to 3.6\% to the attention time.

%% file: figures/compensation.tex
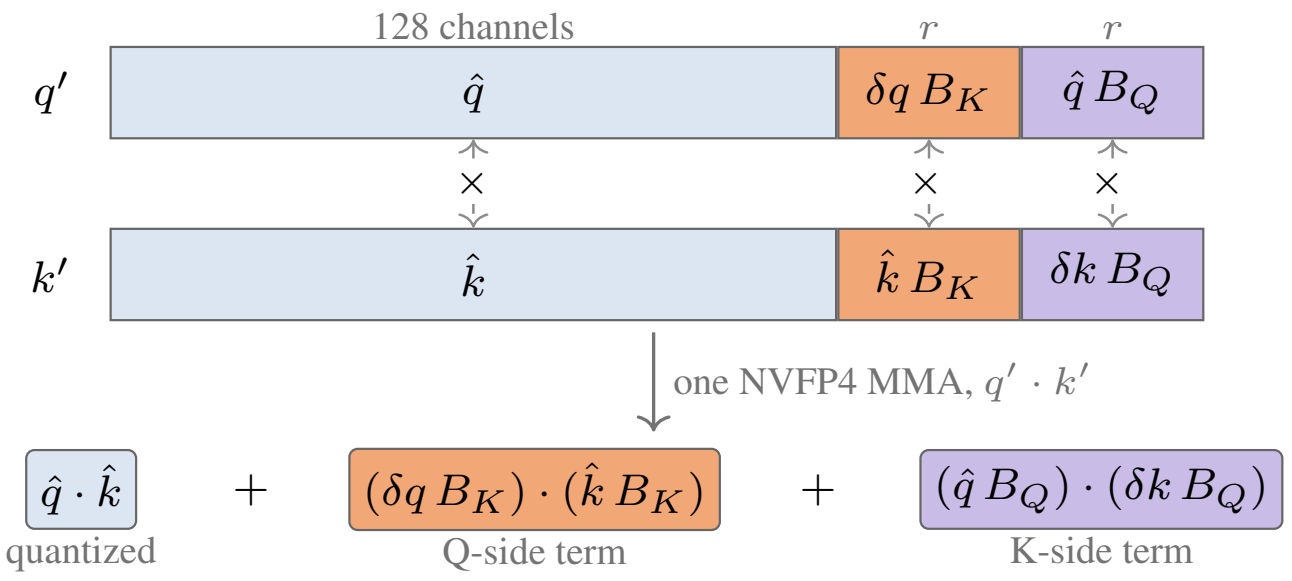
\begin{figure}[t]
\centering
\definecolor{opfill}{HTML}{DCE6F2}
\definecolor{qsidefill}{HTML}{F2A777}
\definecolor{ksidefill}{HTML}{C9BCE6}
\resizebox{\colfigwidth}{!}{%
\begin{tikzpicture}[font=\footnotesize, x=1.2cm, y=1.05cm,
  seg/.style={draw=black!60, line width=0.4pt, minimum height=0.58cm, inner sep=1pt},
  dims/.style={font=\scriptsize, text=black!55, inner sep=1pt},
  pair/.style={<->, black!45, line width=0.45pt, densely dashed},
  term/.style={draw=black!60, line width=0.4pt, rounded corners=1.5pt, inner sep=2.5pt, minimum height=0.5cm}]
\node[anchor=east] at (0.9, 3.6) {$q'$};
\node[seg, fill=opfill, minimum width=4.608cm, anchor=west] (q0) at (1.0, 3.6) {$\hat q$};
\node[seg, fill=qsidefill, minimum width=1.152cm, anchor=west] (q1) at (q0.east) {$\delta q\,B_K$};
\node[seg, fill=ksidefill, minimum width=1.152cm, anchor=west] (q2) at (q1.east) {$\hat q\,B_Q$};
\node[anchor=east] at (0.9, 2.5) {$k'$};
\node[seg, fill=opfill, minimum width=4.608cm, anchor=west] (k0) at (1.0, 2.5) {$\hat k$};
\node[seg, fill=qsidefill, minimum width=1.152cm, anchor=west] (k1) at (k0.east) {$\hat k\,B_K$};
\node[seg, fill=ksidefill, minimum width=1.152cm, anchor=west] (k2) at (k1.east) {$\delta k\,B_Q$};
\node[dims, anchor=south] at (q0.north) {128 channels};
\node[dims, anchor=south] at (q1.north) {$r$};
\node[dims, anchor=south] at (q2.north) {$r$};
\foreach \a/\b in {q0/k0, q1/k1, q2/k2}
  \draw[pair] (\a.south) -- (\b.north);
\node[font=\scriptsize, fill=white, inner sep=1pt] at (2.92, 3.05) {$\times$};
\node[font=\scriptsize, fill=white, inner sep=1pt] at (5.32, 3.05) {$\times$};
\node[font=\scriptsize, fill=white, inner sep=1pt] at (6.28, 3.05) {$\times$};
\node[term, fill=opfill] (t0) at (0.85, 1.2) {$\hat q\cdot\hat k$};
\node at (1.75, 1.2) {$+$};
\node[term, fill=qsidefill] (t1) at (3.25, 1.2) {$(\delta q\,B_K)\cdot(\hat k\,B_K)$};
\node at (4.75, 1.2) {$+$};
\node[term, fill=ksidefill] (t2) at (6.25, 1.2) {$(\hat q\,B_Q)\cdot(\delta k\,B_Q)$};
\node[dims, anchor=north] at (t0.south) {quantized};
\node[dims, anchor=north] at (t1.south) {Q-side term};
\node[dims, anchor=north] at (t2.south) {K-side term};
\draw[->, black!55, line width=0.6pt] (3.88, 2.15) -- (3.88, 1.55);
\node[dims, anchor=west] at (3.95, 1.85) {one NVFP4 MMA, $q' \cdot k'$};
\end{tikzpicture}}
\caption{\textbf{Principal residual compensation of QK.} Extra channels for the $Q$-side and the $K$-side term let one NVFP4
MMA add both terms to the quantized score.}
\label{fig:compensation}
\end{figure}

%% file: docs/performance.tex
\subsection{Kernel implementation and optimizations}
\label{sec:performance}

\begin{figure}[t]
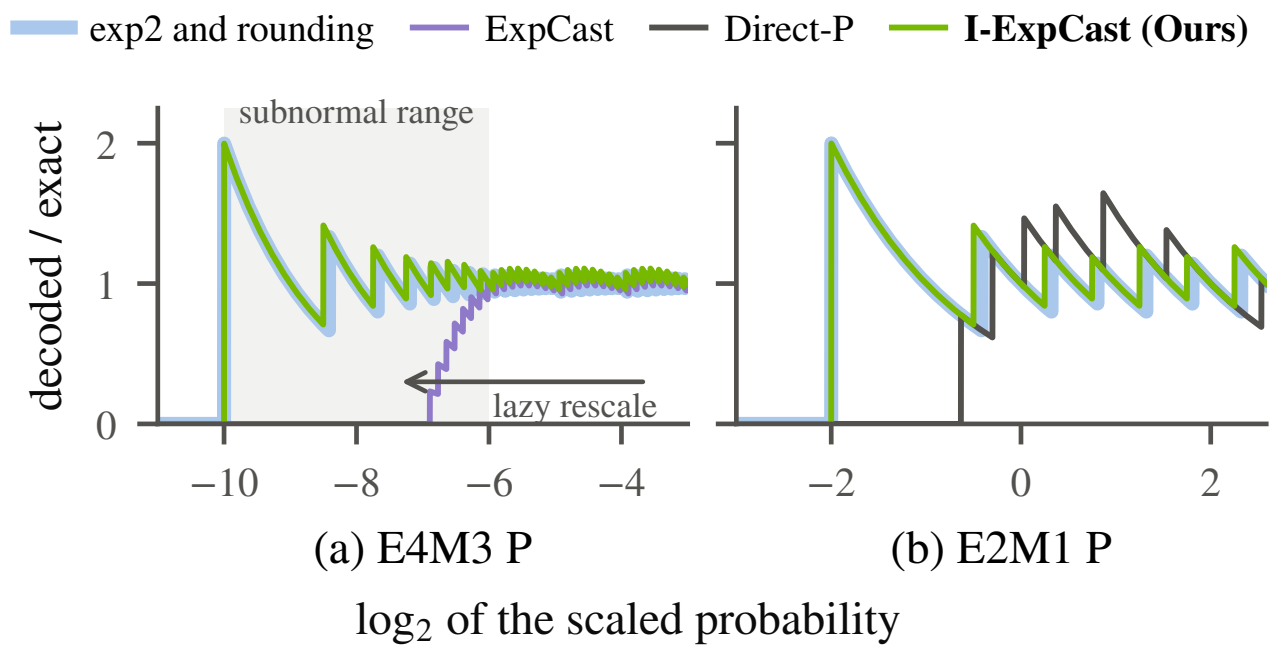

\centering
\colplotwide{plots/p_codes.pdf}
\caption{\textbf{Accuracy of I-ExpCast.} Decoded value of $P$ over its exact value, per way of forming the E4M3 codes (a) and
the~E2M1~codes~(b).}
\label{fig:p-construction}
\end{figure}

We implement the kernels of \deco in CuTe DSL for GB300, VR200, and RTX PRO 6000. On the datacenter GPUs, the softmax between the two low-precision MMAs bounds the kernel and hides their speed
advantage~\citep{vcattention,fa4}. We propose I-ExpCast, which forms $P$ with fewer instructions and without the SFU.
We also adapt the lazy rescale and the row sum in the $PV$ MMA to our kernels.

\label{sec:kernel-p} Within the softmax, the exponential is the bottleneck, since it runs on the SFU and handles far fewer scores per
cycle than the MMAs~\citep{fa4}. FlashAttention-4 emulates part of the exponentials on the FMA units, which takes
many more instructions than the SFU and lengthens the kernel on GB300 and VR200.
ExpCast~\citep{vcattention} and Direct-P~\citep{fa4fp4} bypass the SFU and form the codes of $P$ from a linear function
of the score, ExpCast in E4M3 and Direct-P in E2M1. However, neither method rounds to the nearest code, and neither
covers the subnormal codes.

\begin{table}[t]
\centering
\tablefont
\setlength{\tabcolsep}{3pt}
\caption{\textbf{Rounding error of P under the lazy rescale.} Relative L2 error (\%) on Wan2.2 with only $P$ rounded,
per threshold $T$ of the lazy rescale. Instr. counts the instructions per four scores.}
\label{tab:p-cast}
\input{docs/tables/p_cast}
\end{table}

\input{docs/floats/eval_e2e}

I-ExpCast, our improved ExpCast, forms the probability in FP16 and leaves the rounding to a hardware conversion. FP16
has ten mantissa bits and represents every value of E4M3 and E2M1 exactly, so the conversion rounds to
the nearest code of either format, subnormal codes included. Let $x$ be the base-2 logarithm of a scaled probability, a
linear function of its score $s_j$. Read as an integer, the FP16 bit pattern of $2^x$ is close to $x$ plus the exponent
bias 15, scaled by $2^{10}$ for the ten mantissa bits,
\begin{equation}
\operatorname{bits}_{\mathrm{FP16}}\big(2^x\big) \approx 1024\,(x + 15) .
\label{eq:i-expcast}
\end{equation}
I-ExpCast evaluates this linear function in the subnormal range of FP32, whose values are evenly spaced multiples
of $2^{-149}$ with the multiple as their bit pattern. One FMA therefore maps the score to this pattern and rounds the result to an
integer. Three instructions per pair of scores form the codes,
\begin{equation}
\hat p_j = \Qq\Big(\operatorname{sat}_{16}\big(\operatorname{bits}_{\mathrm{FP32}}(a\,s_j + c)\big)\Big) ,
\label{eq:i-expcast-steps}
\end{equation}
where $a\,s_j + c = 2^{-149}\cdot 1024\,(x + 15)$, the saturating pack $\operatorname{sat}_{16}$ narrows the pattern to
the 16 bits of FP16, and $\Qq$ converts it to E4M3 or E2M1. I-ExpCast thus stays close to the exponential followed by
rounding, whereas ExpCast and Direct-P do not (\cref{fig:p-construction}).

\label{sec:lazy-rescale} \label{sec:rowsum} We further adapt two existing techniques to these kernels. The lazy
rescale moves the row base only when a tile exceeds it by more than a factor of $2^T$. At $T = 8$, as in FlashAttention-4, probabilities below $2^{-9}$ of the row base round to zero, whereas $T = 4$ keeps them down to $2^{-13}$ and still
skips most rescales. At this
threshold, more of the small probabilities fall into the subnormal range of E4M3, which I-ExpCast still rounds to the
nearest code. With I-ExpCast, the two optimizations can therefore be combined. With only $P$ rounded at $T = 4$, the mean error over all layers of
Wan2.2 is 0.93\% with I-ExpCast, against 3.26\% with ExpCast, and at $T = 8$ it rises to 2.72\% (\cref{tab:p-cast}).
An NVFP4 $P$ is much less sensitive to $T$, since each block of 16 keys carries its own scale. \deco uses $T = 4$
with both formats. A column of ones appended to $V$ lets the $PV$ MMA also return the row sum of $P$~\citep{hifa4}.
\reportonly{

We also implement V smoothing~\citep{vcattention}. An extra high-precision MMA adds the block means of $V$ back
and keeps the output unbiased. More analysis is in the appendix.}

%% file: docs/tables/p_cast.tex
\begin{spreadtabular}{llccccccc}
\toprule
\spancell{2}{1}{\tablehead{P}} & \spancell{2}{1}{\tablehead{Codes from}} & \spancell{2}{1}{\tablehead{Instr.}} & \multicolumn{2}{c}{\tablehead{$T = 0$}} & \multicolumn{2}{c}{\tablehead{$T = 4$}} & \multicolumn{2}{c}{\tablehead{$T = 8$}} \\
\cmidrule(lr){4-5}\cmidrule(lr){6-7}\cmidrule(lr){8-9}
 &  &  & \tablehead{Hard} & \tablehead{All} & \tablehead{Hard} & \tablehead{All} & \tablehead{Hard} & \tablehead{All} \\
\midrule
\spancell{3}{0}{E4M3} & exp2 and rounding & 8 & 0.28 & 0.40 & 0.43 & 0.88 & 1.07 & 2.72 \\
 & ExpCast\tablecite{vcattention} & 7 & 0.44 & 1.02 & 1.41 & 3.26 & 4.37 & 8.88 \\
 & I-ExpCast & 6 & 0.50 & 0.68 & 0.54 & 0.93 & 1.11 & 2.72 \\
\midrule
\spancell{3}{0}{E2M1} & exp2 and rounding & 8 & 1.56 & 2.01 & 1.59 & 2.09 & 1.69 & 2.64 \\
 & Direct-P\tablecite{fa4fp4} & 4 & 4.11 & 5.06 & 4.11 & 5.06 & 4.11 & 5.06 \\
 & I-ExpCast & 6 & 1.51 & 1.82 & 1.54 & 1.90 & 1.63 & 2.45 \\
\bottomrule
\end{spreadtabular}

%% file: docs/floats/eval_e2e.tex
\begin{table*}[tbp]
\centering
\widetablefont
\setlength{\tabcolsep}{2pt}
\caption{\textbf{End-to-end quality on four video models against BF16 attention.} Blue cells mark the best PSNR,
SSIM, and LPIPS within each group with a 4-bit $QK$. SageAttention2 and VC-Attention use their default INT8 $QK$ on B200, and
their rows w/ FP8 $QK$ run on GB300. $^\dagger$Means reported by VC-Attention. With FP8 $PV$,
\deco reaches the generation quality of FP8-$QK$ methods.}
\label{tab:e2e}
\input{docs/tables/eval_e2e}
\end{table*}

%% file: docs/tables/eval_e2e.tex
\begin{joinedtabular}{llYYYYYYYYYYYY}
\toprule
\spancell{2}{1}{\tablehead{Method}} & \spancell{2}{1}{\tablehead{QK/PV}} & \multicolumn{3}{c}{\tablehead{Wan2.2}} & \multicolumn{3}{c}{\tablehead{LongCat-Video}} & \multicolumn{3}{c}{\tablehead{HunyuanVideo-1.5}} & \multicolumn{3}{c}{\tablehead{MiniMax-H3}} \\
\cmidrule(lr){3-5}\cmidrule(lr){6-8}\cmidrule(lr){9-11}\cmidrule(lr){12-14}
 &  & \tablehead{PSNR$\uparrow$} & \tablehead{SSIM$\uparrow$} & \tablehead{LPIPS$\downarrow$} & \tablehead{PSNR$\uparrow$} & \tablehead{SSIM$\uparrow$} & \tablehead{LPIPS$\downarrow$} & \tablehead{PSNR$\uparrow$} & \tablehead{SSIM$\uparrow$} & \tablehead{LPIPS$\downarrow$} & \tablehead{PSNR$\uparrow$} & \tablehead{SSIM$\uparrow$} & \tablehead{LPIPS$\downarrow$} \\
\midrule
BF16 & 16/16 & 24.54 & 0.850 & 0.112 & 30.09 & 0.929 & 0.038 & 18.98 & 0.709 & 0.241 & 21.40 & 0.752 & 0.186 \\
\qkpv\tablecite{tensorrtllm} & 16/8 & 20.66 & 0.748 & 0.197 & 23.50 & 0.817 & 0.107 & 16.04 & 0.601 & 0.352 & 20.19 & 0.717 & 0.217 \\
\midrule
Per-tensor FP8\tablecite{fp8formats} & \spancell{6}{0}{8/8} & 17.95 & 0.658 & 0.283 & 20.52 & 0.728 & 0.164 & 14.89 & 0.548 & 0.410 & 17.13 & 0.601 & 0.356 \\
MXFP8\tablecite{microscaling} &  & 18.01 & 0.662 & 0.278 & 20.50 & 0.732 & 0.163 & 14.06 & 0.521 & 0.451 & 16.85 & 0.590 & 0.365 \\
SageAttention2\tablecite{sage2}$^\dagger$ &  & 20.3 & 0.730 & 0.206 & 23.1 & 0.793 & 0.122 & 15.6 & 0.581 & 0.382 & 19.9 & 0.723 & 0.252 \\
\quad w/ FP8 QK &  & 17.45 & 0.646 & 0.293 & 20.81 & 0.742 & 0.158 & 14.74 & 0.551 & 0.411 & 16.98 & 0.600 & 0.355 \\
VC-Attention\tablecite{vcattention}$^\dagger$ &  & 20.5 & 0.744 & 0.191 & 23.8 & 0.807 & 0.109 & 17.5 & 0.648 & 0.301 & 20.2 & 0.726 & 0.248 \\
\quad w/ FP8 QK &  & 18.53 & 0.690 & 0.250 & 22.23 & 0.785 & 0.128 & 16.11 & 0.608 & 0.342 & 17.70 & 0.633 & 0.310 \\
\midrule
NVFP4 QK, FP8 PV\tablecite{tensorrtllm} & \spancell{3}{0}{4/8} & 15.34 & 0.553 & 0.417 & 17.85 & 0.630 & 0.258 & 11.89 & 0.430 & 0.581 & 14.47 & 0.510 & 0.484 \\
Attn-QAT\tablecite{attnqat} &  & 15.33 & 0.556 & 0.420 & 16.85 & 0.549 & 0.373 & 10.15 & 0.360 & 0.672 & 15.16 & 0.537 & 0.452 \\
\textbf{\deco (Ours)} &  & \best{19.42} & \best{0.723} & \best{0.216} & \best{22.53} & \best{0.800} & \best{0.115} & \best{15.97} & \best{0.595} & \best{0.356} & \best{17.22} & \best{0.618} & \best{0.329} \\
\midrule
NVFP4\tablecite{nvfp4pretrain} & \spancell{4}{0}{4/4} & 12.40 & 0.453 & 0.545 & 13.12 & 0.450 & 0.446 & 8.06 & 0.260 & 0.777 & 13.84 & 0.506 & 0.501 \\
SageAttention3\tablecite{sage3}$^\dagger$ &  & 13.9 & 0.510 & 0.466 & 15.1 & 0.526 & 0.382 & 10.9 & 0.391 & 0.652 & 16.1 & 0.587 & 0.408 \\
VC-Attention, 4-bit\tablecite{vcattention}$^\dagger$ &  & 16.8 & 0.614 & 0.318 & 18.7 & 0.663 & 0.226 & 13.8 & 0.486 & 0.481 & 16.6 & \best{0.605} & 0.380 \\
\textbf{\deco (Ours)} &  & \best{17.36} & \best{0.644} & \best{0.292} & \best{21.16} & \best{0.756} & \best{0.145} & \best{14.95} & \best{0.555} & \best{0.399} & \best{16.73} & 0.598 & \best{0.352} \\
\bottomrule
\end{joinedtabular}

%% file: docs/setup.tex
\subsection{Setup}
\label{sec:setup}

\input{docs/floats/eval_error_main}

\paragraph{Models.} We evaluate \deco on four video models, Wan2.2-T2V-A14B~\citep{wan} at 720p,
LongCat-Video~\citep{longcatvideo} at 480p, HunyuanVideo-1.5~\citep{hunyuanvideo15} at 720p, and
MiniMax-H3~\citep{minimaxh3} at $1344 \times 768$ with audio. For a fair comparison with the other methods, MiniMax-H3
runs with the 8-step LoRA of MiniMax-H3-Acc-LoRAs~\citep{h3lora}.

\paragraph{Baselines and metrics.} We compare \deco with three groups of methods, each in its fastest kernel. The
baseline is BF16 attention. The state-of-the-art methods are VC-Attention~\citep{vcattention} and
SageAttention2~\citep{sage2} at 8 bits, VC-Attention and SageAttention3~\citep{sage3} at 4 bits, and the kernel of
Attn-QAT~\citep{attnqat} with a 4-bit $QK$ and an 8-bit $PV$ (without retraining the model). The third group holds the
uncompensated formats: \qkpv{} with a 16-bit $QK$ and an 8-bit $PV$, per-tensor FP8, MXFP8, NVFP4 $QK$ with FP8 $PV$,
and NVFP4.
\input{figures/example}
End to end, we measure the PSNR~\citep{psnr}, SSIM~\citep{ssim}, and LPIPS~\citep{lpips} of each video against the
BF16 video of the same prompt and seed on 100 prompts of MovieGen Bench~\citep{moviegen}, and the speedup in
TensorRT-LLM VisualGen on seven of these prompts. At the kernel level, we measure the relative L2
error~\citep{higham} of the attention output against FP32 attention and~the~\mbox{attention}~time.

\paragraph{Implementations.} \label{sec:pv} \deco uses $r = 32$ directions per side as a trade-off between accuracy and
performance. By default, it runs FP8 $PV$ on GB300 and VR200 and NVFP4 $PV$ on RTX PRO 6000. Following VC-Attention~\citep{vcattention}, we smooth $V$ in both configurations by subtracting its mean
over each block of keys before quantizing it. The appendix gives further details and results.

%% file: docs/floats/eval_error_main.tex
\begin{table}[tbp]
\centering
\tablefont
\setlength{\tabcolsep}{3pt}
\caption{\textbf{Kernel-level accuracy.} Relative L2 error (\%) of the attention output. Blue marks the lowest error per
4-bit-$QK$ group.}
\label{tab:kernel-error}
\input{docs/tables/eval_error_main}
\end{table}

%% file: docs/tables/eval_error_main.tex
\begin{joinedtabular}{llYYYY}
\toprule
\spancell{2}{1}{\tablehead{Method}} & \spancell{2}{1}{\tablehead{QK/PV}} & \multicolumn{2}{c}{\tablehead{Wan2.2}} & \multicolumn{2}{c}{\tablehead{MiniMax-H3}} \\
\cmidrule(lr){3-4}\cmidrule(lr){5-6}
 &  & \tablehead{Hard} & \tablehead{All} & \tablehead{Hard} & \tablehead{All} \\
\midrule
BF16 & 16/16 & 0.15 & 0.17 & 0.17 & 0.17 \\
\midrule
SageAttention2, FP8 QK\tablecite{sage2} & \spancell{2}{0}{8/8} & 13.03 & 4.10 & 21.23 & 4.55 \\
VC-Attention, FP8 QK\tablecite{vcattention} &  & 3.67 & 2.64 & 6.65 & 2.85 \\
\midrule
NVFP4 QK, FP8 PV\tablecite{tensorrtllm} & \spancell{3}{0}{4/8} & 30.56 & 10.57 & 28.92 & 10.69 \\
Attn-QAT\tablecite{attnqat} &  & 30.19 & 10.60 & 28.25 & 11.10 \\
\textbf{\deco (Ours)} &  & \best{2.80} & \best{3.73} & \best{6.32} & \best{4.40} \\
\midrule
SageAttention3\tablecite{sage3} & \spancell{3}{0}{4/4} & 21.86 & 9.19 & 24.07 & 8.96 \\
VC-Attention, 4-bit\tablecite{vcattention} &  & 21.06 & 8.52 & 23.46 & 8.05 \\
\textbf{\deco (Ours)} &  & \best{5.68} & \best{5.09} & \best{8.24} & \best{5.67} \\
\bottomrule
\end{joinedtabular}

%% file: figures/example.tex
\begin{figure*}[tbp]
\centering
\setlength{\tabcolsep}{1pt}
\begin{tabular}{ccc}
\includegraphics[width=0.328\linewidth]{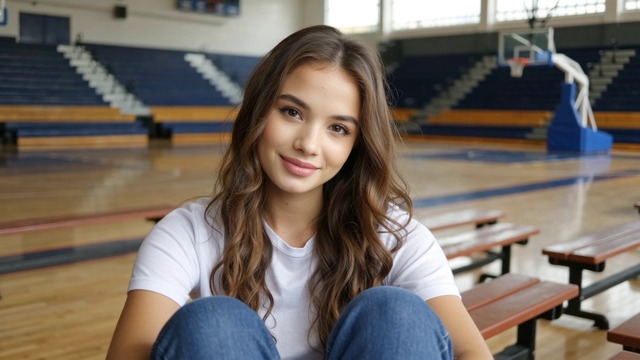} &
\includegraphics[width=0.328\linewidth]{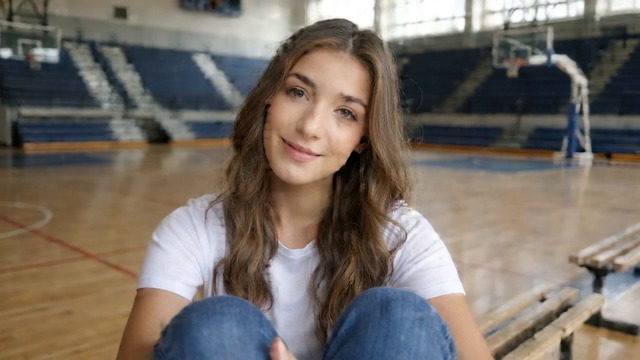} &
\includegraphics[width=0.328\linewidth]{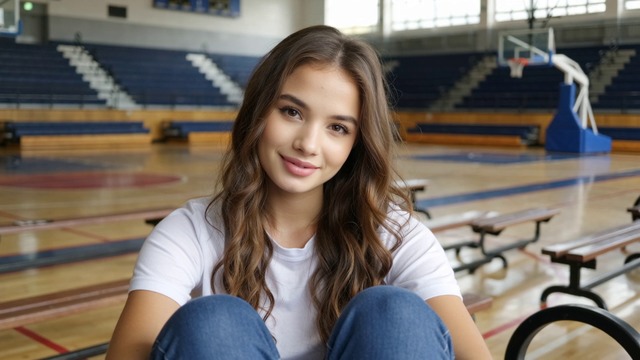} \\
\includegraphics[width=0.328\linewidth]{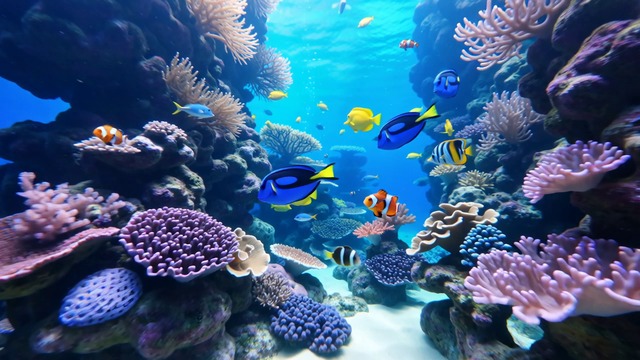} &
\includegraphics[width=0.328\linewidth]{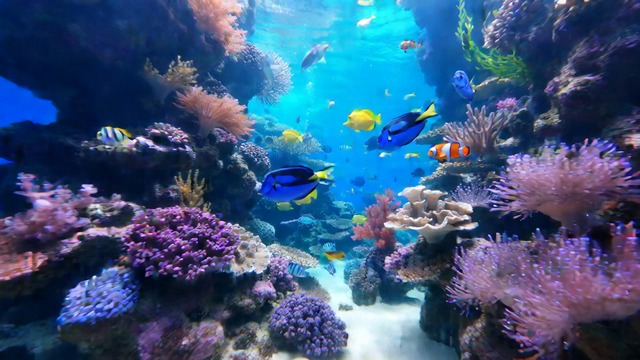} &
\includegraphics[width=0.328\linewidth]{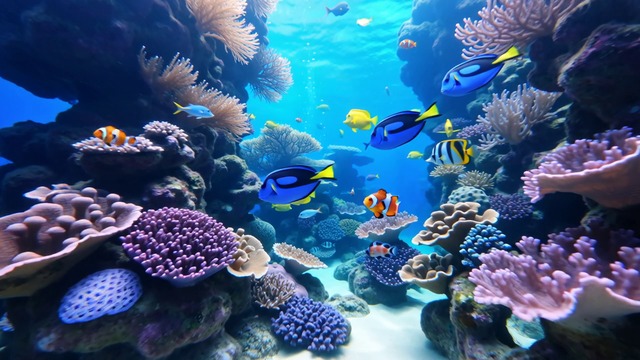} \\[1pt]
\footnotesize BF16 attention & \footnotesize NVFP4 $QK$, FP8 $PV$ & \footnotesize \deco \\
\end{tabular}
\caption{\textbf{Wan2.2 videos with an NVFP4 QK.} Frame 40 of 81 for the two prompts whose LPIPS under both
kernels is closest to the median over 100 prompts. With \deco, the frames follow those of BF16 attention and retain visual fidelity.}
\label{fig:example}
\end{figure*}

%% file: docs/results.tex
\subsection{Accuracy}
\label{sec:accuracy}
\label{sec:e2e}

\input{docs/floats/eval_e2e}
\input{figures/example}
\paragraph{End-to-end quality.} \deco reaches the generation quality of FP8-$QK$ methods and narrows the gap of a
4-bit $QK$ to INT8-$QK$ attention (\cref{tab:e2e}). On every model, it is more accurate than per-tensor FP8, MXFP8, and
SageAttention2 with FP8 $QK$. Compared with VC-Attention with FP8 $QK$, the most
accurate FP8-$QK$ method, \deco is more accurate on Wan2.2 and LongCat-Video and less on the other two models. The INT8-$QK$ methods, which VC-Attention
reports on B200, remain more accurate. The uncompensated 4-bit $QK$ is 5 to 6~dB in PSNR
behind VC-Attention with INT8 $QK$, while \deco is 1 to 3~dB behind. Its frames are also closer to those of BF16 attention
than those of the uncompensated kernel (\cref{fig:example}).

With NVFP4 $PV$, \deco has better PSNR and LPIPS than SageAttention3 and 4-bit VC-Attention on every model, and better
SSIM except against 4-bit VC-Attention on MiniMax-H3. On Wan2.2, its PSNR is 17.36~dB, against 13.9~dB for
SageAttention3 and 16.8~dB for 4-bit VC-Attention. The fully 4-bit \deco matches SageAttention2 with FP8 $QK$ on every
model and is within about 1~dB of VC-Attention with FP8 $QK$.

The fidelity metrics also capture the drift of the sampling trajectory, which exists even between BF16 kernels.
The BF16 row is the BF16 attention of cuDNN, which reaches 24.54~dB on Wan2.2. \deco is 3 to 8~dB below this row and within 3~dB of \qkpv.

\input{docs/floats/eval_error_main}
\paragraph{Kernel-level accuracy.} We evaluate the kernels on attention calls captured from Wan2.2 and MiniMax-H3 in
the hard layers and in every layer. \Cref{tab:kernel-error} shows that \deco reaches the accuracy of
FP8-$QK$ kernels. On the hard layers of both models, it has the lowest error of all kernels with an FP8 or 4-bit $QK$. On Wan2.2, it reaches 2.80\%, against 3.67\% for VC-Attention with FP8 $QK$ and 30.56\% for the uncompensated
kernel. Over all layers, its error lies between those of VC-Attention and SageAttention2 with FP8 $QK$ on both
models, 3.73\% on Wan2.2 against 2.64\% and 4.10\%. With NVFP4 $PV$, \deco has the lowest error of the fully 4-bit kernels. On the hard layers of Wan2.2, its 5.68\% is about a quarter of the 21.86\% of SageAttention3 and the
21.06\% of 4-bit VC-Attention.\reportonly{ It is also below the 13.03\% of SageAttention2 with FP8 $QK$.
Over all layers, its 5.09\% is above the 4.10\% of SageAttention2 and the 2.64\% of VC-Attention with FP8 $QK$.}

\subsection{Performance}
\label{sec:perf}
\label{sec:speed-accuracy}
\label{sec:trtllm}

\begin{figure*}[tbp]
\centering
\includegraphics[width=\linewidth]{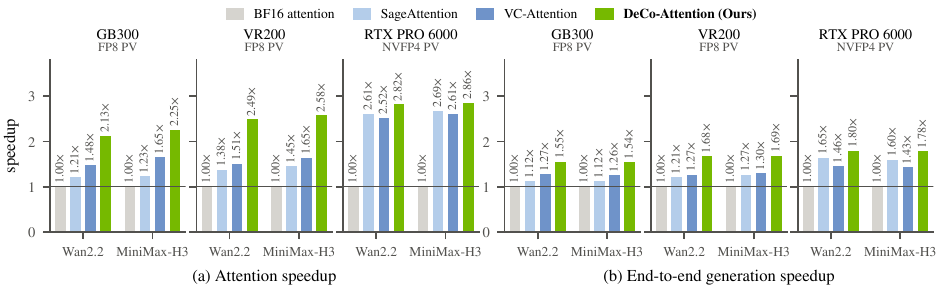}
\caption{\textbf{Kernel-level and end-to-end speedup.} (a) Speedup of the attention kernel over the fastest BF16
kernel. (b) Speedup of the whole generation in TensorRT-LLM VisualGen over BF16 attention. The SageAttention and
VC-Attention bars show their FP8-$QK$ kernels on GB300 and VR200 and their 4-bit kernels on RTX PRO 6000.}
\label{fig:kernel-time}
\end{figure*}

\Cref{fig:kernel-time} shows that \deco is the fastest of the compared attention kernels and gives the largest
end-to-end speedup on every GPU. On Wan2.2, its kernel runs 2.13$\times$, 2.49$\times$, and 2.82$\times$ faster than
BF16 attention on GB300, VR200, and RTX PRO 6000, excluding the preparation step of \cref{sec:basis}, and the whole
generation runs 1.55$\times$, 1.68$\times$, and 1.80$\times$ faster. On GB300 and VR200, \deco is 1.44$\times$ and
1.65$\times$ faster than VC-Attention, the fastest prior method, at the kernel level and 1.22$\times$ and 1.32$\times$ faster
end to end. On RTX PRO 6000, where SageAttention3 is the fastest prior method, \deco is 1.08$\times$ faster at the kernel level and 1.09$\times$ faster end to end. On MiniMax-H3, the speedups are similar, 2.25$\times$ to 2.86$\times$ at the kernel level and 1.54$\times$ to 1.78$\times$ end to end.

\subsection{Ablations}
\label{sec:ablation}

\begin{table}[tbp]
\centering
\tablefont
\caption{\textbf{Accuracy ablation.} Relative L2 error (\%) of \deco with FP8 $PV$ without the $Q$-side term, the
$K$-side term, or both, in emulation.}
\label{tab:abl-acc}
\input{docs/tables/ablation_accuracy}
\end{table}

\paragraph{Accuracy.} \Cref{tab:abl-acc} ablates the two terms of the compensation in an emulation of the GB300 kernel.
The baseline without both terms keeps the Hadamard rotation and the smoothing of $V$. The two terms
together reduce the mean error on the hard layers from 18.70\% to 2.80\% on Wan2.2 and from 29.40\% to 6.31\% on
MiniMax-H3. Most of this reduction comes from the $K$-side term, without which the error on Wan2.2 rises to 18.35\%. The
$Q$-side term has a smaller effect and still reduces the error from 3.97\% to 2.80\% on Wan2.2 and from 7.30\% to
6.31\% on MiniMax-H3.

\begin{table}[tbp]
\centering
\tablefont
\caption{\textbf{Performance ablation.} The optimizations of \cref{sec:performance} added one at a time to the \deco FP8-$PV$
kernel on GB300, with the time of a Wan2.2 call, the SM clock under the power limit, and the error on the hard layers.
The reference rows use an uncompensated FP8 $QK$ with one scale per tensor.}
\label{tab:abl-perf}
\input{docs/tables/ablation_perf}
\end{table}

\paragraph{Performance.} \label{sec:compute} \Cref{tab:abl-perf} adds the optimizations of \cref{sec:performance} one
at a time to our GB300 kernel and compares it with an FP8-$QK$ reference, the same kernel with an
uncompensated FP8 $QK$. Without the optimizations, the softmax bounds the kernel and hides the gain of
the 4-bit $QK$. Our kernel takes 79.0~ms, only slightly below the 83.8~ms of the NVFP4 $QK$ with FP8 $PV$ kernel of
TensorRT-LLM and the 82.1~ms of the reference. I-ExpCast shortens the kernel to
62.2~ms. The row sum in the $PV$ MMA and the lazy rescale shorten it to 55.8~ms, where the MMAs nearly bound
the kernel. With the smoothing of $V$, the complete kernel takes 58.0~ms, 1.36$\times$ faster than without the optimizations and 2.14$\times$ faster than
BF16 attention. It does more work per cycle and draws more power, which lowers the SM clock under the power limit from 2017 to
1582~MHz and offsets part of the gain. At a fixed clock, the same changes give 1.67$\times$ instead of 1.36$\times$. The same optimizations speed up the FP8-$QK$ reference by only 1.20$\times$, since its slower FP8 MMAs then bound
the kernel. As a result, \deco is 1.17$\times$ faster than the FP8-$QK$ reference, and achieves better accuracy on the
hard layers, 2.80\% against 3.60\%. The speedup of \deco thus comes from the 4-bit $QK$ MMA and the softmax optimizations together.

%% file: docs/tables/ablation_accuracy.tex
\begin{spreadtabular}{cccccc}
\toprule
\spancell{2}{1}{\tablehead{\makecell{Q-side\\term}}} & \spancell{2}{1}{\tablehead{\makecell{K-side\\term}}} & \multicolumn{2}{c}{\tablehead{Wan2.2}} & \multicolumn{2}{c}{\tablehead{MiniMax-H3}} \\
\cmidrule(lr){3-4}\cmidrule(lr){5-6}
 &  & \tablehead{Hard} & \tablehead{All} & \tablehead{Hard} & \tablehead{All} \\
\midrule
\checkmark & \checkmark & 2.80 & 3.73 & 6.31 & 4.40 \\
 & \checkmark & 3.97 & 4.96 & 7.30 & 5.71 \\
\checkmark &  & 18.35 & 7.27 & 29.11 & 9.22 \\
 &  & 18.70 & 8.10 & 29.40 & 10.00 \\
\bottomrule
\end{spreadtabular}

%% file: docs/tables/ablation_perf.tex
\begin{spreadtabular}{lcccc}
\toprule
\tablehead{Kernel} & \tablehead{\makecell{Time\\(ms)}} & \tablehead{Speedup} & \tablehead{\makecell{SM clock\\(MHz)}} & \tablehead{\makecell{Error\\(\%)}} \\
\midrule
BF16 attention & 124.3 & 1.00$\times$ & 1537 & 0.15 \\
NVFP4 QK, FP8 PV\tablecite{tensorrtllm} & 83.8 & 1.48$\times$ & 2070 & 30.56 \\
Our kernel, compensation only & 79.0 & 1.57$\times$ & 2017 & 2.92 \\
\quad + I-ExpCast & 62.2 & 2.00$\times$ & 1777 & 2.94 \\
\quad + Row sum in the PV MMA & 57.3 & 2.17$\times$ & 1612 & 2.94 \\
\quad + Lazy rescale, $T = 4$ & 55.8 & 2.23$\times$ & 1537 & 2.95 \\
\quad + Smoothing of V & 58.0 & 2.14$\times$ & 1582 & 2.80 \\
\midrule
FP8-QK reference & 82.1 & 1.51$\times$ & 1912 & 3.70 \\
\quad + All four above & 68.2 & 1.82$\times$ & 1537 & 3.60 \\
\bottomrule
\end{spreadtabular}

%% file: docs/appendix.tex
\input{docs/numerics}
\input{docs/compensation}
\input{docs/pv}
\input{docs/kernel_details}

\section{Evaluation Details}
\label{sec:eval-details}

This appendix supplements \cref{sec:evaluation}. We give the setup (\cref{sec:eval-setup}), further accuracy
results (\cref{sec:eval-accuracy}), and the times behind the speedups (\cref{sec:eval-performance}).

\subsection{Setup}
\label{sec:eval-setup}

\begin{table}[tbp]
\centering
\tablefont
\caption{\textbf{Models and captures of the kernel-level evaluation.} Output of each model, size of a self-attention
call, and the captured calls.}
\label{tab:captures}
\begin{spreadtabular}{lcc}
\toprule
 & \tablehead{Wan2.2} & \tablehead{MiniMax-H3} \\
\midrule
Resolution & 720p & $1344 \times 768$ \\
Frames & 81 & 243, with audio \\
Tokens per call & 75,600 & 73,512 \\
Heads & 40 & 56 \\
Hard layers & 33, 34, 38 & 49, 45, 1 \\
Hard calls & 18 & 9 \\
All-layer calls & 80 & 50 \\
\bottomrule
\end{spreadtabular}
\end{table}

\begin{table}[tbp]
\centering
\tablefont
\setlength{\tabcolsep}{3pt}
\caption{\textbf{Sources of the implementations.} Where the implementation of each method comes from.}
\label{tab:sources}
\begin{spreadtabular}{lcl}
\toprule
\tablehead{Method} & \tablehead{QK/PV} & \tablehead{Source} \\
\midrule
BF16, FlashAttention-4 & 16/16 & public release \\
BF16, cuDNN & 16/16 & TensorRT-LLM \\
\qkpv & 16/8 & TensorRT-LLM \\
\midrule
Per-tensor FP8 & 8/8 & TensorRT-LLM \\
MXFP8 & 8/8 & TensorRT-LLM \\
SageAttention2, INT8 or FP8 QK & 8/8 & TensorRT-LLM \\
VC-Attention, INT8 or FP8 QK & 8/8 & TensorRT-LLM \\
\midrule
NVFP4 QK, FP8 PV & 4/8 & TensorRT-LLM \\
Attn-QAT & 4/8 & public release \\
\midrule
NVFP4 & 4/4 & TensorRT-LLM \\
SageAttention3 & 4/4 & public release \\
VC-Attention, 4-bit & 4/4 & our reproduction \\
Direct-P & 4/4 & public release \\
\bottomrule
\end{spreadtabular}
\end{table}

We give the models and captures, the software, the settings of every method, and the protocols of the four
measurements, the end-to-end quality, the kernel error, the kernel time and the end-to-end time.

\paragraph{Models and captures.} \Cref{tab:captures} lists the models of the kernel-level evaluation. We capture the
BF16 inputs $Q$, $K$, and $V$ of their self-attention calls, on Wan2.2 those of the conditional guidance branch. The
hard layers are the three layers with the largest error of per-tensor FP8 attention. On Wan2.2, each hard layer is
captured at three steps per expert, and the all-layer calls take every layer at the middle step of each expert, steps
6 and 26. MiniMax-H3\footnote{The public MiniMax-H3 checkpoint of MiniMaxAI, with the PDD LoRA MiniMax-H3-Acc-LoRAs of
Alibaba PAI.} runs with its PDD LoRA as in VC-Attention, which reduces the sampling to eight forward passes. The base
model is guidance-distilled and runs without classifier-free guidance. Its calls hold text, video, and audio tokens in
one joint attention. The hard layers are captured at passes 0, 4, and 7, and the all-layer calls take all 50 layers
at pass 4.

\paragraph{Software.} Every kernel that we run uses PyTorch 2.14 and CUDA 13.4. Our kernels and FlashAttention-4
4.0.0b37 are built with CuTe DSL 4.8.1, and the other baselines with the versions of their releases. cuDNN is 9.25 on
GB300 and 9.26 on VR200 and RTX PRO 6000. On RTX PRO 6000, the kernels run in the TensorRT-LLM 1.4.0rc2 container.

\input{docs/floats/eval_e2e_vbench}
\input{docs/floats/eval_e2e_vbench_motion}

\paragraph{Method settings.} Every baseline keeps the quantization scheme of its method, and where a method has
several kernels we time the fastest. \Cref{tab:sources} lists where each implementation comes from.
\begin{itemize}[leftmargin=*,itemsep=2pt]
  \item \emph{BF16 attention.} The reference videos of the end-to-end quality come from FlashAttention-4.
    The BF16 row of \cref{tab:e2e} is the BF16 attention of cuDNN, scored against them, and the VBench tables hold
    both.
  \item \emph{\qkpv, per-tensor FP8, MXFP8, NVFP4 $QK$ with FP8 $PV$, and NVFP4} are the configurations of
    TensorRT-LLM VisualGen with these formats. NVFP4 $QK$ with FP8 $PV$ subtracts the mean of $K$ and applies
    SmoothQuant scales~\citep{smoothquant} before the quantization.
  \item \emph{SageAttention} has three configurations. SageAttention2 scales $Q$ per token and $K$ per 16 tokens,
    with INT8 or FP8 $QK$, and SageAttention3 is fully 4-bit.
  \item \emph{VC-Attention} groups the video tokens of a call by k-means and smooths $V$ per tile of 128 tokens. As
    for SageAttention, we run its configurations with INT8 $QK$, with FP8 $QK$, and fully 4-bit.
  \item \emph{Attn-QAT} runs without the retraining and casts $V$ to E4M3 without a scale. A prompt whose output is
    not finite is generated again.
  \item \emph{Reported rows.} The rows of \cref{tab:e2e,tab:e2e-vbench} marked with a dagger are means that
    VC-Attention reports, quoted unchanged. They are SageAttention2 and VC-Attention with INT8
    $QK$ on B200, since their INT8 $QK$ needs the INT8 tensor cores of B200, and SageAttention3 and 4-bit VC-Attention
    on RTX PRO 6000. Our own runs of the two INT8 configurations on B200 reproduce the reported PSNR to within about 1.4~dB
    (\cref{tab:reported-int8}).
    That of SageAttention2 differs by about 0.5~dB at most. That of VC-Attention is within 0.1~dB on HunyuanVideo-1.5, 0.7
    and 0.8~dB lower on Wan2.2 and MiniMax-H3, and 1.4~dB higher on LongCat-Video. The tables quote the reported
    values.
\end{itemize}

\input{docs/floats/reported_int8}

\paragraph{End-to-end quality.} Each model generates 100 videos for the first 100 prompts of the extended prompt file
of MovieGen Bench, with the prompt index as seed and the same initial noise for every method. The kernel under test
computes every self-attention call of every layer and step. Each model runs with its official default settings, except
that LongCat-Video runs only its first stage at 480p and MiniMax-H3 runs the distilled pipeline of the captures, as in
VC-Attention. LPIPS uses AlexNet features. VBench~\citep{vbench} scores each video without a
reference in six dimensions. Differences between two runs are paired per prompt and given with the 95\% Student-t confidence interval of their
mean.

\input{docs/floats/eval_e2e_paired}
\input{docs/floats/eval_e2e_paired_pv4}
\input{docs/floats/eval_error_full}
\input{docs/floats/eval_time_full}

\paragraph{Kernel error.} We measure the relative L2 error of the attention output against the FP32 attention of the
BF16 inputs,
\begin{equation}
\varepsilon = \bigg(\frac{\sum_h \lVert \hat O_h - O_h\rVert_F^2}{\sum_h \lVert O_h\rVert_F^2}\bigg)^{1/2},
\label{eq:metric}
\end{equation}
where $O_h$ and $\hat O_h$ are the reference and the output under test of head $h$ on the same 1024 query rows, 16
random blocks of 64 rows. We report the mean over the hard layers and over all layers. The errors come from the GB300
kernels, except those of the INT8 kernels, which run on B200, and of SageAttention3 and 4-bit VC-Attention, which run
on RTX PRO 6000.

\paragraph{Emulation.} \label{sec:abl-emulation} The accuracy ablations emulate the GB300 kernels of \deco in PyTorch
and Triton. The emulation forms $Q'$, $K'$, and the smoothed $V$ from the same sampled tokens as the preparation, runs
the online softmax at $T = 4$ with I-ExpCast, and takes the sums of $P$ as the kernels do. It computes the basis in
FP64 and the rescale factors of the online softmax and the final division exactly, whereas the kernels use their own
eigensolver and approximate instructions. For the settings that also exist as kernel builds, the emulation reproduces
the call error of the kernel to within 0.25\% of its value.

\paragraph{Kernel time.} We time the attention kernel alone, at the default clock of the GPU and at the batch of the
real call, on four captures per model, two of hard layers and two of other layers. Each kernel runs after a warm-up, in
two passes with the methods in opposite order. We report the median over the captures and the passes. The speedup is
over the fastest BF16 kernel of the same session. A kernel that runs only together with the quantization of its
operands is timed in the profiler. The preparation of a method is its work outside the kernel, the quantization of the
operands and, where the method has them, the smoothing, rotation, compensation, and grouping. We time it in the same
way, and \cref{tab:time-full} gives its median. For VC-Attention, this is the time of a step that reuses its
grouping. A step that groups again with k-means takes about an order of magnitude longer. The 1.67$\times$ of
\cref{sec:ablation} comes from a separate measurement with the SM clock fixed at 1500~MHz, in which the kernel takes
101.1~ms without the four steps of \cref{tab:abl-perf} and 60.5~ms with them (\cref{tab:abl-fixed-clock}).
\Cref{tab:abl-perf} itself is at the default clock.

\input{docs/floats/ablation_fixed_clock}

\paragraph{End-to-end time.} \label{sec:trtllm-details} The time of a generation is the time that TensorRT-LLM
VisualGen reports for a request, from the prompt to the decoded video, as the mean over prompts 1--7 after a warm-up
generation. The methods of a model run one after another on the same GPU or host. The BF16 run uses the BF16
attention of cuDNN on GB300 and VR200 and the FA2 prefill of FlashInfer on RTX PRO 6000.

\input{docs/floats/eval_e2e_vbench}
\input{docs/floats/eval_e2e_vbench_motion}
\subsection{Accuracy}
\label{sec:eval-accuracy}
\label{sec:kernel-accuracy}

We show that \deco keeps the VBench scores of BF16 attention, that the paired differences per prompt give the ranking
of \cref{sec:evaluation}, and that it has the lowest worst-case kernel error among the kernels with a 4-bit $QK$.

\paragraph{Quality without a reference.} \deco keeps the VBench scores of BF16 attention
(\cref{tab:e2e-vbench,tab:e2e-vbench-motion}). Among the runs with a 4-bit $QK$, it has the best score in 19 of the
24 cells of \cref{tab:e2e-vbench} and in 13 of the 16 ranked cells of \cref{tab:e2e-vbench-motion}. Its largest
drops in \cref{tab:e2e-vbench} are from 0.905 to 0.902 (S.C. on MiniMax-H3, FP8 $PV$)
and from 0.659 to 0.656 (A.Q. on HunyuanVideo-1.5, NVFP4 $PV$). The three kernels with an uncompensated NVFP4 $QK$
lose much more. On HunyuanVideo-1.5, their aesthetic quality is 0.594, 0.591, and 0.513, and their motion smoothness
is 0.9812, 0.9787, and 0.9700, against 0.9893 for the reference and 0.9888 and 0.9881 for \deco. The BF16 attention
of cuDNN shows that the fidelity metrics also measure the drift of the sampling trajectory. Its PSNR against the
reference is only 24.54~dB on Wan2.2, yet its VBench scores are 0.931, 0.700, and 0.668 against 0.930, 0.700, and
0.669.

\paragraph{Paired differences.} \Cref{tab:e2e-paired,tab:e2e-paired-pv4} compare each method with \deco prompt by
prompt, with FP8 $PV$ and with NVFP4 $PV$. A difference is significant when its 95\% confidence interval excludes
zero. Grey entries are significantly worse than \deco, bold entries significantly better, and black entries are not
significant. Every kernel with an uncompensated 4-bit $QK$ is significantly worse than \deco on every model. With
FP8 $PV$, \deco has a significantly lower LPIPS than per-tensor FP8, MXFP8, and SageAttention2 with FP8 $QK$ on
every model. Against VC-Attention with FP8 $QK$, one difference is significant in each direction, in favor of \deco
on Wan2.2 and of VC-Attention on MiniMax-H3. With NVFP4 $PV$, no difference to SageAttention2 with FP8 $QK$ is
significant on any model, and the PSNR of \deco is 0.97 to 1.17~dB below that of VC-Attention with FP8 $QK$. Its
PSNR is 1.95 to 6.15~dB above that of SageAttention3 on every model, all significant, and above that of 4-bit
VC-Attention on every model, significant on LongCat-Video and MiniMax-H3 with 3.67 and 1.22~dB.

\input{docs/floats/eval_e2e_paired}
\input{docs/floats/eval_e2e_paired_pv4}
\paragraph{Every kernel.} \Cref{tab:kernel-error-full} extends \cref{tab:kernel-error} to every kernel and adds the
maximum over the calls. \deco has the lowest mean and maximum among the kernels with a 4-bit $QK$. With FP8 $PV$,
its worst call on the hard layers of Wan2.2 has 3.74\%, against 47.44\% for NVFP4 $QK$ with FP8 $PV$. With NVFP4
$PV$, it has 6.79\%, against 36.02\% for SageAttention3 and 35.67\% for 4-bit VC-Attention. Against the kernels
with an FP8 $QK$, the fully 4-bit \deco has a lower mean error than SageAttention2 on the hard layers, 5.68\% against
13.03\% on Wan2.2 and 8.24\% against 21.23\% on MiniMax-H3, and a higher one over all layers, 5.09\% against 4.10\% and
5.67\% against 4.55\%. VC-Attention with FP8 $QK$ keeps the lowest mean error over all layers, 2.64\% and 2.85\%, below
both configurations of \deco.

\input{docs/floats/eval_error_full}
\input{docs/floats/eval_time_rtx}

\subsection{Performance}
\label{sec:eval-performance}

We show that \deco has the shortest attention time and the shortest generation on every GPU and model.

\paragraph{Kernel speed.} \Cref{tab:time-full,tab:time-rtx} give the times behind \cref{fig:kernel-time}. With FP8
$PV$, a Wan2.2 call takes 57.5~ms on GB300 and 29.2~ms on VR200. The fastest baseline is the fully 4-bit NVFP4 kernel
with 74.8~ms on GB300 and per-tensor FP8 with 43.7~ms on VR200. With NVFP4 $PV$ on RTX PRO 6000, the call takes
234.0~ms, against 253.0~ms for SageAttention3 and 262.1~ms for 4-bit VC-Attention.

\input{docs/floats/trtllm_e2e}

\paragraph{End-to-end speed.} \Cref{tab:trtllm-e2e} gives the generation times. With FP8 $PV$, a Wan2.2 generation takes 194.7~s on GB300 and 108.0~s on VR200, 1.55$\times$ and 1.68$\times$ faster than
with BF16 attention, against 237.8~s and 142.9~s for VC-Attention with FP8 $QK$. On RTX PRO 6000, it takes 900.8~s
with NVFP4 $PV$, 1.80$\times$ faster, against 983.9~s for SageAttention3. The BF16 run of Wan2.2 on GB300 takes 302.7~s, and its attention takes 67.6\% of a traced
denoising step, the two numbers of the introduction.

\input{docs/floats/eval_time_full}
\input{docs/floats/eval_time_rtx}
\input{docs/floats/trtllm_e2e}
\section{Generated Videos}
\label{sec:videos}

\input{results/appendix/videos}
\input{results/appendix/prompts}

\Cref{fig:videos-wan,fig:videos-longcat,fig:videos-hunyuan,fig:videos-minimax} show the videos that each model
generates from one prompt, five frames per row. The
upper block of each figure holds BF16 FlashAttention-4 and three GB300 runs of \cref{tab:e2e}. The lower block holds
our runs of the fully 4-bit methods on RTX PRO 6000 next to BF16 attention on the same GPU, the FA2 prefill of
FlashInfer, with the protocol and seeds of the GB300 runs. In these runs, 4-bit VC-Attention groups the tokens of each call again on every denoising step. The PSNR above a row
is that of the whole clip against the BF16 row of its block. The four prompts of MovieGen Bench follow.
{\small
\begin{itemize}[leftmargin=*,itemsep=1pt]
  \item \textbf{Wan2.2-T2V-A14B.} Prompt \pgfkeysvalueof{/appendix/wan2.2-t2v-a14b/prompt}: \emph{``\pgfkeysvalueof{/appendix/wan2.2-t2v-a14b/fulltext}''}
  \item \textbf{LongCat-Video.} Prompt \pgfkeysvalueof{/appendix/longcat-video/prompt}: \emph{``\pgfkeysvalueof{/appendix/longcat-video/fulltext}''}
  \item \textbf{HunyuanVideo-1.5.} Prompt \pgfkeysvalueof{/appendix/hunyuanvideo-1.5/prompt}: \emph{``\pgfkeysvalueof{/appendix/hunyuanvideo-1.5/fulltext}''}
  \item \textbf{MiniMax-H3.} Prompt \pgfkeysvalueof{/appendix/minimax-h3/prompt}: \emph{``\pgfkeysvalueof{/appendix/minimax-h3/fulltext}''}
\end{itemize}}

\input{figures/appendix_videos}

%% file: docs/numerics.tex
\section{Error of QK Quantization}
\label{sec:app-qk}
\label{sec:derivations}

This appendix supplements \cref{sec:motivation} with the bound on the score error of a floating-point $QK$
(\cref{sec:der-bound}) and the propagation of the score errors through softmax to the output
(\cref{sec:der-softmax}).

\paragraph{Notation.} A query $q$ and the keys $k_j$ are row vectors of $d$ channels, the inputs of the attention. A quantized kernel
rounds them to the low-precision format, $\hat q = \Qq(q)$ and $\hat k_j = \Qq(k_j)$, which leaves the residuals
$\delta q = q - \hat q$ and $\delta k_j = k_j - \hat k_j$, and computes the score $\hat s_j = \hat q\cdot\hat k_j/\sqrt d$ in
place of the exact score $s_j = q\cdot k_j/\sqrt d$. The exact attention weights are $p_j = \exp(s_j)/\sum_l \exp(s_l)$,
and $\tilde p_j$ are the weights of the quantized scores. The captures, the hard layers, and the error metric are those
described in \cref{sec:eval-setup}.

\subsection{Score error of a floating-point QK}
\label{sec:der-bound}

A floating-point format rounds with a relative error that a scale does not lower, and the score errors of an NVFP4
$QK$ can exceed 1.0, beyond which the exponential of softmax amplifies them (\cref{sec:motivation}). We derive the
rounding bound behind the first statement and the bound on the score error behind the second.

\paragraph{Rounding of one element.} An integer format and a floating-point format space their codes differently. INT8
gives a group of elements one scale, which maps the largest magnitude $a$ of the group to the largest code 127. The codes are then $a/127$ apart for every element of the group, with a rounding error of at most half of this spacing. A floating-point format with $t$ mantissa bits places $2^t$ codes in every binade $[2^\ell, 2^{\ell+1})$. The
codes of this binade are $2^{\ell-t}$ apart. Since an element $x$ of the binade has $\lvert x \rvert \ge 2^\ell$, this spacing is at most $2^{-t}\,\lvert x \rvert$, with a rounding error of at most half of it. The rounding errors
of the two formats are thus bounded by
\begin{equation}
\begin{colgather}
\lvert \delta x \rvert \le \frac{a}{254} \quad \text{(INT8)}, \colpair
\lvert \delta x \rvert \le 2^{-(t+1)}\, \lvert x \rvert \quad \text{(floating point)} .
\end{colgather}
\label{eq:rounding-bounds}
\end{equation}
The INT8 bound is absolute and follows the largest element of the group. The floating-point bound is relative to the
element itself, $1/16$ for E4M3 and $1/4$ for E2M1. A scale multiplies an element before the rounding and divides it
afterwards. For INT8, a scale fitted to a smaller group lowers $a$ and thus the error of every element of the group.
For a floating-point format, a scale does not lower the error. The scaled element is rounded with the same relative
error, and dividing by the scale afterwards leaves this relative error unchanged. The relative bound therefore holds
for any scale that keeps the element in its normal range.
With the same per-block scales, E4M3 errs by up to $\lvert x \rvert / 16$ against $a/254$ for INT8, which is larger for
every element above about $a/16$. FP8 therefore has a much larger error than INT8 in \cref{fig:motivation}a.

\paragraph{Score error.} The score error consists of the three terms of the decomposition of the exact product that a quantized kernel leaves
out (\cref{sec:motivation}, \cref{eq:qk-split}),
\begin{equation}
e_j = \hat s_j - s_j = -\big(\underbrace{\delta q\cdot\hat k_j}_{\text{$Q$-side term}} + \underbrace{\hat q\cdot\delta k_j}_{\text{$K$-side term}} + \underbrace{\delta q\cdot\delta k_j}_{\text{second order}}\big)/\sqrt d .
\label{eq:score-error}
\end{equation}
Each first-order term sums a residual times an element of the other operand over the $d$ channels. The triangle
inequality and the floating-point bound on every residual, $\lvert \delta q_c \rvert \le 2^{-(t+1)} \lvert q_c \rvert$, bound the
$Q$-side term by
\begin{equation}
\begin{colgather}
\lvert \delta q\cdot\hat k_j \rvert = \Big\lvert \sum_{c=1}^{d} \delta q_c\, \hat k_{jc} \Big\rvert
\le \sum_{c=1}^{d} \lvert \delta q_c \rvert\, \lvert \hat k_{jc} \rvert \colpair
\le 2^{-(t+1)} \sum_{c=1}^{d} \lvert q_c \rvert\, \lvert \hat k_{jc} \rvert
\approx 2^{-(t+1)} \sum_{c=1}^{d} \lvert q_c\, k_{jc} \rvert ,
\end{colgather}
\label{eq:term-bound}
\end{equation}
where the last step replaces the quantized element $\hat k_{jc}$ by the exact one, which holds to first order. The
$K$-side term has the same bound with the roles of $q$ and $k_j$ exchanged. Adding the two bounds and dividing by
$\sqrt d$ as in \cref{eq:score-error} bounds the score error by
\begin{equation}
\lvert e_j \rvert \lesssim \frac{2 \cdot 2^{-(t+1)}}{\sqrt d} \sum_{c=1}^{d} \lvert q_c\, k_{jc} \rvert ,
\label{eq:score-bound}
\end{equation}
The prefactor $2 \cdot 2^{-(t+1)} = 2^{-t}$ is $1/8$ in E4M3 and $1/2$ in E2M1. It multiplies the summed magnitudes of
the channel products over $\sqrt d$, which is the scaling of the score itself. The bound follows the magnitudes of the channel products and not the score,
since large products of opposite signs cancel in the score while their residuals remain. Some heads hold large values
in a few channels of $Q$ and $K$, the outliers of \cref{sec:motivation}. There the summed magnitudes of the channel
products are large, and the score error can exceed 1.0 in magnitude (\cref{tab:linearization}).

\subsection{Propagation through softmax}
\label{sec:der-softmax}
\label{sec:num-propagation}

We derive how the score errors change the attention weights and the output (\cref{eq:softmax-linear}). A quantized
kernel applies softmax to the scores $s_j + e_j$ in place of $s_j$. With $p_j = \exp(s_j)/\sum_l \exp(s_l)$, its
weights are
\begin{equation}
\tilde p_j = \frac{\exp(s_j + e_j)}{\sum_l \exp(s_l + e_l)} = \frac{p_j \exp(e_j)}{\sum_l p_l \exp(e_l)} .
\label{eq:softmax-exact}
\end{equation}

Softmax is invariant to a shift common to all scores. An error common to all keys of a query therefore does not
change the weights, and only the differences between the errors of the keys matter, which the ratio of two weights shows directly,
\begin{equation}
\frac{\tilde p_j}{\tilde p_{j'}} = \frac{p_j}{p_{j'}}\, \exp(e_j - e_{j'}) .
\label{eq:softmax-deviation}
\end{equation}
We subtract the $P$-weighted mean error of the query, $\bar e = \sum_l p_l e_l$, and call $e_j - \bar e$ the centered
error of key $j$. The weights depend on the centered errors alone,
$\tilde p_j = p_j \exp(e_j - \bar e) / \sum_l p_l \exp(e_l - \bar e)$.

We next linearize the weights in the centered errors. This first-order form relates the output error linearly to
the score errors, which is what lets \cref{sec:method} measure the score error by its energy and compensate it
term by term. For centered errors well below 1.0, $\exp(e_j - \bar e) \approx 1 + e_j - \bar e$. The
denominator is then $\sum_l p_l\,(1 + e_l - \bar e) = \sum_l p_l + \sum_l p_l e_l - \bar e \sum_l p_l = 1 + \bar e - \bar e = 1$,
using $\sum_l p_l = 1$ and the definition of $\bar e$. The weights become
\begin{equation}
\tilde p_j \approx p_j\,(1 + e_j - \bar e) ,
\label{eq:softmax-first-order}
\end{equation}
which is \cref{eq:softmax-linear}. Each weight changes by its centered error as a relative amount, and the changes
$\delta p_j = \tilde p_j - p_j \approx p_j\,(e_j - \bar e)$ sum to zero, as the normalization of softmax requires.
The first-order approximation of \cref{eq:softmax-first-order} holds while the centered errors stay below 1.0. Beyond that, the exponential of
\cref{eq:softmax-deviation} grows faster than the linear term, and the weights move by more than the first-order approximation
predicts.

Let $v_j$ be the value of token $j$ and $o = \sum_j p_j v_j$ the attention output. With $P$ and $V$ exact, the
quantized scores change the output by $\delta o_{\mathrm{QK}} = \sum_j \delta p_j\, v_j$. Inserting
$\delta p_j \approx p_j\,(e_j - \bar e)$ gives its first-order approximation,
\begin{equation}
\delta o^{(1)}_{\mathrm{QK}} = \sum_j p_j\,(e_j - \bar e)\, v_j .
\label{eq:softmax-output}
\end{equation}
To first order, the output error is thus linear in the centered score errors. The next paragraph measures the error
of this approximation on the captures.

\begin{table}[tbp]
\centering
\tablefont
\setlength{\tabcolsep}{2pt}
\caption{\textbf{Error of the first-order approximation.} Rotated NVFP4 $QK$ without ($\times$) and with (\checkmark) the
compensation of \deco, $P$ and $V$ exact, on the hard and on all layers of Wan2.2 and MiniMax-H3. The compensation
lowers all three quantities.}
\label{tab:linearization}
\input{docs/tables/linearization}
\end{table}

\paragraph{Error of the first-order approximation.} The error of the approximation is the norm of the difference between $\delta o_{\mathrm{QK}}$ and its first-order approximation, relative to the norm of $\delta o_{\mathrm{QK}}$,
\begin{equation}
\frac{\lVert \delta o_{\mathrm{QK}} - \delta o^{(1)}_{\mathrm{QK}} \rVert}{\lVert \delta o_{\mathrm{QK}} \rVert} ,
\label{eq:missed-error}
\end{equation}
with the norms taken over the queries of a head, or over those of all heads for the pooled value.
\Cref{tab:linearization} has four rows. The first two give the share of the keys outside the linear range, which are
the keys with $\lvert e_j - \bar e \rvert > 1.0$, and the attention weight that these keys hold. The last two give
the error of the approximation for the median head and pooled over the heads. Over the four columns of \cref{tab:linearization}, the NVFP4 $QK$ without compensation puts at most 7.1\% of the attention weight on these keys, yet the error of the approximation is 5.9 to 28.5\% of the output error of the median head and 68.4 to 97.9\% of the pooled output error. The output error thus comes from a minority of heads that leave the
linear range. With the compensation, these keys carry at most 0.23\% of the weight, and the error of the approximation is 3.0 to 10.9\% of the output error of the median head and 8.7 to 56.7\% of the pooled output error. The compensation thus brings the output error back toward the linear range, as stated in \cref{sec:qk}.

\begin{figure}[tbp]
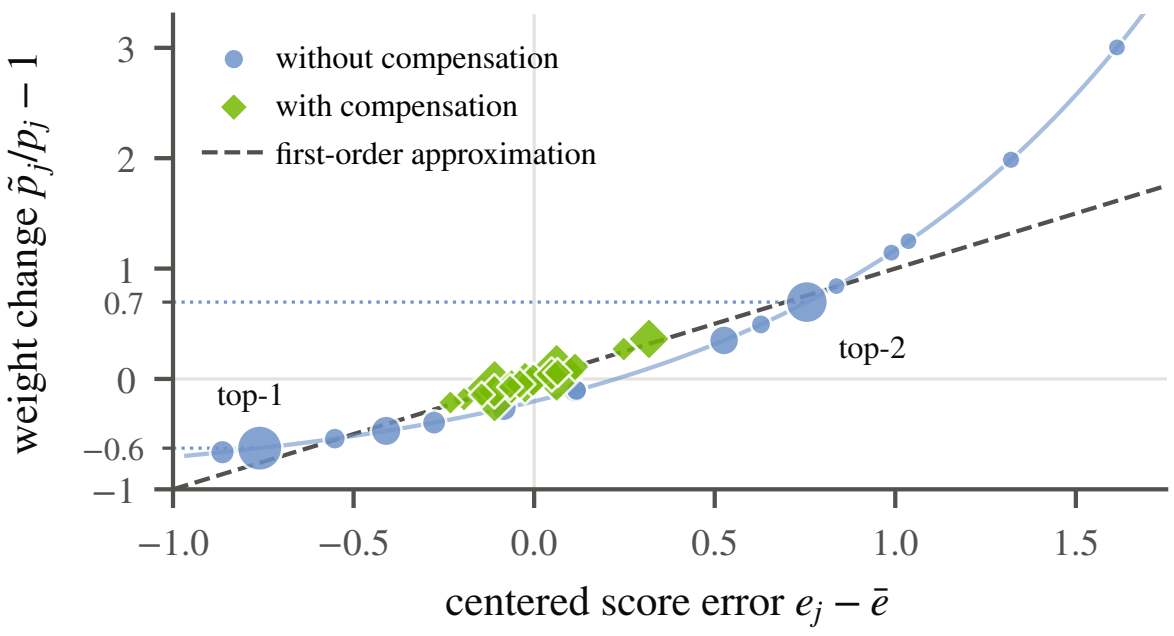

\centering
\colplot{plots/score_error_wan22-t2v-a14b.pdf}
\caption{\textbf{Weight changes of one query.} Relative weight change of the 16 leading keys of one query of a hard
layer of Wan2.2 against their centered score errors, without and with the compensation of \deco. Marker area is the
exact weight. The dashed line is the first-order approximation of \cref{eq:softmax-first-order}. The compensated keys
lie on the line, and without compensation the two leading keys swap their order.}
\label{fig:rounding}
\end{figure}

\paragraph{Weight changes of one query.} \Cref{fig:rounding} shows how the weights of the 16 leading keys of one query
change with their centered score errors, for a hard layer of Wan2.2 under the NVFP4 $QK$ without compensation and with
the compensated $QK$ of \deco{} (\cref{sec:eval-setup}). The dashed line is the first-order approximation of
\cref{eq:softmax-first-order}, on which a centered error of 0.5 raises the weight of its key by 50\%. With the
compensated $QK$, the centered errors stay within about a third, which keeps every key on the line, and the two leading
keys change their weights by about 11\% and 6\%. Without compensation, the centered errors reach beyond 1.0, where the
exponential of \cref{eq:softmax-deviation} grows much faster than the line. The two leading keys swap their order. The
top-1 key has a centered error of about $-3/4$, which lowers its weight by about 60\%, and the top-2 key has a centered
error of about $+3/4$, which raises its weight by about 70\%. The largest weight now lies on a different key, whose
value dominates the output. A score error below one unit thus produces a large output error, which is the amplification
described in \cref{sec:motivation}.

%% file: docs/tables/linearization.tex
\begin{spreadtabular}{lccccc}
\toprule
 & \spancell{2}{1}{\tablehead{Compensation}} & \multicolumn{2}{c}{\tablehead{Wan2.2}} & \multicolumn{2}{c}{\tablehead{MiniMax-H3}} \\
\cmidrule(lr){3-4}\cmidrule(lr){5-6}
 &  & \tablehead{Hard} & \tablehead{All} & \tablehead{Hard} & \tablehead{All} \\
\midrule
\spancell{2}{0}{Keys with $\lvert e_j - \bar e \rvert > 1.0$ (\%)} & $\times$ & 7.4 & 1.3 & 17 & 4.8 \\
 & \checkmark & 2.1 & 0.19 & 5.0 & 0.32 \\
\addlinespace
\spancell{2}{0}{Attention weight of these keys (\%)} & $\times$ & 0.67 & 0.095 & 7.1 & 1.4 \\
 & \checkmark & 0.017 & 0.0012 & 0.23 & 0.015 \\
\addlinespace
\spancell{2}{0}{Error of \cref{eq:missed-error}, median head (\%)} & $\times$ & 5.9 & 10.5 & 28.5 & 18.8 \\
 & \checkmark & 3.0 & 5.7 & 10.9 & 9.4 \\
\addlinespace
\spancell{2}{0}{Error of \cref{eq:missed-error}, pooled (\%)} & $\times$ & 93.5 & 68.4 & 97.9 & 83.9 \\
 & \checkmark & 17.1 & 8.7 & 56.7 & 28.7 \\
\bottomrule
\end{spreadtabular}

%% file: docs/compensation.tex
\section{Principal Residual Compensation}
\label{sec:app-comp}

This appendix supplements \cref{sec:qk}. We analyze the energy of the first-order terms and the share that the
compensation leaves, with the Hadamard rotation as the step that lets the eigenvalues measure this energy
(\cref{sec:der-energy}), split the error of the compensated $QK$ into its sources and show how the rotation and the
compensation add up (\cref{sec:qk-alone}), vary the rank (\cref{sec:abl-rank}), and give the preparation step by step
with its shared basis and its token sample (\cref{sec:prep-details}).

\subsection{Energy of the first-order terms}
\label{sec:der-energy}
\label{sec:residual-direction}

\deco reduces the error of a low-precision $QK$ numerically. It compensates the first-order terms of
\cref{eq:qk-split} along the $r$ leading eigenvectors of $G_K = \hat K^\top \hat K$ for the $Q$-side term and of
$G_Q = \hat Q^\top \hat Q$ for the $K$-side term (\cref{sec:qk}). \Cref{eq:qk-energy} measures the $Q$-side term by its
energy, its sum of squares over all keys, and splits this energy into one term per eigenvector of $G_K$. We show which
part of each term the compensation leaves, and that the eigenvalues predict it once the rotation has spread the
residual evenly.

The $K$-side term $\hat q\cdot\delta k_j$ has the same form with the operands exchanged. For one key, its energy over
all queries is
\begin{equation}
\begin{colgather}
\sum_i (\hat q_i\cdot\delta k_j)^2 = \delta k_j\, G_Q\, \delta k_j^\top = \sum_{m=1}^{d} \mu_m\, (\delta k_j\, w_m)^2,
\colpair G_Q = \hat Q^\top \hat Q = \sum_m \mu_m w_m w_m^\top ,
\end{colgather}
\label{eq:qk-energy-k}
\end{equation}
where $\mu_m$ and $w_m$ are the eigenvalues and eigenvectors of $G_Q$.

The compensation covers the $r$ leading eigenvectors, so the terms beyond the $r$-th remain. Summed over the queries
$i$ of a head, the share of the energy of the $Q$-side term that remains is
\begin{equation}
\eta(r) = \frac{\sum_i \sum_{m > r} \lambda_m\, (\delta q_i\, u_m)^2}{\sum_i \sum_{m} \lambda_m\, (\delta q_i\, u_m)^2} ,
\label{eq:qk-remainder}
\end{equation}
and that of the $K$-side term has $\mu_m$, $w_m$ and the sum over the keys. For a basis other than these eigenvectors,
$\eta(r)$ is the energy of the part of the term outside the $r$ directions of the basis over the energy of the term.
\input{docs/floats/rotation_residual}

\paragraph{Direction of the residual.} By \cref{eq:qk-energy}, the error energy along an
eigenvector $u_m$ is $\lambda_m\, (\delta q\, u_m)^2$. The eigenvalues alone predict this energy only if the residual
is spread evenly over the eigenvectors. The eigenvectors $u_1, \dots, u_d$ are orthonormal and span all $d$ channels, so
$\sum_m u_m u_m^\top = I$, and the squared components of the residual along them add up to its energy,
\begin{equation}
\begin{colgather}
\sum_m (\delta q\, u_m)^2 = \sum_m \delta q\, u_m u_m^\top\, \delta q^\top \colbreak
= \delta q\, \Big( \sum_m u_m u_m^\top \Big)\, \delta q^\top = \lVert \delta q \rVert^2 .
\end{colgather}
\label{eq:residual-components}
\end{equation}
If the residual is spread evenly over the eigenvectors, every component holds the same share, $(\delta q\, u_m)^2 = \lVert \delta q \rVert^2 / d$. The
error energy of \cref{eq:qk-energy} then reduces to $(\lVert \delta q \rVert^2 / d) \sum_m \lambda_m$, which follows
from the eigenvalues alone. We measure how far the residual is from this case by the direction factor $\gamma$, the
measured error energy divided by this value,
\begin{equation}
\gamma = \frac{\sum_i \sum_m \lambda_m\, (\delta q_i\, u_m)^2}{\frac{1}{d}\, \sum_i \lVert \delta q_i \rVert^2\, \sum_m \lambda_m} ,
\label{eq:direction-factor}
\end{equation}
with $i$ over the queries of the head. An evenly spread residual has $\gamma = 1$. The $K$-side term has the same factor
with $\delta k_j$, $\mu_m$ and $w_m$ and the sum over the keys $j$.

\paragraph{Hadamard rotation.} The Hadamard rotation is meant to spread the residual evenly over the eigenvectors
(\cref{sec:qk}), so that the eigenvalues measure the error energy. \Cref{tab:rotation-residual} tests this with the
direction factor $\gamma$ and with the first-order error energy of \cref{eq:qk-energy}, each without and with the
rotation. Without the rotation, $\gamma$ is 1.20 to 1.63 over the four
sets, because the leading eigenvectors lie along the outlier channels, which also carry the largest rounding errors.
The rotation brings $\gamma$ to 1.01 to 1.13. Each term of \cref{eq:qk-energy} is then nearly proportional to its
eigenvalue, and the first-order error energy falls to 0.64 of its value without the rotation on the hard layers of Wan2.2 and to 0.80 to
0.85 on the other three sets, before any compensation.

With the rotation, $(\delta q\, u_m)^2$ is thus nearly the same along every $u_m$. The remainder $\eta(r)$ of
\cref{eq:qk-remainder} is then close to the share of the energy of $K$ outside the $r$ leading eigenvectors, which is
one minus the share that the basis holds in \cref{fig:motivation}b.

\input{docs/floats/qk_residual}

\paragraph{Measured remainder.} \Cref{tab:qk-residual} gives the measured $\eta(r)$ of both terms over all layers of Wan2.2, with the rotation. With the
eigenvectors of $G_K$ and $G_Q$ from all tokens as the bases, 25.6\% of the $Q$-side term and 26.4\% of the $K$-side
term remain at $r = 32$. These values are close to the complements of the 74.3\% of the energy of $K$ and the 73.7\% of the energy
of $Q$ that the 32 leading eigenvectors hold (\cref{tab:basis-directions}), as the proportionality of \cref{sec:qk}
predicts. A random basis leaves 74.8\%, close to $1 - r/d$, because $r$ random directions hold on average a share
$r/d$ of the energy. A few principal directions thus hold most of the error of both terms. \Cref{sec:prep-details} discusses the two
remaining rows of the table.

\paragraph{Effect of the two terms.} Of the two terms, the $K$-side term has by far the larger effect on the output on
the hard layers. There, the kernel error on Wan2.2 is 18.35\% without the compensation of the $K$-side term and
3.97\% without that of the $Q$-side term (\cref{tab:abl-acc}). The reason is that softmax passes on only the
differences of the score errors between the keys of a query (\cref{eq:softmax-deviation}). For two keys $j$ and $j'$
of one query, the errors of the two terms differ by
\begin{equation}
\begin{colgather}
\delta q\cdot(\hat k_j - \hat k_{j'}) = \delta q\cdot(k_j - k_{j'}) - \delta q\cdot(\delta k_j - \delta k_{j'}) , \colpair
\hat q\cdot(\delta k_j - \delta k_{j'}) .
\end{colgather}
\label{eq:side-differences}
\end{equation}
Two similar keys can round in opposite directions, so the difference $\delta k_j - \delta k_{j'}$ of their residuals
is not small. In the $K$-side term, the whole query $\hat q$ multiplies it. In the $Q$-side term, only the residual
$\delta q$ multiplies it, which makes it a second-order term, next to $\delta q$ times the difference of the keys
themselves, which is small for similar keys. The $K$-side term therefore dominates wherever the weight of a query lies on similar keys. On the hard layers it
does, as the ablation of \cref{tab:abl-acc} shows.

\subsection{Error of the compensated QK}
\label{sec:qk-alone}
\label{sec:rotation-compensation}
\label{sec:abl-rotation}

\input{docs/floats/qk_rotation}

\input{docs/floats/qk_rotation_parts}

We show that the truncation drops the energy that $\eta(r)$ counts, how much of the error the compensation removes,
and how the Hadamard rotation and the compensation add up.

\paragraph{Truncation.} \Cref{eq:qk-truncation} expands the $Q$-side term of a query in the eigenvectors $u_m$ of $G_K$ and keeps the first $r$
terms. The dropped part is the sum over $m > r$. Its energy over all keys is
\begin{equation}
\begin{colgather}
\sum_j \Big( \sum_{m > r} (\delta q\, u_m)(\hat k_j\, u_m) \Big)^2 \colbreak
= \sum_{m > r} \sum_{m' > r} (\delta q\, u_m)(\delta q\, u_{m'}) \sum_j (\hat k_j\, u_m)(\hat k_j\, u_{m'}) .
\end{colgather}
\label{eq:qk-dropped}
\end{equation}
The inner sum over the keys is $u_m^\top G_K\, u_{m'}$. It equals $\lambda_m$ for $m' = m$ and zero otherwise, because
the $u_m$ are orthonormal eigenvectors of $G_K$. The cross terms therefore vanish, and the dropped part has the energy
\begin{equation}
\sum_j \Big( \sum_{m > r} (\delta q\, u_m)(\hat k_j\, u_m) \Big)^2 = \sum_{m > r} \lambda_m\, (\delta q\, u_m)^2 ,
\label{eq:qk-dropped-energy}
\end{equation}
which are the terms of \cref{eq:qk-energy} beyond the $r$-th, the numerator of \cref{eq:qk-remainder}. \deco thus
compensates the first $r$ terms of \cref{eq:qk-energy} and drops the other $d - r$. The eigenvalues are in descending
order, $\lambda_1 \ge \dots \ge \lambda_d$, so the dropped directions are those along which the keys have the least
energy. The same holds for the
$K$-side term.

\paragraph{Parts of the error.} The compensated $QK$ of \cref{sec:qk} departs from the exact scores in three places. The second-order term
$\delta q\cdot\delta k$ is neglected in \cref{eq:qk-split}, each first-order term is truncated to $r$ directions in
\cref{eq:qk-truncation}, and the projections onto $B$ are rounded to NVFP4 in \cref{eq:qk-channels}. Two more runs separate the three parts. One leaves the projections in FP32, which removes their rounding. The other
also compensates all 128 directions, which leaves only the second-order term. The differences
between the squared mean errors of the runs give each part (\cref{tab:qk-rotation-parts}).

\Cref{tab:qk-rotation} gives the error of $QK$ alone, with $P$ and $V$ exact. With the rotation, the compensation
lowers the mean error on the hard layers of Wan2.2 from 18.57\% to 2.10\% and over all layers from 8.00\% to 3.45\%,
and on MiniMax-H3 from 30.14\% to 5.90\% and from 9.98\% to 4.19\%. \Cref{tab:qk-rotation-parts}
gives the parts of these errors. On Wan2.2 with the rotation, the first-order terms in the 32 compensated directions
amount to 18.47\% on the hard layers and to 7.27\% over all layers, most of the error of the uncompensated $QK$. Of
the parts that remain, the first-order terms in the dropped directions are the largest on every set, 1.69\% and
3.29\% on Wan2.2.

\paragraph{Rotation and compensation.} The rotation helps in three ways, and its gain adds to that of the
compensation. First, it lowers the score error before any compensation, on the hard layers of Wan2.2 from 30.12\% to
18.57\% (\cref{tab:qk-rotation}). It spreads the residuals of $Q$ and $K$ over all directions, which takes them off the directions in which the
other operand is largest (\cref{sec:der-energy}). Second, the evenly spread
residual lets the eigenvalues measure the error, on which the choice of the compensated directions rests. Third, it
evens out the values within a block of 16 channels, which lowers the two smaller parts of the compensated error
(\cref{tab:qk-rotation-parts}). The rounding of the projections falls from 2.73\% to 1.01\% and the second-order term
from 1.60\% to 0.75\%. Most of this gain comes from the heads in which two large values share a block, where NVFP4
rounds the smaller one coarsely.

The cost is a weaker compensation of the first-order terms. Without the rotation, the residual lies along the leading eigenvectors,
which the compensation covers. The rotation moves a small part of it into the dropped directions, whose error rises
from 1.62\% to 1.69\%. With the compensation, the rotation lowers the error of the $QK$ from
3.56\% to 2.10\% and the error of the kernel from 3.94\% to 2.80\%.

Blocks with one dominant value do not gain from the rotation. The block scale stores the largest value of a block
almost exactly, so such a block is quantized well as it is. The rotation spreads this value over the channels, where every part of it is rounded to 4 bits. Some heads of the hard layers of MiniMax-H3 are of
this kind. There the unrotated operands have the smaller parts of the $QK$ error and the smaller error of the kernel,
5.82\% against 6.31\% (\cref{tab:qk-rotation,tab:qk-rotation-parts}). Over all layers of MiniMax-H3, the error of
the kernel is 4.40\% with and without the rotation, and on Wan2.2 the rotation lowers it. The kernels therefore keep it
for every model.

\subsection{Compensation rank}
\label{sec:abl-rank}

\input{docs/floats/ablation_rank}

We show that the error falls with the rank $r$ and that $r = 32$ is the largest rank that does not add an MMA to the
$QK$ on GB300. \Cref{tab:abl-rank} varies the rank. On the
hard layers of Wan2.2, $r = 16$ raises the error from 2.80\% to 3.97\% and $r = 64$ lowers it to 2.42\%, and over all
layers $r = 64$ lowers it from 3.73\% to 2.81\%.

A higher rank costs time in the preparation and in the kernel. The preparation projects every token of $Q$ and $K$
onto $r$ directions twice, and at $r = 32$ it takes 2.7 to 3.6\% of the attention time (\cref{sec:prep-details}). In
the kernel, the operands $Q'$ and $K'$ have $128 + 2r$ channels, 192 at $r = 32$, which is 50\% more than the 128 of
$Q$ and $K$, so every tile loads 50\% more bytes of them. The $QK$ also takes more NVFP4 MMAs per tile once its
channels exceed the reduction depth of an MMA, which is up to 96 on GB300, 128 on VR200, and 64 on RTX PRO 6000. On
GB300, $r = 16$ and $r = 32$ take two MMAs per tile, as many as the uncompensated $QK$, and $r = 64$ takes a third.
On VR200, every rank above zero takes two MMAs, one more than the uncompensated $QK$. On RTX PRO 6000, $r = 16$ and
$r = 32$ take three MMAs per tile and $r = 64$ four.

We therefore use $r = 32$ on every GPU. It is the largest rank that does not add an MMA on GB300, and it takes as many
MMAs as $r = 16$ on VR200 and RTX PRO 6000. On VR200, $r = 64$ would take the same two MMAs, but the kernels keep one rank for all GPUs.

\subsection{Preparation}
\label{sec:prep-details}
\label{sec:prep}

\begin{algorithm}[tbp]
\caption{Preparation of one head}
\label{alg:prep}
\algfont
\begin{algorithmic}[1]
\Require $Q, K, V \in \mathbb{R}^{N \times d}$, rank $r$, rotation $H$, block size of $V$
\Ensure $Q', K'$ with $d + 2r$ channels, $\hat V$, block means $\bar v$
\State $\mathcal S \gets$ sample of $\max(N/256,\, 2r)$ tokens
\State $G \gets$ \cref{eq:shared-basis} on $Q_{\mathcal S} H$ and $K_{\mathcal S} H$
\State $B \gets$ the $r$ leading eigenvectors of $G$
\For{each token $i$} \Comment{pass over $Q$ and $K$}
  \State $q \gets q_i H$, \ $\hat q \gets \Qq(q)$, \ $\delta q \gets q - \hat q$
  \State $q'_i \gets [\,\hat q \mid \Qq(\delta q\, B) \mid \Qq(\hat q\, B)\,]$
  \State $k \gets k_i H$, \ $\hat k \gets \Qq(k)$, \ $\delta k \gets k - \hat k$
  \State $k'_i \gets [\,\hat k \mid \Qq(\hat k\, B) \mid \Qq(\delta k\, B)\,]$
\EndFor
\For{each block $b$ of keys} \Comment{pass over $V$}
  \State $\bar v_b \gets$ mean of $v_j$ over $j \in b$
  \State $\hat v_j \gets \Qq(v_j - \bar v_b)$ for $j \in b$
\EndFor
\end{algorithmic}
\end{algorithm}

We give the preparation of \cref{sec:basis} step by step. \Cref{alg:prep} lists it for one head. It runs as a chain
of fused kernels, an eigensolver for the bases of all heads and one pass each over $Q$ and $K$ and over $V$.
\begin{itemize}[leftmargin=*,itemsep=2pt]
  \item \emph{Rotation.} $H$ is a normalized Hadamard matrix of order 128 with random signs on its rows. Since
    $HH^\top = I$, the rotation leaves the scores and the eigenvalues of $G_Q$ and $G_K$ unchanged.
  \item \emph{Sampling.} The sample holds $1/256$ of the tokens, at least $2r = 64$, and is drawn before the pass over
    $Q$ and $K$, which needs $B$. On a sequence of text, audio and video tokens, half of the sample comes from the text
    and audio tokens, or all of them when they are fewer, and the rest from the video tokens, so that both groups weigh
    equally in the basis.
  \item \emph{$Q'$ and $K'$.} Each operand holds three groups of channels, for $q'$ the 128 channels of $\hat q$, the
    32 of $\delta q\, B$ and the 32 of $\hat q\, B$. $\Qq$ rounds every group to NVFP4, E2M1 elements with an unsigned
    E4M3 scale per block $b$ of 16 channels,
    \begin{equation}
    \begin{colgather}
    s_g = 2^{\lceil \log_2 (a_g / (6 \cdot 448)) \rceil} , \qquad
    \sigma_b = \Qq_{\mathrm{E4M3}}\Big( \frac{\max_{c \in b} \lvert x_c \rvert}{6\, s_g} \Big) , \colpair
    \hat x_c = s_g\, \sigma_b\, \Qq_{\mathrm{E2M1}}\Big( \frac{x_c}{s_g\, \sigma_b} \Big) ,
    \end{colgather}
    \label{eq:nvfp4}
    \end{equation}
    where $a_g$ is the largest magnitude of the group in the head and 6 and 448 are the largest values of E2M1 and
    E4M3. The block scale $\sigma_b$ maps the largest element of its block to 6, and $s_g$ keeps every block scale at or
    below 448. Each group has its own $s_g$, because the projections of the residual are much smaller than the other
    two groups.
  \item \emph{$V$.} With FP8 $PV$, the blocks hold 128 keys and $\Qq$ rounds to E4M3 with one power-of-two scale
    per tensor. With NVFP4 $PV$, the blocks hold 16 keys (\cref{sec:num-pv}).
\end{itemize}

\input{docs/floats/basis_directions}

\input{docs/floats/ablation_basis}

\paragraph{Shared basis.}\label{sec:abl-compensation} We show that the leading subspaces of $Q$ and $K$ overlap enough for one shared basis to hold nearly as much of the
energy of each operand as separate bases, and that a basis of channels or a random basis leaves most of the error.
\deco shares one basis between $Q$ and $K$ (\cref{sec:basis}). \Cref{tab:basis-directions} measures the overlap of the
two leading subspaces as the mean share of a leading eigenvector of $G_Q$ that lies in the span of the leading
eigenvectors of $G_K$, which is 1 for identical subspaces and $r/d$ for unrelated ones. At $r = 32$ the overlap is
0.67 to 0.76 over the four sets, with 0.75 over all layers of Wan2.2 as quoted in \cref{sec:basis}. Over all layers of
Wan2.2, the shared basis holds 71.9\% of the energy of $K$, against 74.3\% for separate bases. Separate bases give
nearly the same error (\cref{tab:abl-basis}), 2.91\% against 2.80\% on the hard layers of Wan2.2 and 5.97\% against
6.31\% on those of MiniMax-H3.

\Cref{tab:abl-basis} also replaces the principal directions. The channel basis takes the 32 channels with the largest
diagonal entries of $G$, so that the extra channels hold the residuals and the quantized values of these channels, as
ARCQuant does for outlier channels~\citep{arcquant}. With the rotation, this basis leaves about as much error as a
random basis, because the rotation spreads the energy over all channels. Without the rotation, the channel basis reaches 4.48\% on
the hard layers of Wan2.2 and 7.75\% on those of MiniMax-H3, against 3.94\% and 5.82\% for the unrotated shared
basis (\cref{tab:qk-rotation}). With the
same number of extra channels, the principal directions thus remove the most error on every set, which shows the
benefit of compensating along them.

\input{docs/floats/ablation_sample}

\paragraph{Token sample.}\label{sec:abl-sample} We show that a basis estimated from a small sample of the tokens holds nearly as much energy and gives nearly the same
error as a basis from all tokens. \deco estimates the basis from a random sample of $1/256$ of the tokens
(\cref{sec:basis}). Over all layers, the sampled basis holds 69.0\% of the energy of $K$ on Wan2.2 and 77.5\% on
MiniMax-H3, against 71.9\% and 82.7\% for the shared basis from all tokens (\cref{tab:basis-directions}). On Wan2.2,
it leaves 30.8\% and 31.3\% of the two first-order terms, against 27.8\% and 28.7\% (\cref{tab:qk-residual}). Sharing
and sampling together thus lower the energy of $K$ in the basis over all layers of Wan2.2 from 74.3\% to 69.0\%, the
two values of \cref{sec:basis}. A basis from $1/64$ or
from all of the tokens lowers the error on the hard layers of Wan2.2 from 2.80\% to 2.77\% and on those of MiniMax-H3
from 6.31\% to 6.07\% and 5.78\% (\cref{tab:abl-sample}). The larger gain on MiniMax-H3 comes from the token types.
Half of the sample is drawn from the text and audio tokens, whereas a basis from all tokens
follows the video tokens.

\paragraph{Common scale of the three channel groups.} The MMA over the 192 channels of $Q'$ and $K'$ sums the three
groups into one score, which can carry only one scale $S$. The kernel takes as $S$ the largest of the three products of
a $Q'$ group scale and its $K'$ group scale. I-ExpCast can raise $S$ further, because it folds $S$ into the multiplier
$a$ of its FMA, an FP32 subnormal, which keeps $S$ at or above $2^{-10}$ with FP8 $PV$ and $2^{-6}$ with NVFP4 $PV$
(\cref{eq:i-expcast-steps}). A group whose product lies below $S$ shrinks its block scales by the ratio, which leaves
the scores unchanged. A block scale below $2^{-6}$, the end of the normal range of E4M3, loses accuracy. The kernel
therefore spreads the shrink over the block scales of $Q'$ and $K'$. The group scales and $S$ are powers of two, so
the shrink shifts the exponent of a block scale without a second rounding.

\paragraph{Cost.} Per head, the attention takes $N^2\,(d + 2r)$ multiply-adds for the $QK$ and $N^2 d$ for the $PV$,
and the preparation $2Nd^2$ for the rotation of $Q$ and $K$ and $4Ndr$ for the four projections. The Gram matrices of
the sample and the block means of $V$ add little, and the eigendecomposition of a $d \times d$ matrix does not grow
with $N$. The preparation is thus linear in $N$ and the attention quadratic, with the ratio
\begin{equation}
\frac{N\,(2d^2 + 4dr)}{N^2\,(2d + 2r)} = \frac{d\,(d + 2r)}{N\,(d + r)} ,
\label{eq:prep-ratio}
\end{equation}
about $154/N$ for $d = 128$ and $r = 32$, or 0.2\% for the 75,600 tokens of a Wan2.2 call. In time, the preparation
adds 2.7 to 3.6\% to the attention (\cref{tab:time-full}), as stated in \cref{sec:basis}. This exceeds its share of
the operations, partly because the rotation and the projections run as BF16 and TF32 MMAs, which are slower than the NVFP4
and FP8 MMAs of the attention, and because every operand is read and written once.

%% file: docs/floats/rotation_residual.tex
\begin{table}[tbp]
\centering
\tablefont
\setlength{\tabcolsep}{3pt}
\caption{\textbf{Residual without and with the rotation.} Direction factor $\gamma$ of \cref{eq:direction-factor} and
first-order error energy relative to its value without the rotation, without ($\times$) and with (\checkmark) the
Hadamard rotation, as medians over the heads and
means of the $Q$ and the $K$ side. The rotation brings $\gamma$ close to 1 and lowers the first-order error energy on
every set.}
\label{tab:rotation-residual}
\input{docs/tables/rotation_residual}
\end{table}

%% file: docs/tables/rotation_residual.tex
\begin{spreadtabular}{lccccc}
\toprule
 & \spancell{2}{1}{\tablehead{Rotation}} & \multicolumn{2}{c}{\tablehead{Wan2.2}} & \multicolumn{2}{c}{\tablehead{MiniMax-H3}} \\
\cmidrule(lr){3-4}\cmidrule(lr){5-6}
 &  & \tablehead{Hard} & \tablehead{All} & \tablehead{Hard} & \tablehead{All} \\
\midrule
\spancell{2}{0}{Direction factor $\gamma$} & $\times$ & 1.63 & 1.20 & 1.62 & 1.33 \\
 & \checkmark & 1.02 & 1.01 & 1.13 & 1.03 \\
\addlinespace
\spancell{2}{0}{First-order error energy} & $\times$ & 1.00 & 1.00 & 1.00 & 1.00 \\
 & \checkmark & 0.64 & 0.85 & 0.80 & 0.83 \\
\bottomrule
\end{spreadtabular}

%% file: docs/floats/qk_residual.tex
\begin{table}[tbp]
\centering
\tablefont
\setlength{\tabcolsep}{3pt}
\caption{\textbf{First-order terms outside the compensation.} Remainder $\eta(r)$ (\%) of \cref{eq:qk-remainder} of the
$Q$-side and the $K$-side term of the rotated NVFP4 $QK$ at $r = 16$, 32 and 64, median over the heads of all layers of
Wan2.2, for separate bases of the leading eigenvectors of each operand, one basis shared by $Q$ and $K$, the shared
basis from the $1/256$ token sample of the kernels, and a random basis. The eigenvectors of each operand leave about a quarter of each term at $r = 32$, a random basis three quarters.}
\label{tab:qk-residual}
\input{docs/tables/qk_residual}
\end{table}

%% file: docs/tables/qk_residual.tex
\begin{spreadtabular}{llcccccc}
\toprule
\spancell{2}{1}{\tablehead{Basis}} & \spancell{2}{1}{\tablehead{Tokens}} & \multicolumn{3}{c}{\tablehead{$Q$-side term, $r =$}} & \multicolumn{3}{c}{\tablehead{$K$-side term, $r =$}} \\
\cmidrule(lr){3-5}\cmidrule(lr){6-8}
 &  & \tablehead{16} & \tablehead{32} & \tablehead{64} & \tablehead{16} & \tablehead{32} & \tablehead{64} \\
\midrule
separate & all & 42.1 & 25.6 & 10.7 & 43.0 & 26.4 & 11.3 \\
shared & all & 44.6 & 27.8 & 11.9 & 45.7 & 28.7 & 12.6 \\
\textbf{shared (Ours)} & 1/256 & 46.9 & 30.8 & 14.3 & 47.8 & 31.3 & 14.9 \\
random & none & 87.4 & 74.8 & 49.6 & 87.4 & 74.8 & 49.6 \\
\bottomrule
\end{spreadtabular}

%% file: docs/floats/qk_rotation.tex
\begin{table}[tbp]
\centering
\tablefont
\setlength{\tabcolsep}{3pt}
\caption{\textbf{Error without and with the rotation.} Mean relative L2 error (\%) without ($\times$) and with
(\checkmark) the Hadamard rotation, for $QK$ alone with $P$ and $V$ exact, uncompensated and compensated at $r = 32$,
and for the whole kernel with FP8 $PV$. The rotation lowers the errors on Wan2.2 and raises those on the
hard layers of MiniMax-H3.}
\label{tab:qk-rotation}
\input{docs/tables/qk_rotation}
\end{table}

%% file: docs/tables/qk_rotation.tex
\begin{spreadtabular}{lccccc}
\toprule
 & \spancell{2}{1}{\tablehead{Rotation}} & \multicolumn{2}{c}{\tablehead{Wan2.2}} & \multicolumn{2}{c}{\tablehead{MiniMax-H3}} \\
\cmidrule(lr){3-4}\cmidrule(lr){5-6}
 &  & \tablehead{Hard} & \tablehead{All} & \tablehead{Hard} & \tablehead{All} \\
\midrule
\spancell{2}{0}{Uncompensated $QK$} & $\times$ & 30.12 & 10.29 & 28.14 & 10.82 \\
 & \checkmark & 18.57 & 8.00 & 30.14 & 9.98 \\
\addlinespace
\spancell{2}{0}{Compensated $QK$} & $\times$ & 3.56 & 3.57 & 5.45 & 4.19 \\
 & \checkmark & 2.10 & 3.45 & 5.90 & 4.19 \\
\addlinespace
\spancell{2}{0}{Whole kernel} & $\times$ & 3.94 & 3.83 & 5.82 & 4.40 \\
 & \checkmark & 2.80 & 3.73 & 6.31 & 4.40 \\
\bottomrule
\end{spreadtabular}

%% file: docs/floats/qk_rotation_parts.tex
\begin{table}[tbp]
\centering
\tablefont
\setlength{\tabcolsep}{3pt}
\caption{\textbf{Parts of the QK error without and with the rotation.} Each part as the error (\%) that it would give
alone. The parts add in energy, that is, in their squares. The first three make up the uncompensated $QK$ of
\cref{tab:qk-rotation}, and the last three the compensated $QK$. The first-order terms are split into those along the
$r = 32$ compensated directions and those along the dropped directions. On Wan2.2, the rotation lowers the first-order terms along the compensated directions, the second-order term, and the
rounding. The terms along the dropped directions rise marginally.}
\label{tab:qk-rotation-parts}
\input{docs/tables/qk_rotation_parts}
\end{table}

%% file: docs/tables/qk_rotation_parts.tex
\begin{spreadtabular}{lccccc}
\toprule
 & \spancell{2}{1}{\tablehead{Rotation}} & \multicolumn{2}{c}{\tablehead{Wan2.2}} & \multicolumn{2}{c}{\tablehead{MiniMax-H3}} \\
\cmidrule(lr){3-4}\cmidrule(lr){5-6}
 &  & \tablehead{Hard} & \tablehead{All} & \tablehead{Hard} & \tablehead{All} \\
\midrule
\spancell{2}{0}{First order, compensated} & $\times$ & 30.03 & 9.75 & 27.77 & 10.06 \\
 & \checkmark & 18.47 & 7.27 & 29.74 & 9.13 \\
\addlinespace
\spancell{2}{0}{First order, dropped} & $\times$ & 1.62 & 3.20 & 4.35 & 3.91 \\
 & \checkmark & 1.69 & 3.29 & 4.64 & 3.97 \\
\addlinespace
\spancell{2}{0}{Second-order term} & $\times$ & 1.60 & 0.74 & 1.36 & 0.78 \\
 & \checkmark & 0.75 & 0.55 & 1.56 & 0.73 \\
\addlinespace
\spancell{2}{0}{Rounding of the projections} & $\times$ & 2.73 & 1.39 & 2.99 & 1.28 \\
 & \checkmark & 1.01 & 0.87 & 3.30 & 1.14 \\
\bottomrule
\end{spreadtabular}

%% file: docs/floats/ablation_rank.tex
\begin{table}[tbp]
\centering
\tablefont
\setlength{\tabcolsep}{3pt}
\caption{\textbf{Rank of the compensation.} $QK$ channels, NVFP4 MMAs of the $QK$ per tile on GB300 and VR200, and
relative L2 error (\%) of \deco with FP8 $PV$, per rank $r$. $r = 32$ is the largest rank with two MMAs
on GB300.}
\label{tab:abl-rank}
\input{docs/tables/ablation_rank}
\end{table}

%% file: docs/tables/ablation_rank.tex
\begin{spreadtabular}{lccccccc}
\toprule
\spancell{2}{1}{\tablehead{$r$}} & \spancell{2}{1}{\tablehead{Channels}} & \multicolumn{2}{c}{\tablehead{MMAs}} & \multicolumn{2}{c}{\tablehead{Wan2.2}} & \multicolumn{2}{c}{\tablehead{MiniMax-H3}} \\
\cmidrule(lr){3-4}\cmidrule(lr){5-6}\cmidrule(lr){7-8}
 &  & \tablehead{GB300} & \tablehead{VR200} & \tablehead{Hard} & \tablehead{All} & \tablehead{Hard} & \tablehead{All} \\
\midrule
0 & 128 & 2 & 1 & 18.70 & 8.10 & 29.40 & 10.00 \\
16 & 160 & 2 & 2 & 3.97 & 4.77 & 10.08 & 5.89 \\
\textbf{32 (Ours)} & 192 & 2 & 2 & 2.80 & 3.73 & 6.31 & 4.40 \\
64 & 256 & 3 & 2 & 2.42 & 2.81 & 4.92 & 3.18 \\
\bottomrule
\end{spreadtabular}

%% file: docs/floats/basis_directions.tex
\begin{table}[tbp]
\centering
\tablefont
\setlength{\tabcolsep}{3pt}
\caption{\textbf{Energy in the basis.} Share (\%) of the energy of $Q$ and $K$ in the 32 directions of separate bases
per operand, of one basis shared by $Q$ and $K$ from all tokens, and of the shared basis from the token sample of the
kernels, and the overlap of the leading subspaces of $Q$ and $K$, as medians over the heads. Sharing and sampling each
cost a few points of energy.}
\label{tab:basis-directions}
\input{docs/tables/basis_directions}
\end{table}

%% file: docs/tables/basis_directions.tex
\begin{spreadtabular}{lllcccc}
\toprule
\spancell{2}{1}{\tablehead{Basis}} & \spancell{2}{1}{\tablehead{Tokens}} &  & \multicolumn{2}{c}{\tablehead{Wan2.2}} & \multicolumn{2}{c}{\tablehead{MiniMax-H3}} \\
\cmidrule(lr){4-5}\cmidrule(lr){6-7}
 &  &  & \tablehead{Hard} & \tablehead{All} & \tablehead{Hard} & \tablehead{All} \\
\midrule
\spancell{2}{0}{separate} & \spancell{2}{0}{all} & $Q$ & 86.0 & 73.7 & 93.8 & 83.2 \\
 &  & $K$ & 87.4 & 74.3 & 96.3 & 85.1 \\
\addlinespace
\spancell{2}{0}{shared} & \spancell{2}{0}{all} & $Q$ & 83.8 & 71.1 & 92.5 & 80.8 \\
 &  & $K$ & 85.4 & 71.9 & 95.1 & 82.7 \\
\addlinespace
\spancell{2}{0}{\textbf{shared (Ours)}} & \spancell{2}{0}{1/256} & $Q$ & 82.6 & 68.5 & 88.9 & 75.8 \\
 &  & $K$ & 83.8 & 69.0 & 92.4 & 77.5 \\
\midrule
\multicolumn{3}{l}{Overlap of $Q$ and $K$} & 0.76 & 0.75 & 0.73 & 0.67 \\
\bottomrule
\end{spreadtabular}

%% file: docs/floats/ablation_basis.tex
\begin{table}[tbp]
\centering
\tablefont
\setlength{\tabcolsep}{3pt}
\caption{\textbf{Basis of the compensation.} Relative L2 error (\%) of \deco with FP8 $PV$ with the
shared basis, with separate bases $B_K$ and $B_Q$ per term, with the 32 channels of largest energy as the basis, and
with a random orthonormal basis, with (\checkmark) or without ($\times$) the Hadamard rotation. Separate bases change
the error by less than half a point. With the rotation, a channel basis or a random basis leaves most of the error.}
\label{tab:abl-basis}
\input{docs/tables/ablation_basis}
\end{table}

%% file: docs/tables/ablation_basis.tex
\begin{spreadtabular}{lccccc}
\toprule
\spancell{2}{1}{\tablehead{Basis}} & \spancell{2}{1}{\tablehead{Rotation}} & \multicolumn{2}{c}{\tablehead{Wan2.2}} & \multicolumn{2}{c}{\tablehead{MiniMax-H3}} \\
\cmidrule(lr){3-4}\cmidrule(lr){5-6}
 &  & \tablehead{Hard} & \tablehead{All} & \tablehead{Hard} & \tablehead{All} \\
\midrule
\textbf{shared (Ours)} & \checkmark & 2.80 & 3.73 & 6.31 & 4.40 \\
separate & \checkmark & 2.91 & 3.77 & 5.97 & 4.35 \\
channels & \checkmark & 15.16 & 6.78 & 24.99 & 8.22 \\
channels & $\times$ & 4.48 & 4.56 & 7.75 & 5.84 \\
random & \checkmark & 14.56 & 6.80 & 25.74 & 8.49 \\
\bottomrule
\end{spreadtabular}

%% file: docs/floats/ablation_sample.tex
\begin{table}[tbp]
\centering
\tablefont
\setlength{\tabcolsep}{3pt}
\caption{\textbf{Token sample of the basis.} Relative L2 error (\%) of \deco with FP8 $PV$ with the basis
estimated from $1/256$, $1/64$, or all of the tokens. The sample costs 0.03 points on the hard layers of Wan2.2 and
about half a point on those of MiniMax-H3.}
\label{tab:abl-sample}
\input{docs/tables/ablation_sample}
\end{table}

%% file: docs/tables/ablation_sample.tex
\begin{spreadtabular}{lcccc}
\toprule
\spancell{2}{1}{\tablehead{Tokens}} & \multicolumn{2}{c}{\tablehead{Wan2.2}} & \multicolumn{2}{c}{\tablehead{MiniMax-H3}} \\
\cmidrule(lr){2-3}\cmidrule(lr){4-5}
 & \tablehead{Hard} & \tablehead{All} & \tablehead{Hard} & \tablehead{All} \\
\midrule
\textbf{1/256 (Ours)} & 2.80 & 3.73 & 6.31 & 4.40 \\
1/64 & 2.77 & 3.65 & 6.07 & 4.26 \\
all & 2.77 & 3.62 & 5.78 & 3.99 \\
\bottomrule
\end{spreadtabular}

%% file: docs/pv.tex
\section{Smoothing of V}
\label{sec:app-pv}
\label{sec:num-pv}

\deco smooths $V$ before it quantizes $V$, following the V-Smooth of VC-Attention (\cref{sec:setup})~\citep{vcattention}.
We give the method, show what the smoothing removes from the error of each operand of the $PV$ product, and show why
the FP8-$PV$ kernels smooth per 128 keys and the NVFP4-$PV$ kernels per 16 keys.

\paragraph{Method.} Every key $j$ has a value vector $v_j$, and the keys of a block $b$ of consecutive keys share their
mean $\bar v_b$. The smoothing takes this shared part out of the low-precision $PV$. It subtracts $\bar v_b$ from the
values of its block, the kernel multiplies $P$ with the centered $V$ in low precision, and a small product in high
precision adds the block means back. Let $\phi_{ij}$ be the exponential of the score of key $j$ relative to the row
base of row $i$ and $b(j)$ the block that holds key $j$. The numerator of the attention output splits exactly into the
centered values and the block means,
\begin{equation}
\begin{colgather}
\sum_j \phi_{ij}\, v_j = \sum_j \phi_{ij}\,\big(v_j - \bar v_{b(j)}\big) + \sum_b L_{ib}\,\bar v_b , \colpair
L_{ib} = \sum_{j \in b} \phi_{ij} , \qquad \bar v_b = \frac{1}{\lvert b \rvert}\sum_{j \in b} v_j .
\end{colgather}
\label{eq:centering-split}
\end{equation}
The kernel rounds only the first term and adds the second with its block sums $\hat L_{ib}$ of $P$,
\begin{equation}
o_i = \frac{\sum_j \hat\phi_{ij}\,\widehat{\big(v_j - \bar v_{b(j)}\big)} + \sum_b \hat L_{ib}\,\bar v_b}{\sum_b \hat L_{ib}} ,
\label{eq:centering}
\end{equation}
where $\hat\phi_{ij}$ is the quantized $\phi_{ij}$ of the $PV$ MMA and $\hat L_{ib}$ is the block sum of $P$ that the
kernel forms (\cref{sec:kernel-ablations}). VC-Attention reorders the tokens by k-means and subtracts the mean of
each tile of 128 tokens. \deco keeps the token order, in which neighboring video tokens are similar, and smooths per
block of 16 keys with NVFP4 $PV$. The small blocks follow $V$ more closely than tiles of 128 tokens and do not need a
grouping in token order (\cref{tab:abl-smooth-full}).

\begin{figure}[tbp]
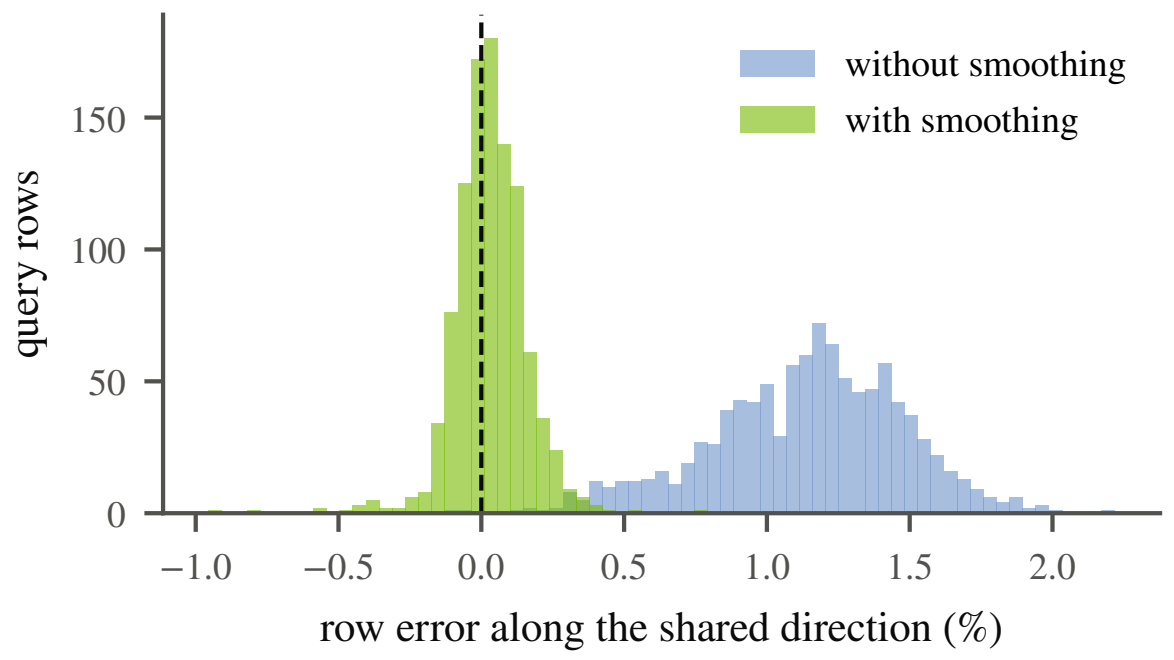

\centering
\colplot{plots/smoothing_wan22-t2v-a14b.pdf}
\caption{\textbf{Bias of a 4-bit P.} Error of each query row of a typical head of Wan2.2 along the mean error of the
head, with $P$ rounded to NVFP4 and $V$ exact, without and with smoothing per 16 keys. Without the smoothing the row
errors lean to one side, with it they center near zero.}
\label{fig:smoothing}
\end{figure}

\paragraph{Effect on P and V.} The smoothing lowers the error of both operands of the $PV$. For $P$, it returns the shared part of every contribution that a 4-bit $P$ drops. An NVFP4 $P$ rounds every probability below 1/24 of its block maximum to zero, so
without the smoothing the output loses the whole contribution $\phi_{ij}\, v_j$ of such a key. With the smoothing,
the part $\phi_{ij}\, \bar v_{b(j)}$ of this contribution returns to the output through the block sum $\hat L_{ib}$,
which the NVFP4-$PV$ kernels form from most probabilities before the rounding (\cref{sec:kernel-ablations}). Only the
part $\phi_{ij}\,(v_j - \bar v_{b(j)})$ is lost. For $V$, the centered values have zero mean in every block, so these
losses have both signs and largely cancel, where the losses of the uncentered values all point along the block mean.
The centered values are also smaller than $V$, so a floating-point format rounds them with a smaller error, and the
block means stay in high precision. \Cref{fig:smoothing} shows the result. Without the smoothing, the errors of the
rows lean to one side, and with it they center near zero. The cost is one more MMA, which multiplies the block sums
of $P$ with the block means in high precision and which smaller blocks need more often.

\input{docs/floats/ablation_smooth_full}

\paragraph{Block size.} The block size follows from the accuracy and from what bounds each kernel.
\Cref{tab:abl-smooth-full} gives the error per block size.
An FP8 $P$ and $V$ round finely already, so the smoothing does not lower the error by much. On the hard layers of Wan2.2, blocks of 128
keys lower the error from 2.95\% to 2.80\% and blocks of 16 keys to 2.66\%, but the smaller blocks need the extra MMA
four times as often, once per two key tiles instead of eight. The FP8-$PV$ kernel already runs close to the time of
its MMAs (\cref{sec:rebase}), so every extra MMA lengthens it. It therefore smooths per 128 keys, which lengthens the
kernel on GB300 from 55.8 to 58.0~ms on a Wan2.2 call (\cref{tab:abl-perf}). An NVFP4 $PV$ gains much more, and most
from small blocks. Blocks of 16 keys lower the error from 7.50\% to 5.68\% on the hard layers of Wan2.2 and from
5.81\% to 5.09\% over all layers, against 6.58\% and 5.43\% with 128 keys. On GB300 and VR200, the softmax warps
bound the NVFP4-$PV$ kernel (\cref{sec:kernel-ablations}), so its tensor cores have room for the extra MMAs. The
NVFP4-$PV$ kernels therefore smooth per 16 keys.

%% file: docs/floats/ablation_smooth_full.tex
\begin{table}[tbp]
\centering
\tablefont
\setlength{\tabcolsep}{3pt}
\caption{\textbf{Block of the smoothing.} Relative L2 error (\%) of \deco with $V$ as
it is (none) and smoothed per block of 16, 32 or 128 keys, per $PV$ format. The smoothing does not lower the error by much with FP8 $PV$ and lowers it most with NVFP4 $PV$ and
small blocks.}
\label{tab:abl-smooth-full}
\input{docs/tables/ablation_smooth_full}
\end{table}

%% file: docs/tables/ablation_smooth_full.tex
\begin{spreadtabular}{llcccc}
\toprule
\spancell{2}{1}{\tablehead{PV}} & \spancell{2}{1}{\tablehead{Block}} & \multicolumn{2}{c}{\tablehead{Wan2.2}} & \multicolumn{2}{c}{\tablehead{MiniMax-H3}} \\
\cmidrule(lr){3-4}\cmidrule(lr){5-6}
 &  & \tablehead{Hard} & \tablehead{All} & \tablehead{Hard} & \tablehead{All} \\
\midrule
\spancell{4}{0}{FP8} & none & 2.95 & 3.75 & 6.37 & 4.44 \\
 & 16 & 2.66 & 3.70 & 6.29 & 4.38 \\
 & 32 & 2.73 & 3.71 & 6.31 & 4.39 \\
 & \textbf{128 (Ours)} & 2.80 & 3.73 & 6.31 & 4.40 \\
\midrule
\spancell{4}{0}{NVFP4} & none & 7.50 & 5.81 & 9.16 & 6.55 \\
 & \textbf{16 (Ours)} & 5.68 & 5.09 & 8.24 & 5.66 \\
 & 32 & 6.15 & 5.26 & 8.37 & 5.78 \\
 & 128 & 6.58 & 5.43 & 8.44 & 5.85 \\
\bottomrule
\end{spreadtabular}

%% file: docs/kernel_details.tex
\section{Kernels}
\label{sec:kernel-details}

This appendix supplements \cref{sec:performance}. On GB300 and VR200, the softmax between the two MMAs bounds the
kernel, so the faster 4-bit $QK$ MMA shortens the kernel only when the softmax is shortened as well. The three
optimizations of \cref{sec:performance} do this. We derive the bit pattern of I-ExpCast, bound the error of its
approximation and measure its accuracy and speed in the kernels (\cref{sec:kernel-p-details}), measure the row sum and
the block sums of $P$ in the $PV$ MMA (\cref{sec:kernel-ablations}), and measure the threshold of the lazy rescale
(\cref{sec:rebase}).

\subsection{I-ExpCast}
\label{sec:kernel-p-details}

\begin{figure*}[tbp]
\centering
\small
\setlength{\tabcolsep}{6pt}
\begin{tabular}{p{0.2\textwidth}p{0.2\textwidth}p{0.2\textwidth}p{0.2\textwidth}}
\toprule
\ttfamily\fontsize{6}{7}\selectfont x = ffma2(s, a, c)\newline h = cvt.f16x2(x)\newline p = ex2.approx.f16x2(h)\newline q = cvt.e4m3x2.f16x2(p) &
\ttfamily\fontsize{6}{7}\selectfont b = ffma2(s, 8a, c)\newline h = cvt.relu.f16x2(b)\newline y = fma.f16x2(h, 1, 1024)\newline q = prmt(y01, y23) &
\ttfamily\fontsize{6}{7}\selectfont t = ffma2(s, a, c)\newline w = cvt.pack.sat.u16(t)\newline q = cvt.e4m3x2.f16x2(w) &
\ttfamily\fontsize{6}{7}\selectfont u = ffma2(s, a, c)\newline q = cvt.relu.e2m1x2.f32(u) \\
\addlinespace
8 instructions & 7 instructions & 6 instructions & 4 instructions \\
\bottomrule
\addlinespace[3pt]
\multicolumn{1}{c}{(a) exp2 on the SFU} & \multicolumn{1}{c}{(b) ExpCast} & \multicolumn{1}{c}{(c) I-ExpCast} &
\multicolumn{1}{c}{(d) Direct-P} \\
\end{tabular}
\caption{\textbf{Cost of forming P.} Instructions that form the codes of $P$ from a pair of scores $s$ in the softmax
warps, in PTX names, with their count per four scores on GB300 and VR200, for exp2 on the SFU, ExpCast of VC-Attention,
I-ExpCast, and Direct-P. Only (a) uses the SFU, for 2 of its 8 instructions. The constants $a$ and $c$
fold in the score scale, the row base, and the $P$ scale, and in Direct-P its linear map. (a) to (c) form E4M3 codes
and (d) E2M1 codes, which I-ExpCast forms with an E2M1 conversion as its last instruction.}
\label{fig:p-instructions}
\end{figure*}

We show that I-ExpCast overestimates a probability by at most 6\%, which moves the attention weights by less than a
centered score error of 0.06, and that it forms $P$ faster than every other way in the kernels at the error of the
exponential on the SFU.

\paragraph{Bit pattern.} An FP16 number holds a sign bit, an exponent field $E$ of five bits, and a mantissa field
$M$ of ten bits. For $1 \le E \le 30$, its value is $(1 + M/1024)\cdot 2^{E - 15}$, where 15 is the exponent bias of
the format, and the 16 bits of a positive number read as the integer $1024\,E + M$. Let $x = n + f$ with the integer
$n = \lfloor x \rfloor$ and $0 \le f < 1$. Then $2^x = 2^f \cdot 2^n$ with $1 \le 2^f < 2$, so its fields are
$E = n + 15$ and $M = 1024\,(2^f - 1)$, and its bit pattern is
\begin{equation}
\begin{colgather}
\operatorname{bits}_{\mathrm{FP16}}\big(2^x\big) = 1024\,(n + 15) + 1024\,(2^f - 1) \colpair
\approx 1024\,(n + 15) + 1024\,f = 1024\,(x + 15) ,
\end{colgather}
\label{eq:i-expcast-pattern}
\end{equation}
where the approximation replaces $2^f - 1$ by $f$. This is \cref{eq:i-expcast}. It holds for $-14 \le n \le 15$,
which covers every value of E4M3 and E2M1.

\paragraph{Approximation error.} The pattern $1024\,(x + 15)$ has the exponent field $n + 15$ and the mantissa field
$1024\,f$, so it is the FP16 number $2^n\,(1 + f)$, whereas $2^x = 2^n\, 2^f$. I-ExpCast thus forms the probability
times
\begin{equation}
c(f) = \frac{1 + f}{2^f} , \qquad 1 \le c(f) \le \frac{2}{e \ln 2} ,
\label{eq:i-expcast-error}
\end{equation}
which is exact at integer $x$ and largest at $f = 1/\ln 2 - 1$, an overestimate of up to 6\%. The kernel divides the
output by the row sum of the same $P$, so the weight of key $j$ is $p_j\, c_j / \sum_l p_l\, c_l$. A factor that all
keys share cancels in this ratio. Since every $c_j$ lies between 1 and 1.06, each weight changes by about 6\% at most.
By \cref{eq:softmax-linear}, this is the effect of a centered score error of about 0.06, far below the 1.0 at which the
softmax leaves its linear range (\cref{sec:der-softmax}). The codes of I-ExpCast therefore follow those of the
exponential and rounding over the whole range, subnormal codes included (\cref{fig:p-construction}).

\begin{table}[tbp]
\centering
\tablefont
\setlength{\tabcolsep}{3pt}
\caption{\textbf{Forming P in the kernels.} Attention time of a Wan2.2 call and relative L2 error (\%) of the
FP8-$PV$ and NVFP4-$PV$ kernels of \deco on GB300 with each way of forming $P$, as the mean over the hard layers and
over all layers. ExpCast is that of VC-Attention, SFU computes every exponential on the special-function unit, and
the polynomial of FlashAttention-4 replaces it for one eighth of the score pairs. $^\dagger$Direct-P is timed in its
released kernel against the NVFP4-$PV$ kernel of the same session. In every set of calls, some of its outputs are
not finite, so its errors are nan. I-ExpCast is the fastest way in both formats at the error of the exponentials on
the SFU.}
\label{tab:forming-p}
\input{docs/tables/forming_p}
\end{table}

\paragraph{Accuracy.} With only $P$ rounded, I-ExpCast stays close to the exponential followed by rounding in both
formats (\cref{tab:p-cast}). The kernel scales $P$ down by $2^T$ to leave room for the lazy rescale
(\cref{sec:rebase}). A larger threshold therefore moves more of the small probabilities into the subnormal range of
the format, where the ways of forming the codes differ most. For an E4M3 $P$, ExpCast makes the subnormal codes too
small. Over all layers of Wan2.2, its error grows from 1.02\% at $T = 0$ to 3.26\% at $T = 4$ and 8.88\% at $T = 8$,
whereas I-ExpCast has 0.68\%, 0.93\% and 2.72\%, against 0.40\%, 0.88\% and 2.72\% for the exponential and rounding.
For an E2M1 $P$, Direct-P~\citep{fa4fp4} forms the codes from a linear function of the score and has 5.06\% at every
threshold, whereas I-ExpCast has 1.90\% at $T = 4$, below the 2.09\% of the exponential and rounding. In the kernels
(\cref{tab:forming-p}), I-ExpCast has 2.80\% on the hard layers of Wan2.2 with FP8 $PV$, against 2.78\% for the
exponentials on the SFU, whereas ExpCast raises this error to 3.13\%.

\paragraph{Performance.} I-ExpCast takes 6 instructions per four scores and does not use the SFU
(\cref{fig:p-instructions}). The times of \cref{tab:forming-p} show two things. First, I-ExpCast saves a large part
of the kernel time. With every exponential on the SFU, the FP8-$PV$ kernel takes 21.5\% longer. Second, it saves more
with NVFP4 $PV$, where the SFU costs 34.3\%, because the faster 4-bit $PV$ MMA leaves the softmax a larger share of the
kernel. The polynomial of FlashAttention-4~\citep{fa4} does not close this gap. A score pair takes 13 instructions on
the polynomial against 4 on the SFU. With the polynomial on one eighth of the score pairs and the SFU on the others,
the FP8-$PV$ kernel takes 25.7\% longer than with I-ExpCast on GB300 and 36.1\% longer on VR200, more than with the SFU
alone. This is the slowdown that \cref{sec:kernel-p} refers to. ExpCast also keeps the exponentials off the SFU, but
its 7 instructions take 3.6\% longer. Direct-P needs only 4 instructions, but its released kernel takes 40.2\% longer than the NVFP4-$PV$ kernel of
\deco. In every set of calls, some of its outputs are not finite, which leaves its error unmeasured
(\cref{tab:forming-p}).

\subsection{Row sum and block sums}
\label{sec:kernel-ablations}

\begin{table}[tbp]
\centering
\tablefont
\caption{\textbf{Row sum of P.} Attention time of a Wan2.2 call and relative L2 error (\%) of the FP8-$PV$ kernel of
\deco on GB300 with each way of summing $P$, without the smoothing of $V$, as the mean over the hard layers and over
all layers. The sum in the $PV$ MMA is the fastest and does not cost accuracy.}
\label{tab:rowsum}
\input{docs/tables/row_sum}
\end{table}

\begin{table}[tbp]
\centering
\tablefont
\caption{\textbf{Threshold of the lazy rescale.} Attention time of a Wan2.2 call and relative L2 error (\%) of the
FP8-$PV$ and NVFP4-$PV$ kernels of \deco on GB300 at three thresholds $T$, as the mean over the hard layers and over all
layers. $T = 4$ keeps the error of $T = 0$ at nearly the time of $T = 8$.}
\label{tab:threshold}
\input{docs/tables/kernel_threshold}
\end{table}

We show that the row sum of $P$ in the $PV$ MMA does not cost accuracy against a sum on the ALU and saves 7.2\% of the
kernel time, and we give how the kernels form the block sums that the smoothing of $V$ needs. The ones column appended
to $V$ sums the $P$ that the $PV$ multiplies, in the format of the $PV$, so the denominator matches the rounded
numerator (\cref{sec:rowsum}). \Cref{tab:rowsum} compares it with a sum on the ALU of the softmax warps. The ALU sum
of the same E4M3 values gives the same error in every column, and takes 7.2\% longer. The row sum through the E4M3
$PV$ MMA thus does not lose accuracy and moves the sum from the softmax warps, which bound the kernel, to the tensor cores.
\Cref{tab:abl-perf} measures the same step before the lazy rescale, with 62.2 against 57.3~ms. Summing the FP16 values
before the rounding takes 5.4\% longer than the MMA and raises the error, from 2.95\% to 3.02\% on the hard layers of
Wan2.2, because this sum does not match the rounded $P$ of the numerator.

The smoothing of $V$ needs the sums of $P$ per block of keys (\cref{sec:num-pv}). With FP8 $PV$, each of eight
consecutive key tiles has its ones in a column of its own, and one BF16 MMA per eight key tiles adds the block means
to the output. With a 4-bit $P$, sums formed by the $PV$ MMA would omit the probabilities that round to zero. The
NVFP4-$PV$ kernels therefore sum the unquantized $P$ on the ALU of the softmax warps where their time allows. On RTX
PRO 6000, the MMAs bound the kernel, and it accumulates the row sum and all block sums in registers. On GB300 and
VR200, the softmax warps bound the kernel. These kernels sum three of the four pairs of 16-key blocks of a tile on the
ALU and leave the fourth pair to the $PV$ MMA. Summing it on the ALU as well would take about 6\% longer and lower the
error of the kernel by about 1\%.

\subsection{Threshold of the lazy rescale}
\label{sec:rebase}

We show that the threshold $T = 4$ keeps the error of a rescale at every tile and nearly the speed of $T = 8$. The
lazy rescale of \cref{sec:lazy-rescale} moves the row base only when a tile exceeds it by more than a factor of
$2^T$, so a probability reaches up to $2^T$ of the row base. FP8 kernels scale $P$ by $2^{8 - T}$ before the
conversion, which keeps it below the largest E4M3 value of 448. The smallest E4M3 subnormal $2^{-9}$ then stands for
\begin{equation}
2^{-9} / 2^{8 - T} = 2^{T - 17}
\label{eq:threshold-floor}
\end{equation}
of the row base, $2^{-13}$ at $T = 4$ and $2^{-9}$ at $T = 8$, the two values of \cref{sec:lazy-rescale}.
\Cref{tab:threshold} measures both kernels on GB300 at $T = 0$, 4 and 8. With an FP8 $P$, $T = 8$ raises the mean
error over all layers from 3.73\% to 4.50\% on Wan2.2 and from 4.40\% to 5.05\% on MiniMax-H3. $T = 0$ changes the
errors only in the second decimal but lengthens the attention by 2.5\%. The block scales of an NVFP4 $P$ follow the
block maxima, so its errors differ only in the second decimal between the thresholds, and $T = 0$ lengthens its
attention by 4.5\%. Both kernels therefore keep $T = 4$, which is within 0.5\% of the time of $T = 8$.

With the three optimizations of this appendix, the MMAs of a tile of the FP8-$PV$ kernel on GB300 take, at their
full rate, about nine tenths of the measured time of the tile, so the kernel runs close to the time of its MMAs
(\cref{sec:ablation}).

%% file: docs/tables/forming_p.tex
\begin{spreadtabular}{llccccc}
\toprule
\spancell{2}{1}{\tablehead{PV}} & \spancell{2}{1}{\tablehead{Exponential}} & \spancell{2}{1}{\tablehead{Time (ms)}} & \multicolumn{2}{c}{\tablehead{Wan2.2}} & \multicolumn{2}{c}{\tablehead{MiniMax-H3}} \\
\cmidrule(lr){4-5}\cmidrule(lr){6-7}
 &  &  & \tablehead{Hard} & \tablehead{All} & \tablehead{Hard} & \tablehead{All} \\
\midrule
\spancell{4}{0}{FP8} & \textbf{I-ExpCast (Ours)} & 57.83 & 2.80 & 3.73 & 6.32 & 4.40 \\
 & ExpCast & 60.05 (+3.6\%) & 3.13 & 4.82 & 6.87 & 5.08 \\
 & SFU & 70.24 (+21.5\%) & 2.78 & 3.69 & 6.27 & 4.37 \\
 & SFU and polynomial & 72.67 (+25.7\%) & 2.78 & 3.69 & 6.27 & 4.37 \\
\midrule
\spancell{3}{0}{NVFP4} & \textbf{I-ExpCast (Ours)} & 54.17 & 5.68 & 5.09 & 8.24 & 5.67 \\
 & SFU & 72.77 (+34.3\%) & 5.67 & 5.07 & 8.23 & 5.65 \\
 & Direct-P$^\dagger$ & 75.44 (+40.2\%) & nan & nan & nan & nan \\
\bottomrule
\end{spreadtabular}

%% file: docs/tables/row_sum.tex
\begin{spreadtabular}{lccccc}
\toprule
\spancell{2}{1}{\tablehead{Row sum of P}} & \spancell{2}{1}{\tablehead{Time (ms)}} & \multicolumn{2}{c}{\tablehead{Wan2.2}} & \multicolumn{2}{c}{\tablehead{MiniMax-H3}} \\
\cmidrule(lr){3-4}\cmidrule(lr){5-6}
 &  & \tablehead{Hard} & \tablehead{All} & \tablehead{Hard} & \tablehead{All} \\
\midrule
\textbf{$PV$ MMA (Ours)} & 54.93 & 2.95 & 3.75 & 6.37 & 4.44 \\
ALU, E4M3 values & 58.89 (+7.2\%) & 2.95 & 3.75 & 6.37 & 4.44 \\
ALU, FP16 values & 57.91 (+5.4\%) & 3.02 & 3.79 & 6.44 & 4.48 \\
\bottomrule
\end{spreadtabular}

%% file: docs/tables/kernel_threshold.tex
\begin{spreadtabular}{ccccccc}
\toprule
\spancell{2}{1}{\tablehead{PV}} & \spancell{2}{1}{\tablehead{$T$}} & \spancell{2}{1}{\tablehead{Time (ms)}} & \multicolumn{2}{c}{\tablehead{Wan2.2}} & \multicolumn{2}{c}{\tablehead{MiniMax-H3}} \\
\cmidrule(lr){4-5}\cmidrule(lr){6-7}
 &  &  & \tablehead{Hard} & \tablehead{All} & \tablehead{Hard} & \tablehead{All} \\
\midrule
\spancell{3}{0}{FP8} & 0 & 59.09 (+2.5\%) & 2.79 & 3.70 & 6.27 & 4.36 \\
 & \textbf{4 (Ours)} & 57.63 & 2.80 & 3.73 & 6.32 & 4.40 \\
 & 8 & 57.36 (-0.5\%) & 2.99 & 4.50 & 6.85 & 5.05 \\
\midrule
\spancell{3}{0}{NVFP4} & 0 & 56.63 (+4.5\%) & 5.67 & 5.07 & 8.24 & 5.65 \\
 & \textbf{4 (Ours)} & 54.19 & 5.68 & 5.09 & 8.24 & 5.67 \\
 & 8 & 54.01 (-0.3\%) & 5.70 & 5.13 & 8.28 & 5.70 \\
\bottomrule
\end{spreadtabular}

%% file: docs/floats/eval_e2e_vbench.tex
\begin{table*}[tbp]
\centering
\tablefont
\setlength{\tabcolsep}{3pt}
\caption{\textbf{End-to-end quality without a reference.} VBench subject consistency (S.C.), imaging quality (I.Q.),
and aesthetic quality (A.Q.) of the videos of \cref{tab:e2e}. Blue cells mark the best score within each
group with a 4-bit $QK$. The row of FlashAttention-4 holds the reference videos, and the row of cuDNN
those of the BF16 attention of cuDNN, the BF16 row of \cref{tab:e2e}. $^\dagger$Means reported by VC-Attention,
which does not report A.Q., read against its own BF16 row.}
\label{tab:e2e-vbench}
\input{docs/tables/eval_e2e_vbench}
\end{table*}

%% file: docs/tables/eval_e2e_vbench.tex
\begin{joinedtabular}{llYYYYYYYYYYYY}
\toprule
\spancell{2}{1}{\tablehead{Method}} & \spancell{2}{1}{\tablehead{QK/PV}} & \multicolumn{3}{c}{\tablehead{Wan2.2}} & \multicolumn{3}{c}{\tablehead{LongCat-Video}} & \multicolumn{3}{c}{\tablehead{HunyuanVideo-1.5}} & \multicolumn{3}{c}{\tablehead{MiniMax-H3}} \\
\cmidrule(lr){3-5}\cmidrule(lr){6-8}\cmidrule(lr){9-11}\cmidrule(lr){12-14}
 &  & \tablehead{S.C.$\uparrow$} & \tablehead{I.Q.$\uparrow$} & \tablehead{A.Q.$\uparrow$} & \tablehead{S.C.$\uparrow$} & \tablehead{I.Q.$\uparrow$} & \tablehead{A.Q.$\uparrow$} & \tablehead{S.C.$\uparrow$} & \tablehead{I.Q.$\uparrow$} & \tablehead{A.Q.$\uparrow$} & \tablehead{S.C.$\uparrow$} & \tablehead{I.Q.$\uparrow$} & \tablehead{A.Q.$\uparrow$} \\
\midrule
BF16, FlashAttention-4 & 16/16 & 0.930 & 0.700 & 0.669 & 0.938 & 0.683 & 0.678 & 0.922 & 0.672 & 0.659 & 0.905 & 0.686 & 0.647 \\
BF16$^\dagger$ & 16/16 & 0.932 & 0.703 & -- & 0.941 & 0.694 & -- & 0.933 & 0.680 & -- & 0.899 & 0.668 & -- \\
BF16, cuDNN & 16/16 & 0.931 & 0.700 & 0.668 & 0.938 & 0.684 & 0.679 & 0.924 & 0.670 & 0.658 & 0.904 & 0.686 & 0.647 \\
\qkpv\tablecite{tensorrtllm} & 16/8 & 0.931 & 0.700 & 0.672 & 0.937 & 0.682 & 0.680 & 0.923 & 0.668 & 0.656 & 0.906 & 0.686 & 0.647 \\
\midrule
Per-tensor FP8\tablecite{fp8formats} & \spancell{6}{0}{8/8} & 0.927 & 0.695 & 0.671 & 0.938 & 0.681 & 0.677 & 0.915 & 0.671 & 0.653 & 0.902 & 0.677 & 0.644 \\
MXFP8\tablecite{microscaling} &  & 0.927 & 0.696 & 0.669 & 0.937 & 0.681 & 0.677 & 0.917 & 0.666 & 0.650 & 0.904 & 0.675 & 0.647 \\
SageAttention2\tablecite{sage2}$^\dagger$ &  & 0.931 & 0.703 & -- & 0.941 & 0.697 & -- & 0.931 & 0.676 & -- & 0.897 & 0.671 & -- \\
\quad w/ FP8 QK &  & 0.931 & 0.698 & 0.674 & 0.938 & 0.681 & 0.679 & 0.920 & 0.670 & 0.653 & 0.904 & 0.676 & 0.645 \\
VC-Attention\tablecite{vcattention}$^\dagger$ &  & 0.932 & 0.705 & -- & 0.941 & 0.696 & -- & 0.931 & 0.677 & -- & 0.897 & 0.669 & -- \\
\quad w/ FP8 QK &  & 0.931 & 0.700 & 0.669 & 0.938 & 0.683 & 0.679 & 0.921 & 0.674 & 0.657 & 0.904 & 0.686 & 0.646 \\
\midrule
NVFP4 QK, FP8 PV\tablecite{tensorrtllm} & \spancell{3}{0}{4/8} & 0.918 & 0.689 & 0.657 & 0.935 & 0.681 & \best{0.676} & 0.899 & 0.645 & 0.594 & 0.899 & 0.668 & 0.636 \\
Attn-QAT\tablecite{attnqat} &  & 0.919 & 0.684 & 0.661 & 0.911 & 0.658 & 0.613 & 0.895 & 0.642 & 0.591 & 0.896 & 0.669 & 0.635 \\
\textbf{\deco (Ours)} &  & \best{0.931} & \best{0.699} & \best{0.669} & \best{0.937} & \best{0.683} & \best{0.676} & \best{0.921} & \best{0.675} & \best{0.659} & \best{0.902} & \best{0.687} & \best{0.647} \\
\midrule
NVFP4\tablecite{nvfp4pretrain} & \spancell{4}{0}{4/4} & 0.918 & 0.684 & 0.660 & 0.932 & 0.685 & 0.672 & 0.877 & 0.616 & 0.513 & 0.900 & 0.673 & 0.637 \\
SageAttention3\tablecite{sage3}$^\dagger$ &  & 0.925 & 0.697 & -- & 0.938 & 0.692 & -- & 0.927 & 0.677 & -- & 0.907 & 0.686 & -- \\
VC-Attention, 4-bit\tablecite{vcattention}$^\dagger$ &  & 0.931 & \best{0.707} & -- & \best{0.939} & \best{0.695} & -- & \best{0.934} & \best{0.679} & -- & \best{0.909} & 0.686 & -- \\
\textbf{\deco (Ours)} &  & \best{0.932} & 0.700 & \best{0.671} & \best{0.939} & 0.682 & \best{0.680} & 0.920 & 0.676 & \best{0.656} & 0.905 & \best{0.690} & \best{0.646} \\
\bottomrule
\end{joinedtabular}

%% file: docs/floats/eval_e2e_vbench_motion.tex
\begin{table*}[tbp]
\centering
\tablefont
\setlength{\tabcolsep}{2pt}
\caption{\textbf{End-to-end motion and prompt alignment without a reference.} VBench motion smoothness (M.S.),
dynamic degree (D.D., the fraction of videos rated dynamic), and overall consistency (O.C., the similarity of video and
prompt) of the runs of \cref{tab:e2e-vbench}. Blue cells mark the best M.S. and O.C. within each group with a 4-bit
$QK$, and the D.D. closest to that of BF16 FlashAttention-4.}
\label{tab:e2e-vbench-motion}
\input{docs/tables/eval_e2e_vbench_motion}
\end{table*}

%% file: docs/tables/eval_e2e_vbench_motion.tex
\begin{joinedtabular}{llYYYYYYYYYYYY}
\toprule
\spancell{2}{1}{\tablehead{Method}} & \spancell{2}{1}{\tablehead{QK/PV}} & \multicolumn{3}{c}{\tablehead{Wan2.2}} & \multicolumn{3}{c}{\tablehead{LongCat-Video}} & \multicolumn{3}{c}{\tablehead{HunyuanVideo-1.5}} & \multicolumn{3}{c}{\tablehead{MiniMax-H3}} \\
\cmidrule(lr){3-5}\cmidrule(lr){6-8}\cmidrule(lr){9-11}\cmidrule(lr){12-14}
 &  & \tablehead{M.S.$\uparrow$} & \tablehead{D.D.} & \tablehead{O.C.$\uparrow$} & \tablehead{M.S.$\uparrow$} & \tablehead{D.D.} & \tablehead{O.C.$\uparrow$} & \tablehead{M.S.$\uparrow$} & \tablehead{D.D.} & \tablehead{O.C.$\uparrow$} & \tablehead{M.S.$\uparrow$} & \tablehead{D.D.} & \tablehead{O.C.$\uparrow$} \\
\midrule
BF16, FlashAttention-4 & 16/16 & 0.9766 & 0.61 & 0.268 & 0.9834 & 0.80 & 0.262 & 0.9893 & 0.74 & 0.266 & 0.9887 & 0.80 & 0.272 \\
BF16, cuDNN & 16/16 & 0.9765 & 0.62 & 0.268 & 0.9834 & 0.80 & 0.263 & 0.9894 & 0.75 & 0.265 & 0.9887 & 0.81 & 0.272 \\
\qkpv\tablecite{tensorrtllm} & 16/8 & 0.9765 & 0.62 & 0.266 & 0.9832 & 0.81 & 0.263 & 0.9889 & 0.75 & 0.263 & 0.9888 & 0.78 & 0.273 \\
\midrule
Per-tensor FP8\tablecite{fp8formats} & \spancell{4}{0}{8/8} & 0.9763 & 0.61 & 0.268 & 0.9831 & 0.80 & 0.262 & 0.9866 & 0.73 & 0.267 & 0.9884 & 0.76 & 0.270 \\
MXFP8\tablecite{microscaling} &  & 0.9764 & 0.60 & 0.269 & 0.9830 & 0.81 & 0.262 & 0.9872 & 0.75 & 0.267 & 0.9881 & 0.77 & 0.269 \\
SageAttention2, FP8 QK\tablecite{sage2} &  & 0.9770 & 0.61 & 0.269 & 0.9832 & 0.81 & 0.262 & 0.9884 & 0.75 & 0.263 & 0.9882 & 0.78 & 0.270 \\
VC-Attention, FP8 QK\tablecite{vcattention} &  & 0.9765 & 0.60 & 0.268 & 0.9836 & 0.80 & 0.263 & 0.9890 & 0.72 & 0.265 & 0.9887 & 0.81 & 0.271 \\
\midrule
NVFP4 QK, FP8 PV\tablecite{tensorrtllm} & \spancell{3}{0}{4/8} & 0.9758 & \best{0.60} & \best{0.268} & 0.9823 & 0.82 & 0.262 & 0.9812 & \best{0.74} & \best{0.268} & 0.9872 & 0.69 & 0.269 \\
Attn-QAT\tablecite{attnqat} &  & 0.9765 & \best{0.62} & \best{0.268} & 0.9737 & 0.60 & 0.253 & 0.9787 & 0.80 & 0.265 & 0.9879 & 0.76 & 0.270 \\
\textbf{\deco (Ours)} &  & \best{0.9767} & \best{0.60} & 0.266 & \best{0.9834} & \best{0.80} & \best{0.263} & \best{0.9888} & 0.71 & 0.264 & \best{0.9885} & \best{0.80} & \best{0.271} \\
\midrule
NVFP4\tablecite{nvfp4pretrain} & \spancell{2}{0}{4/4} & 0.9728 & \best{0.60} & \best{0.271} & 0.9742 & 0.84 & 0.261 & 0.9700 & \best{0.74} & 0.243 & \best{0.9885} & 0.76 & 0.268 \\
\textbf{\deco (Ours)} &  & \best{0.9766} & 0.63 & 0.268 & \best{0.9833} & \best{0.80} & \best{0.262} & \best{0.9881} & 0.75 & \best{0.265} & \best{0.9885} & \best{0.82} & \best{0.270} \\
\bottomrule
\end{joinedtabular}

%% file: docs/floats/reported_int8.tex
\begin{table}[tbp]
\centering
\tablefont
\setlength{\tabcolsep}{3pt}
\caption{\textbf{Reported and reproduced INT8 rows.} PSNR (dB) of SageAttention2 and VC-Attention with INT8 $QK$ as
VC-Attention reports it and as our runs on B200 measure it over the 100 prompts. Our runs reproduce the reported
values to within about 1.4~dB.}
\label{tab:reported-int8}
\input{docs/tables/reported_int8}
\end{table}

%% file: docs/tables/reported_int8.tex
\begin{spreadtabular}{lcccc}
\toprule
\spancell{2}{1}{\tablehead{Model}} & \multicolumn{2}{c}{\tablehead{SageAttention2}} & \multicolumn{2}{c}{\tablehead{VC-Attention}} \\
\cmidrule(lr){2-3}\cmidrule(lr){4-5}
 & \tablehead{reported} & \tablehead{ours} & \tablehead{reported} & \tablehead{ours} \\
\midrule
Wan2.2 & 20.30 & 20.08 & 20.50 & 19.78 \\
LongCat-Video & 23.10 & 22.93 & 23.80 & 25.21 \\
HunyuanVideo-1.5 & 15.60 & 15.88 & 17.50 & 17.41 \\
MiniMax-H3 & 19.90 & 19.36 & 20.20 & 19.39 \\
\bottomrule
\end{spreadtabular}

%% file: docs/floats/eval_e2e_paired.tex
\begin{table*}[tbp]
\centering
\scriptsize
\setlength{\tabcolsep}{3pt}
\caption{\textbf{Paired end-to-end differences with FP8 PV.} PSNR (dB) and LPIPS of each method minus those of \deco
with FP8 $PV$ on the same prompt, as the mean over the 100 prompts with the half width of its 95\%
confidence interval. Beyond its interval, a bold entry is better than \deco and a grey entry worse. A black entry
has an interval that contains zero. \Cref{tab:e2e} holds the absolute values.}
\label{tab:e2e-paired}
\input{docs/tables/eval_e2e_paired}
\end{table*}

%% file: docs/tables/eval_e2e_paired.tex
\begin{spreadtabular}{llcccccccc}
\toprule
\spancell{2}{1}{\tablehead{Method}} & \spancell{2}{1}{\tablehead{QK/PV}} & \multicolumn{2}{c}{\tablehead{Wan2.2}} & \multicolumn{2}{c}{\tablehead{LongCat-Video}} & \multicolumn{2}{c}{\tablehead{HunyuanVideo-1.5}} & \multicolumn{2}{c}{\tablehead{MiniMax-H3}} \\
\cmidrule(lr){3-4}\cmidrule(lr){5-6}\cmidrule(lr){7-8}\cmidrule(lr){9-10}
 &  & \tablehead{$\Delta$PSNR} & \tablehead{$\Delta$LPIPS} & \tablehead{$\Delta$PSNR} & \tablehead{$\Delta$LPIPS} & \tablehead{$\Delta$PSNR} & \tablehead{$\Delta$LPIPS} & \tablehead{$\Delta$PSNR} & \tablehead{$\Delta$LPIPS} \\
\midrule
\qkpv\tablecite{tensorrtllm} & 16/8 & \textbf{+1.24 $\pm$ 0.58} & \textbf{-0.019 $\pm$ 0.017} & \textbf{+0.97 $\pm$ 0.64} & -0.008 $\pm$ 0.013 & +0.07 $\pm$ 0.40 & -0.003 $\pm$ 0.018 & \textbf{+2.97 $\pm$ 0.47} & \textbf{-0.112 $\pm$ 0.017} \\
\midrule
Per-tensor FP8\tablecite{fp8formats} & \spancell{4}{0}{8/8} & \textcolor{black!45}{-1.47 $\pm$ 0.47} & \textcolor{black!45}{+0.067 $\pm$ 0.017} & \textcolor{black!45}{-2.02 $\pm$ 0.65} & \textcolor{black!45}{+0.049 $\pm$ 0.015} & \textcolor{black!45}{-1.08 $\pm$ 0.37} & \textcolor{black!45}{+0.054 $\pm$ 0.017} & -0.09 $\pm$ 0.37 & \textcolor{black!45}{+0.027 $\pm$ 0.015} \\
MXFP8\tablecite{microscaling} &  & \textcolor{black!45}{-1.41 $\pm$ 0.47} & \textcolor{black!45}{+0.061 $\pm$ 0.017} & \textcolor{black!45}{-2.03 $\pm$ 0.55} & \textcolor{black!45}{+0.047 $\pm$ 0.014} & \textcolor{black!45}{-1.91 $\pm$ 0.38} & \textcolor{black!45}{+0.096 $\pm$ 0.019} & \textcolor{black!45}{-0.37 $\pm$ 0.35} & \textcolor{black!45}{+0.035 $\pm$ 0.015} \\
SageAttention2, FP8 QK\tablecite{sage2} &  & \textcolor{black!45}{-1.97 $\pm$ 0.50} & \textcolor{black!45}{+0.077 $\pm$ 0.019} & \textcolor{black!45}{-1.72 $\pm$ 0.58} & \textcolor{black!45}{+0.043 $\pm$ 0.015} & \textcolor{black!45}{-1.23 $\pm$ 0.35} & \textcolor{black!45}{+0.055 $\pm$ 0.017} & -0.24 $\pm$ 0.36 & \textcolor{black!45}{+0.026 $\pm$ 0.015} \\
VC-Attention, FP8 QK\tablecite{vcattention} &  & \textcolor{black!45}{-0.90 $\pm$ 0.48} & \textcolor{black!45}{+0.034 $\pm$ 0.017} & -0.30 $\pm$ 0.60 & +0.013 $\pm$ 0.017 & +0.14 $\pm$ 0.40 & -0.013 $\pm$ 0.017 & \textbf{+0.48 $\pm$ 0.38} & \textbf{-0.020 $\pm$ 0.017} \\
\midrule
NVFP4 QK, FP8 PV\tablecite{tensorrtllm} & \spancell{2}{0}{4/8} & \textcolor{black!45}{-4.09 $\pm$ 0.56} & \textcolor{black!45}{+0.201 $\pm$ 0.019} & \textcolor{black!45}{-4.68 $\pm$ 0.63} & \textcolor{black!45}{+0.143 $\pm$ 0.017} & \textcolor{black!45}{-4.08 $\pm$ 0.42} & \textcolor{black!45}{+0.226 $\pm$ 0.018} & \textcolor{black!45}{-2.75 $\pm$ 0.41} & \textcolor{black!45}{+0.155 $\pm$ 0.018} \\
Attn-QAT\tablecite{attnqat} &  & \textcolor{black!45}{-4.09 $\pm$ 0.54} & \textcolor{black!45}{+0.204 $\pm$ 0.019} & \textcolor{black!45}{-5.68 $\pm$ 0.70} & \textcolor{black!45}{+0.258 $\pm$ 0.017} & \textcolor{black!45}{-5.82 $\pm$ 0.50} & \textcolor{black!45}{+0.316 $\pm$ 0.021} & \textcolor{black!45}{-2.06 $\pm$ 0.35} & \textcolor{black!45}{+0.122 $\pm$ 0.016} \\
\bottomrule
\end{spreadtabular}

%% file: docs/floats/eval_e2e_paired_pv4.tex
\begin{table*}[tbp]
\centering
\scriptsize
\setlength{\tabcolsep}{3pt}
\caption{\textbf{Paired end-to-end differences with NVFP4 PV.} PSNR (dB) and LPIPS of each method minus those of
\deco with NVFP4 $PV$, as in \cref{tab:e2e-paired}. SageAttention3 and 4-bit VC-Attention are paired on prompts 0 to
7. \Cref{tab:e2e} holds the absolute values.}
\label{tab:e2e-paired-pv4}
\input{docs/tables/eval_e2e_paired_pv4}
\end{table*}

%% file: docs/tables/eval_e2e_paired_pv4.tex
\begin{spreadtabular}{llcccccccc}
\toprule
\spancell{2}{1}{\tablehead{Method}} & \spancell{2}{1}{\tablehead{QK/PV}} & \multicolumn{2}{c}{\tablehead{Wan2.2}} & \multicolumn{2}{c}{\tablehead{LongCat-Video}} & \multicolumn{2}{c}{\tablehead{HunyuanVideo-1.5}} & \multicolumn{2}{c}{\tablehead{MiniMax-H3}} \\
\cmidrule(lr){3-4}\cmidrule(lr){5-6}\cmidrule(lr){7-8}\cmidrule(lr){9-10}
 &  & \tablehead{$\Delta$PSNR} & \tablehead{$\Delta$LPIPS} & \tablehead{$\Delta$PSNR} & \tablehead{$\Delta$LPIPS} & \tablehead{$\Delta$PSNR} & \tablehead{$\Delta$LPIPS} & \tablehead{$\Delta$PSNR} & \tablehead{$\Delta$LPIPS} \\
\midrule
SageAttention2, FP8 QK\tablecite{sage2} & \spancell{2}{0}{8/8} & +0.09 $\pm$ 0.43 & +0.001 $\pm$ 0.017 & -0.35 $\pm$ 0.40 & +0.013 $\pm$ 0.014 & -0.21 $\pm$ 0.38 & +0.012 $\pm$ 0.018 & +0.26 $\pm$ 0.29 & +0.003 $\pm$ 0.013 \\
VC-Attention, FP8 QK\tablecite{vcattention} &  & \textbf{+1.17 $\pm$ 0.51} & \textbf{-0.042 $\pm$ 0.020} & \textbf{+1.07 $\pm$ 0.60} & -0.017 $\pm$ 0.017 & \textbf{+1.16 $\pm$ 0.39} & \textbf{-0.056 $\pm$ 0.018} & \textbf{+0.97 $\pm$ 0.41} & \textbf{-0.042 $\pm$ 0.017} \\
\midrule
NVFP4\tablecite{nvfp4pretrain} & \spancell{3}{0}{4/4} & \textcolor{black!45}{-4.95 $\pm$ 0.61} & \textcolor{black!45}{+0.253 $\pm$ 0.023} & \textcolor{black!45}{-8.05 $\pm$ 0.61} & \textcolor{black!45}{+0.301 $\pm$ 0.020} & \textcolor{black!45}{-6.90 $\pm$ 0.52} & \textcolor{black!45}{+0.378 $\pm$ 0.022} & \textcolor{black!45}{-2.88 $\pm$ 0.37} & \textcolor{black!45}{+0.148 $\pm$ 0.017} \\
SageAttention3\tablecite{sage3} &  & \textcolor{black!45}{-3.77 $\pm$ 2.44} & \textcolor{black!45}{+0.119 $\pm$ 0.037} & \textcolor{black!45}{-6.15 $\pm$ 2.00} & \textcolor{black!45}{+0.180 $\pm$ 0.048} & \textcolor{black!45}{-5.62 $\pm$ 1.71} & \textcolor{black!45}{+0.310 $\pm$ 0.089} & \textcolor{black!45}{-1.95 $\pm$ 1.27} & \textcolor{black!45}{+0.093 $\pm$ 0.060} \\
VC-Attention, 4-bit\tablecite{vcattention} &  & -1.21 $\pm$ 1.78 & +0.053 $\pm$ 0.069 & \textcolor{black!45}{-3.67 $\pm$ 1.50} & \textcolor{black!45}{+0.105 $\pm$ 0.046} & -0.57 $\pm$ 1.12 & +0.029 $\pm$ 0.058 & \textcolor{black!45}{-1.22 $\pm$ 1.14} & \textcolor{black!45}{+0.079 $\pm$ 0.062} \\
\bottomrule
\end{spreadtabular}

%% file: docs/floats/eval_error_full.tex
\begin{table*}[tbp]
\centering
\tablefont
\caption{\textbf{Accuracy of every kernel.} Relative L2 error (\%) of every method, as mean / maximum over the hard
layers and over all layers. The hard layers are captured at three times as many steps as all layers, so the two
columns hold different calls. SageAttention2 and VC-Attention use their default INT8 $QK$ on B200, and their rows w/ FP8
$QK$ run on GB300. Blue marks the lowest error of each 4-bit-$QK$ group.}
\label{tab:kernel-error-full}
\input{docs/tables/eval_error_full}
\end{table*}

%% file: docs/tables/eval_error_full.tex
\begin{spreadtabular}{ll!{\extracolsep{0pt}}cccc}
\toprule
\spancell{2}{1}{\tablehead{Method}} & \spancell{2}{1}{\tablehead{QK/PV}} & \multicolumn{2}{c}{\tablehead{Wan2.2}} & \multicolumn{2}{c}{\tablehead{MiniMax-H3}} \\
\cmidrule(lr){3-4}\cmidrule(lr){5-6}
 &  & \tablehead{Hard} & \tablehead{All} & \tablehead{Hard} & \tablehead{All} \\
\midrule
BF16 & 16/16 & 0.15 / 0.17 & 0.17 / 0.19 & 0.17 / 0.20 & 0.17 / 0.19 \\
\qkpv\tablecite{tensorrtllm} & 16/8 & 1.90 / 2.28 & 1.20 / 2.27 & 1.76 / 2.17 & 1.27 / 2.03 \\
\midrule
Per-tensor FP8\tablecite{fp8formats} & \spancell{6}{0}{8/8} & 20.29 / 35.18 & 5.29 / 29.69 & 21.42 / 56.52 & 5.14 / 46.44 \\
MXFP8\tablecite{microscaling} &  & 18.21 / 33.28 & 4.81 / 33.16 & 21.87 / 57.12 & 4.95 / 48.17 \\
SageAttention2\tablecite{sage2} &  & 2.18 / 2.79 & 1.37 / 2.43 & 3.32 / 5.42 & 1.52 / 4.68 \\
\quad w/ FP8 QK &  & 13.03 / 26.29 & 4.10 / 26.64 & 21.23 / 56.42 & 4.55 / 44.85 \\
VC-Attention\tablecite{vcattention} &  & 1.83 / 2.32 & 1.57 / 2.82 & 1.93 / 2.43 & 1.46 / 2.15 \\
\quad w/ FP8 QK &  & 3.67 / 5.78 & 2.64 / 5.25 & 6.65 / 11.88 & 2.85 / 9.51 \\
\midrule
NVFP4 QK, FP8 PV\tablecite{tensorrtllm} & \spancell{3}{0}{4/8} & 30.56 / 47.44 & 10.57 / 47.19 & 28.92 / 59.90 & 10.69 / 50.58 \\
Attn-QAT\tablecite{attnqat} &  & 30.19 / 46.41 & 10.60 / 45.99 & 28.25 / 63.30 & 11.10 / 53.43 \\
\textbf{\deco (Ours)} &  & \cellcolor{bestfill}2.80 / 3.74 & \cellcolor{bestfill}3.73 / 5.50 & \cellcolor{bestfill}6.32 / 10.52 & \cellcolor{bestfill}4.40 / 7.79 \\
\midrule
NVFP4\tablecite{nvfp4pretrain} & \spancell{4}{0}{4/4} & 31.46 / 47.63 & 12.97 / 47.18 & 30.02 / 64.53 & 13.87 / 54.65 \\
SageAttention3\tablecite{sage3} &  & 21.86 / 36.02 & 9.19 / 34.90 & 24.07 / 53.67 & 8.96 / 45.02 \\
VC-Attention, 4-bit\tablecite{vcattention} &  & 21.06 / 35.67 & 8.52 / 34.54 & 23.46 / 53.48 & 8.05 / 44.86 \\
\textbf{\deco (Ours)} &  & \cellcolor{bestfill}5.68 / 6.79 & \cellcolor{bestfill}5.09 / 7.81 & \cellcolor{bestfill}8.24 / 11.37 & \cellcolor{bestfill}5.67 / 9.36 \\
\bottomrule
\end{spreadtabular}

%% file: docs/floats/eval_time_full.tex
\begin{table*}[tbp]
\centering
\tablefont
\setlength{\tabcolsep}{3pt}
\caption{\textbf{Kernel speed on GB300 and VR200.} Attention time (ms) of a Wan2.2 and a MiniMax-H3 call, with the
speedup over BF16 attention, the method's preparation outside the kernel, and the backend of each kernel, written GB300 / VR200 where the two GPUs differ and Wan2.2, MiniMax-H3 where the
two models differ. Each speedup is over the BF16 attention of the kernel's own timing session, whose time is
within 2\% of the BF16 row. A dash marks a method without a kernel for the GPU. Blue marks the shortest attention time of each 4-bit-$QK$ group.}
\label{tab:time-full}
\input{docs/tables/eval_time_full}
\end{table*}

%% file: docs/tables/eval_time_full.tex
\begin{spreadtabular}{lll!{\extracolsep{0pt}}cccccccc}
\toprule
\spancell{3}{2}{\tablehead{Method}} & \spancell{3}{2}{\tablehead{QK/PV}} & \spancell{3}{2}{\tablehead{Backend}} & \multicolumn{4}{c}{\tablehead{GB300}} & \multicolumn{4}{c}{\tablehead{VR200}} \\
\cmidrule(lr){4-7}\cmidrule(lr){8-11}
 &  &  & \multicolumn{2}{c}{\tablehead{Wan2.2}} & \multicolumn{2}{c}{\tablehead{MiniMax-H3}} & \multicolumn{2}{c}{\tablehead{Wan2.2}} & \multicolumn{2}{c}{\tablehead{MiniMax-H3}} \\
\cmidrule(lr){4-5}\cmidrule(lr){6-7}\cmidrule(lr){8-9}\cmidrule(lr){10-11}
 &  &  & \tablehead{attention} & \tablehead{prep.} & \tablehead{attention} & \tablehead{prep.} & \tablehead{attention} & \tablehead{prep.} & \tablehead{attention} & \tablehead{prep.} \\
\midrule
BF16 & 16/16 & FA4, cuDNN / FA4 & 122.5 (1.00$\times$) & 0.00 & 83.5 (1.00$\times$) & 0.00 & 72.6 (1.00$\times$) & 0.00 & 50.5 (1.00$\times$) & 0.00 \\
\qkpv\tablecite{tensorrtllm} & 16/8 & TRT-LLM / FlashInfer & 102.8 (1.19$\times$) & 0.65 & 69.8 (1.20$\times$) & 0.46 & 59.8 (1.22$\times$) & 0.59 & 41.9 (1.21$\times$) & 0.43 \\
\midrule
Per-tensor FP8\tablecite{fp8formats} & \spancell{4}{0}{8/8} & FlashInfer / cuDNN & 79.2 (1.55$\times$) & 1.95 & 52.7 (1.58$\times$) & 1.40 & 43.7 (1.66$\times$) & 1.78 & 29.0 (1.74$\times$) & 1.30 \\
MXFP8\tablecite{microscaling} &  & cuDNN & 82.7 (1.50$\times$) & 3.37 & 54.5 (1.53$\times$) & 2.32 & 44.2 (1.64$\times$) & 3.51 & 29.5 (1.72$\times$) & 2.41 \\
SageAttention2, FP8 QK\tablecite{sage2} &  & TRT-LLM & 103.1 (1.21$\times$) & 1.71 & 67.9 (1.23$\times$) & 1.15 & 52.6 (1.38$\times$) & 1.99 & 34.8 (1.45$\times$) & 1.36 \\
VC-Attention, FP8 QK\tablecite{vcattention} &  & FlashInfer & 83.0 (1.48$\times$) & 3.63 & 51.0 (1.65$\times$) & 2.44 & 48.1 (1.51$\times$) & 2.21 & 30.7 (1.65$\times$) & 1.46 \\
\midrule
NVFP4 QK, FP8 PV\tablecite{tensorrtllm} & \spancell{3}{0}{4/8} & TRT-LLM & 83.7 (1.49$\times$) & 11.53 & 56.3 (1.48$\times$) & 7.90 & 44.9 (1.62$\times$) & 12.59 & 31.3 (1.61$\times$) & 8.67 \\
Attn-QAT\tablecite{attnqat} &  & Attn-QAT & 77.5 (1.60$\times$) & 0.96 & 51.1 (1.62$\times$) & 0.66 & -- & -- & -- & -- \\
\textbf{\deco (Ours)} &  & ours & \cellcolor{bestfill}57.5 (2.13$\times$) & 1.55 & \cellcolor{bestfill}37.1 (2.25$\times$) & 1.09 & \cellcolor{bestfill}29.2 (2.49$\times$) & 1.05 & \cellcolor{bestfill}19.6 (2.58$\times$) & 0.70 \\
\midrule
NVFP4\tablecite{nvfp4pretrain} & \spancell{2}{0}{4/4} & FlashInfer & 74.8 (1.66$\times$) & 0.99 & 49.9 (1.67$\times$) & 0.64 & -- & -- & -- & -- \\
\textbf{\deco (Ours)} &  & ours & \cellcolor{bestfill}54.7 (2.28$\times$) & 1.41 & \cellcolor{bestfill}36.1 (2.32$\times$) & 1.00 & \cellcolor{bestfill}32.0 (2.27$\times$) & 1.02 & \cellcolor{bestfill}21.3 (2.38$\times$) & 0.74 \\
\bottomrule
\end{spreadtabular}

%% file: docs/floats/ablation_fixed_clock.tex
\begin{table}[tbp]
\centering
\tablefont
\caption{\textbf{Softmax optimizations at a fixed clock.} Time of the FP8-$PV$ kernel of \deco and of the FP8-$QK$
reference on a Wan2.2 call on GB300 with the SM clock fixed at 1500~MHz, without the four optimizations of
\cref{tab:abl-perf} and with all of them. At a fixed clock the four optimizations speed our kernel up 1.67$\times$.}
\label{tab:abl-fixed-clock}
\input{docs/tables/ablation_fixed_clock}
\end{table}

%% file: docs/tables/ablation_fixed_clock.tex
\begin{spreadtabular}{lcc}
\toprule
\tablehead{Kernel} & \tablehead{\makecell{Time\\(ms)}} & \tablehead{\makecell{Speedup of the\\optimizations}} \\
\midrule
Our kernel, compensation only & 101.1 & -- \\
\quad + All four optimizations & 60.5 & 1.67$\times$ \\
\midrule
FP8-QK reference & 98.9 & -- \\
\quad + All four optimizations & 69.4 & 1.43$\times$ \\
\bottomrule
\end{spreadtabular}

%% file: docs/floats/eval_time_rtx.tex
\begin{table}[tbp]
\centering
\tablefont
\caption{\textbf{Kernel speed on RTX PRO 6000.} Attention time (ms) with the speedup over BF16 attention. Blue marks
the shortest time.}
\label{tab:time-rtx}
\input{docs/tables/eval_time_rtx}
\end{table}

%% file: docs/tables/eval_time_rtx.tex
\begin{spreadtabular}{ll!{\extracolsep{0pt}}cc}
\toprule
\tablehead{Method} & \tablehead{QK/PV} & \tablehead{Wan2.2} & \tablehead{MiniMax-H3} \\
\midrule
BF16 & 16/16 & 660.3 (1.00$\times$) & 444.0 (1.00$\times$) \\
\midrule
SageAttention3\tablecite{sage3} & \spancell{3}{0}{4/4} & 253.0 (2.61$\times$) & 165.3 (2.69$\times$) \\
VC-Attention, 4-bit\tablecite{vcattention} &  & 262.1 (2.52$\times$) & 170.1 (2.61$\times$) \\
\textbf{\deco (Ours)} &  & \best{234.0 (2.82$\times$)} & \best{155.1 (2.86$\times$)} \\
\bottomrule
\end{spreadtabular}

%% file: docs/floats/trtllm_e2e.tex
\begin{table}[tbp]
\centering
\tablefont
\caption{\textbf{End-to-end speed.} Time (s) of a whole generation in TensorRT-LLM VisualGen with the speedup over BF16
attention on the same GPU. Blue marks the shortest time on each GPU.}
\label{tab:trtllm-e2e}
\input{docs/tables/trtllm_e2e}
\end{table}

%% file: docs/tables/trtllm_e2e.tex
\begin{spreadtabular}{l!{\extracolsep{0pt}}cc}
\toprule
\tablehead{Method} & \tablehead{Wan2.2} & \tablehead{MiniMax-H3} \\
\midrule
\multicolumn{3}{l}{\textit{GB300}} \\
BF16 & 302.7 (1.00$\times$) & 51.9 (1.00$\times$) \\
VC-Attention, FP8 QK & 237.8 (1.27$\times$) & 41.3 (1.26$\times$) \\
SageAttention2, FP8 QK & 270.4 (1.12$\times$) & 46.2 (1.12$\times$) \\
\textbf{\cfgmain (Ours)} & \cellcolor{bestfill}194.7 (1.55$\times$) & \cellcolor{bestfill}33.6 (1.54$\times$) \\
\midrule
\multicolumn{3}{l}{\textit{VR200}} \\
BF16 & 181.0 (1.00$\times$) & 32.2 (1.00$\times$) \\
VC-Attention, FP8 QK & 142.9 (1.27$\times$) & 24.8 (1.30$\times$) \\
SageAttention2, FP8 QK & 149.2 (1.21$\times$) & 25.4 (1.27$\times$) \\
\textbf{\cfgmain (Ours)} & \cellcolor{bestfill}108.0 (1.68$\times$) & \cellcolor{bestfill}19.1 (1.69$\times$) \\
\midrule
\multicolumn{3}{l}{\textit{RTX PRO 6000}} \\
BF16 & 1619.1 (1.00$\times$) & 269.7 (1.00$\times$) \\
SageAttention3 & 983.9 (1.65$\times$) & 168.3 (1.60$\times$) \\
VC-Attention, 4-bit & 1111.7 (1.46$\times$) & 188.8 (1.43$\times$) \\
\textbf{\cfgfour (Ours)} & \cellcolor{bestfill}900.8 (1.80$\times$) & \cellcolor{bestfill}151.3 (1.78$\times$) \\
\bottomrule
\end{spreadtabular}

%% file: results/appendix/videos.tex
\pgfkeyssetvalue{/appendix/blocks}{gb300, rtx}
\pgfkeyssetvalue{/appendix/block/gb300/rows}{reference, deco.fp8pv.vs128, trtllm.sage.fp8, vc.qk8}
\pgfkeyssetvalue{/appendix/block/gb300/gpu}{GB300}
\pgfkeyssetvalue{/appendix/block/gb300/reference}{reference}
\pgfkeyssetvalue{/appendix/block/rtx/rows}{rtx.reference, deco.nvfp4pv.vs16, sage3.nvfp4, vc4.vsmooth}
\pgfkeyssetvalue{/appendix/block/rtx/gpu}{RTX PRO 6000}
\pgfkeyssetvalue{/appendix/block/rtx/reference}{rtx.reference}
\pgfkeyssetvalue{/appendix/label/reference}{BF16 FlashAttention-4}
\pgfkeyssetvalue{/appendix/label/deco.fp8pv.vs128}{\cfgmain}
\pgfkeyssetvalue{/appendix/label/trtllm.sage.fp8}{SageAttention2, FP8 QK}
\pgfkeyssetvalue{/appendix/label/vc.qk8}{VC-Attention, FP8 QK}
\pgfkeyssetvalue{/appendix/label/rtx.reference}{BF16 FlashInfer FA2}
\pgfkeyssetvalue{/appendix/label/deco.nvfp4pv.vs16}{\cfgfour}
\pgfkeyssetvalue{/appendix/label/sage3.nvfp4}{SageAttention3}
\pgfkeyssetvalue{/appendix/label/vc4.vsmooth}{VC-Attention, 4-bit}
\pgfkeyssetvalue{/appendix/wan2.2-t2v-a14b/prompt}{90}
\pgfkeyssetvalue{/appendix/wan2.2-t2v-a14b/frames}{0, 20, 40, 60, and 80}
\pgfkeyssetvalue{/appendix/wan2.2-t2v-a14b/count}{81}
\pgfkeyssetvalue{/appendix/wan2.2-t2v-a14b/text}{A cinematic wide-angle portrait of a man with his face illuminated by the warm glow of a TV screen. \ldots{} A wide-angle shot capturing the man's intense gaze and the warm ambiance of the room.}
\pgfkeyssetvalue{/appendix/wan2.2-t2v-a14b/psnr/reference}{reference}
\pgfkeyssetvalue{/appendix/wan2.2-t2v-a14b/psnr/deco.fp8pv.vs128}{24.70~dB}
\pgfkeyssetvalue{/appendix/wan2.2-t2v-a14b/psnr/trtllm.sage.fp8}{18.09~dB}
\pgfkeyssetvalue{/appendix/wan2.2-t2v-a14b/psnr/vc.qk8}{17.17~dB}
\pgfkeyssetvalue{/appendix/wan2.2-t2v-a14b/psnr/rtx.reference}{reference}
\pgfkeyssetvalue{/appendix/wan2.2-t2v-a14b/psnr/deco.nvfp4pv.vs16}{19.94~dB}
\pgfkeyssetvalue{/appendix/wan2.2-t2v-a14b/psnr/sage3.nvfp4}{17.40~dB}
\pgfkeyssetvalue{/appendix/wan2.2-t2v-a14b/psnr/vc4.vsmooth}{17.62~dB}
\pgfkeyssetvalue{/appendix/longcat-video/prompt}{48}
\pgfkeyssetvalue{/appendix/longcat-video/frames}{0, 23, 46, 69, and 92}
\pgfkeyssetvalue{/appendix/longcat-video/count}{93}
\pgfkeyssetvalue{/appendix/longcat-video/text}{A dynamic scene captured in the style of a vibrant food photography, showcasing a chef skillfully chopping onions in a bustling kitchen. \ldots{} A medium shot with the chef at the center, capturing the intensity of his work.}
\pgfkeyssetvalue{/appendix/longcat-video/psnr/reference}{reference}
\pgfkeyssetvalue{/appendix/longcat-video/psnr/deco.fp8pv.vs128}{17.63~dB}
\pgfkeyssetvalue{/appendix/longcat-video/psnr/trtllm.sage.fp8}{13.54~dB}
\pgfkeyssetvalue{/appendix/longcat-video/psnr/vc.qk8}{11.99~dB}
\pgfkeyssetvalue{/appendix/longcat-video/psnr/rtx.reference}{reference}
\pgfkeyssetvalue{/appendix/longcat-video/psnr/deco.nvfp4pv.vs16}{16.89~dB}
\pgfkeyssetvalue{/appendix/longcat-video/psnr/sage3.nvfp4}{12.36~dB}
\pgfkeyssetvalue{/appendix/longcat-video/psnr/vc4.vsmooth}{14.32~dB}
\pgfkeyssetvalue{/appendix/hunyuanvideo-1.5/prompt}{44}
\pgfkeyssetvalue{/appendix/hunyuanvideo-1.5/frames}{0, 30, 60, 90, and 120}
\pgfkeyssetvalue{/appendix/hunyuanvideo-1.5/count}{121}
\pgfkeyssetvalue{/appendix/hunyuanvideo-1.5/text}{An aerial view of Santorini during the blue hour, capturing the stunning architecture of white Cycladic buildings with blue domes against the twilight sky. \ldots{} A bird's-eye view with a wide-angle lens, focusing on the intricate details of the buildings and the vast expanse of the caldera.}
\pgfkeyssetvalue{/appendix/hunyuanvideo-1.5/psnr/reference}{reference}
\pgfkeyssetvalue{/appendix/hunyuanvideo-1.5/psnr/deco.fp8pv.vs128}{17.41~dB}
\pgfkeyssetvalue{/appendix/hunyuanvideo-1.5/psnr/trtllm.sage.fp8}{14.58~dB}
\pgfkeyssetvalue{/appendix/hunyuanvideo-1.5/psnr/vc.qk8}{13.26~dB}
\pgfkeyssetvalue{/appendix/hunyuanvideo-1.5/psnr/rtx.reference}{reference}
\pgfkeyssetvalue{/appendix/hunyuanvideo-1.5/psnr/deco.nvfp4pv.vs16}{16.79~dB}
\pgfkeyssetvalue{/appendix/hunyuanvideo-1.5/psnr/sage3.nvfp4}{8.57~dB}
\pgfkeyssetvalue{/appendix/hunyuanvideo-1.5/psnr/vc4.vsmooth}{15.57~dB}
\pgfkeyssetvalue{/appendix/minimax-h3/prompt}{11}
\pgfkeyssetvalue{/appendix/minimax-h3/frames}{0, 60, 121, 181, and 242}
\pgfkeyssetvalue{/appendix/minimax-h3/count}{243}
\pgfkeyssetvalue{/appendix/minimax-h3/text}{A cinematic film shot in 70mm, capturing an extreme close-up of a 24-year-old woman's eye as it blinks. \ldots{} A medium shot with a slightly elevated perspective, highlighting the natural movement of her eye.}
\pgfkeyssetvalue{/appendix/minimax-h3/psnr/reference}{reference}
\pgfkeyssetvalue{/appendix/minimax-h3/psnr/deco.fp8pv.vs128}{23.77~dB}
\pgfkeyssetvalue{/appendix/minimax-h3/psnr/trtllm.sage.fp8}{17.27~dB}
\pgfkeyssetvalue{/appendix/minimax-h3/psnr/vc.qk8}{19.52~dB}
\pgfkeyssetvalue{/appendix/minimax-h3/psnr/rtx.reference}{reference}
\pgfkeyssetvalue{/appendix/minimax-h3/psnr/deco.nvfp4pv.vs16}{20.47~dB}
\pgfkeyssetvalue{/appendix/minimax-h3/psnr/sage3.nvfp4}{16.52~dB}
\pgfkeyssetvalue{/appendix/minimax-h3/psnr/vc4.vsmooth}{17.82~dB}

%% file: results/appendix/prompts.tex
\pgfkeyssetvalue{/appendix/wan2.2-t2v-a14b/fulltext}{A cinematic wide-angle portrait of a man with his face illuminated by the warm glow of a TV screen. The man, with a rugged yet determined expression, leans forward slightly against a vintage wooden armchair. His dark hair is slightly disheveled, and he wears a worn leather jacket over a plain white shirt. The background features a cluttered living room with old books, newspapers, and a few framed photos scattered around. The TV shows static, with a faint image of a news broadcast flickering in the corner. The overall scene has a nostalgic and gritty feel, with a rich color palette and a soft, grainy texture. A wide-angle shot capturing the man's intense gaze and the warm ambiance of the room.}
\pgfkeyssetvalue{/appendix/longcat-video/fulltext}{A dynamic scene captured in the style of a vibrant food photography, showcasing a chef skillfully chopping onions in a bustling kitchen. The chef, a middle-aged man with a weathered face and determined expression, skillfully slices the onions with quick, practiced movements. He wears a white apron tied neatly around his waist and a chef's hat perched atop his head. The background is a well-equipped kitchen, with stainless steel appliances and countertops cluttered with various cooking tools and ingredients. Steam rises from a pot on the stove, and sunlight filters through the window, casting a warm glow. A medium shot with the chef at the center, capturing the intensity of his work.}
\pgfkeyssetvalue{/appendix/hunyuanvideo-1.5/fulltext}{An aerial view of Santorini during the blue hour, capturing the stunning architecture of white Cycladic buildings with blue domes against the twilight sky. The caldera views are breathtaking, with the volcanic cliffs and sea below creating a dramatic contrast. The lighting casts a soft, warm glow, enhancing the serene atmosphere. The image has a dreamy, almost ethereal quality, emphasizing the beauty of the setting. A bird's-eye view with a wide-angle lens, focusing on the intricate details of the buildings and the vast expanse of the caldera.}
\pgfkeyssetvalue{/appendix/minimax-h3/fulltext}{A cinematic film shot in 70mm, capturing an extreme close-up of a 24-year-old woman's eye as it blinks. The scene takes place during magic hour in Marrakech, with the vibrant colors of the setting sun casting warm hues over the bustling streets. The depth of field emphasizes the intricate details of her almond-shaped eyes, which reflect the lively atmosphere of the city. Her eyes, framed by long, dark lashes, are set against a backdrop of bustling market stalls, ornate architecture, and the soft shadows of the setting sun. The background features a blend of rich textures and vibrant colors, creating a sense of depth and immersion. A medium shot with a slightly elevated perspective, highlighting the natural movement of her eye.}

%% file: figures/appendix_videos.tex
\pgfkeysgetvalue{/appendix/blocks}{\appendixblocks}
\newcommand{\appendixvideo}[1]{%
\resizebox{\textwidth}{!}{%
\begin{tikzpicture}[font=\footnotesize]
\foreach \block [count=\b from 0] in \appendixblocks {
  \pgfkeysgetvalue{/appendix/block/\block/rows}{\appendixrows}
  \pgfkeysgetvalue{/appendix/block/\block/reference}{\appendixref}
  \node[anchor=south west, inner sep=0, outer sep=0] at (0, {-10.9 * \b + 0.1})
    {\textbf{\pgfkeysvalueof{/appendix/block/\block/gpu}}, PSNR against
     \pgfkeysvalueof{/appendix/label/\appendixref} on the same GPU};
  \foreach \key [count=\row from 0] in \appendixrows {
    \node[anchor=south west, inner sep=0, outer sep=0, yshift=0.6mm] at (0, {-10.9 * \b - 0.5 - 2.5 * \row})
      {\textbf{\pgfkeysvalueof{/appendix/label/\key}}\quad PSNR \pgfkeysvalueof{/appendix/#1/psnr/\key}};
    \foreach \k in {0, ..., 4}
      \node[inner sep=0, anchor=north west] at ({3.52 * \k}, {-10.9 * \b - 0.5 - 2.5 * \row})
        {\includegraphics[width=3.42cm]{results/appendix/#1/\key_\k.jpg}};
  }
}
\end{tikzpicture}}\par\smallskip
\parbox{\textwidth}{\footnotesize\itshape\raggedright ``\pgfkeysvalueof{/appendix/#1/text}''}}
\newcommand{\appendixcaption}[2]{\textbf{Videos of #2.} Prompt \pgfkeysvalueof{/appendix/#1/prompt} of MovieGen Bench, frames
\pgfkeysvalueof{/appendix/#1/frames} of \pgfkeysvalueof{/appendix/#1/count}. The upper block runs on GB300 and the lower
block on RTX PRO 6000, with the PSNR of each clip against BF16 attention on the same GPU.}

\begin{figure*}[p]
\centering
\appendixvideo{wan2.2-t2v-a14b}
\caption[Videos of Wan2.2-T2V-A14B]{\appendixcaption{wan2.2-t2v-a14b}{Wan2.2-T2V-A14B}}
\label{fig:videos-wan}
\end{figure*}

\begin{figure*}[p]
\centering
\appendixvideo{longcat-video}
\caption[Videos of LongCat-Video]{\appendixcaption{longcat-video}{LongCat-Video}}
\label{fig:videos-longcat}
\end{figure*}

\begin{figure*}[p]
\centering
\appendixvideo{hunyuanvideo-1.5}
\caption[Videos of HunyuanVideo-1.5]{\appendixcaption{hunyuanvideo-1.5}{HunyuanVideo-1.5}}
\label{fig:videos-hunyuan}
\end{figure*}

\begin{figure*}[p]
\centering
\appendixvideo{minimax-h3}
\caption[Videos of MiniMax-H3]{\appendixcaption{minimax-h3}{MiniMax-H3}}
\label{fig:videos-minimax}
\end{figure*}